%% file: thesis.tex
\documentclass[12pt,a4paper]{report}
\usepackage{thesis}
\usepackage{lmodern}
\usepackage[numbers,sort&compress]{natbib}
\usepackage[left=2.5cm,top=2.5cm,right=2.5cm,bottom=2.5cm]{geometry}
\usepackage{nomencl}
\usepackage{latexsym}
\usepackage{graphicx}
\usepackage{multirow}
\usepackage{url}
\usepackage{amssymb}
\usepackage{amsmath}
\usepackage{enumitem}
\usepackage{times}
\usepackage{subcaption}
\usepackage{float}
\DeclareMathOperator*{\argmax}{arg\,max}

\makenomenclature

\usepackage{tocloft}

\newcommand{\thesistitle}{Learning to Memorize in Neural Task-Oriented Dialogue Systems}
\newcommand{\thesisauthor}{Chien-Sheng (Jason) Wu}
\newcommand{\programname}{Electronic and Computer Engineering}
\newcommand{\departmentname}{Department of Electronic and Computer Engineering}
\newcommand{\thesisdate}{June 2019}
\newcommand{\signdate}{June 2019}
\newcommand{\thesisdegree}{MPhil}

\newcommand{\supervisorinfo}{Prof. Pascale Fung}
\newcommand{\otherprofinfo}{Prof. Qifeng Chen}
\newcommand{\depheadinfo}{Prof. Bertram Shi}

\begin{document}

\pagenumbering{roman}
\pagestyle{plain}
\setcounter{page}{1}
\addcontentsline{toc}{chapter}{Title Page}

\include{1_title}

\newpage
\addcontentsline{toc}{chapter}{Authorization Page}

\include{2_authorization}

\newpage
\addcontentsline{toc}{chapter}{Signature Page}

\include{3_signature}


\newpage
\addcontentsline{toc}{chapter}{Acknowledgments}
\include{5_acknowledgement}

\newpage
\addcontentsline{toc}{chapter}{Table of Contents}
\tableofcontents

\newpage
\addcontentsline{toc}{chapter}{List of Figures}
\listoffigures

\newpage
\addcontentsline{toc}{chapter}{List of Tables}
\listoftables

\newpage
\addcontentsline{toc}{chapter}{Abstract}
\include{abstract}


\newpage
\pagenumbering{arabic}
\pagestyle{plain}
\setcounter{page}{1}
\include{chapter1}

\include{chapter2}

\include{chapter3}

\include{chapter4}

\include{chapter5}

\include{chapter6}

\newpage
\addcontentsline{toc}{chapter}{Publication}
\include{publication}

\newpage
\addcontentsline{toc}{chapter}{Reference}
\bibliographystyle{IEEEtranN}
\include{reference}


\end{document}

%% file: 1_title.tex
\thispagestyle{empty}
\null\vskip0.5in
\begin{center}
  \begin{LARGE}
    \textbf{\thesistitle}
  \end{LARGE}
  \vfill
  \vspace{20mm}

  \begin{Large}

  by

  \vspace{4mm}

  \textbf{\thesisauthor} \\
  \end{Large}
  
  \vfill
  \vspace{20mm}

  A Thesis Submitted to\\
  The Hong Kong University of Science and Technology \\
  in Partial Fulfillment of the Requirements for\\
  the Degree of Master of Philosophy \\
  in \programname \\
  \vfill \vfill
  \thesisdate, Hong Kong
  \vfill
  \vfill
  \vfill
  \vfill
  Copyright \textsuperscript{\textcopyright} by Chien-Sheng Wu 2019
\end{center}

\vfill

%% file: 2_authorization.tex
\null\skip0.2in
\begin{center}
{\bf \Large \underline{Authorization}}
\end{center}
\vspace{12mm}

I hereby declare that I am the sole author of the thesis.

\vspace{10mm}

I authorize the Hong Kong University of Science and Technology to lend this
thesis to other institutions or individuals for the purpose of scholarly research.

\vspace{10mm}

I further authorize the Hong Kong University of Science and Technology to
reproduce the thesis by photocopying or by other means, in total or in part, at the
request of other institutions or individuals for the purpose of scholarly research.

\vspace{30mm}

\begin{center}
\underline{~~~~~~~~~~~~~~~~~~~~~~~~~~~~~~~~~~~~~~~~~~~~~~~~~~~~~~~~~~~~~~~~~~~~~~}\\
~~~~\thesisauthor \\
~~~~\signdate

\end{center}

%% file: 3_signature.tex
\begin{center}
{\LARGE \thesistitle}\\
\vspace{5mm}
{\large 
by\\
\vspace{3mm}
\thesisauthor
}\\
\vspace{5mm}
This is to certify that I have examined the above MPhil thesis\\
and have found that it is complete and satisfactory in all respects,\\
and that any and all revisions required by\\
the thesis examination committee have been made.
\end{center}

\vspace{15mm}

\begin{center}
\underline{~~~~~~~~~~~~~~~~~~~~~~~~~~~~~~~~~~~~~~~~~~~~~~~~~~~~~~~~~~~~~~~~~~~~~~~~~~~ }\\
\supervisorinfo , Thesis Supervisor
\end{center}


\vspace{15mm}
\begin{center}
\underline{~~~~~~~~~~~~~~~~~~~~~~~~~~~~~~~~~~~~~~~~~~~~~~~~~~~~~~~~~~~~~~~~~~~~~~~~~~~ }\\
\depheadinfo , Head of Department
\end{center}

\vspace{10mm}

\noindent\textbf{Thesis Examination Committee} \\

\begin{tabular}{ll}
    1. \depheadinfo  & Department of Electronic and Computer Engineering \\
    2. \supervisorinfo & Department of Electronic and Computer Engineering \\
    3. \otherprofinfo & Department of Computer Science and Engineering
\end{tabular}

\vfill

\begin{center}
\departmentname\\
\vspace{5mm}
\signdate
\end{center}

%% file: 5_acknowledgement.tex
\centerline{{\bf \Large Acknowledgments}} \vspace{5mm} \noindent

First and foremost I would like to thank my supervisor, Professor Pascale Fung. As an exchange student and a graduate student, Pascale provided me extremely valuable mentorship both in academics and in life. Through her careful guidance in the past two years, I entered the natural language processing field, found my own research interests, published my first top conference paper, and connected to the research and industrial communities. Pascale inspired me by her unique vision and passionate attitude. She trusted me like nobody else, and encouraged me to dream big and focus on big problems. I am extremely lucky and proud to have been advised by Pascale.

I appreciate Professor Bertram Shi and Professor Qifeng Chen for their time in being my thesis examination committee. I also had the fortune to gain research and development experience in the industry during my internship at Salesforce. I would like to thank Dr. Richard Socher for sharing his comprehensive views and giving me on-point advice. Thanks to my great mentor Dr. Caiming Xiong for guiding me to conduct impactful research and giving me the freedom to explore. I would like to express my appreciation to the whole Salesforce research team for the insightful research discussions and brainstorming. 

The last two years at HKUST have been an amazing learning and playing experience with my lab friends. Andrea Madotto, thanks for all the in-depth discussions and close collaboration on dialogue research. Peng Xu, thanks for insightful consultation on affective computing. Genta Winata, thanks for leading the code-switching research group. Nayeon Lee, thanks for collaborating on the fact-checking project. Thanks also to Zhaojiang Lin, Jamin Shin, Jiho Park, Farhad Siddique, and many many others for the great exchange of research experience and ideas. Without their time discussing and working with me, my graduate life would not have gone this well, nor would my thesis. I wish them all the best in their research and look forward to meeting them in the near future.

Finally, and most importantly, I would like to send love to my whole family, Chia-Chin Wu, Mei-Chueh Wang, and Chun-Ping Wu, for their continuous and incomparable love, support, and encouragement. I also want to express my deepest gratitude to Nai-Wen Hu for making me a better person and motivating me to chase my dream. It was they who always stood behind me giving me endless support with all of their hearts. This thesis is dedicated to them.

%% file: abstract.tex
\begin{center}
{\LARGE \thesistitle}\\
\vspace{20mm}
{\large by \\
\thesisauthor\\
\departmentname\\
The Hong Kong University of Science and Technology
}
\end{center}
\vspace{8mm}
\begin{center}
{\LARGE Abstract}
\end{center}
\par
\noindent

Dialogue systems are designed to communicate with human via natural language and help people in many aspects. Task-oriented dialogue systems, in particular, aim to accomplish users goal (e.g., restaurant reservation or ticket booking) in minimal conversational turns. 
The earliest systems were designed with a large amount of hand-crafted rules and templates by experts, which were costly and limited. 
Therefore, data-driven statistical dialogue systems, including the powerful neural-based systems, have considerable attention over last few decades to reduce the cost and provide robustness.

One of the main challenges in building neural task-oriented dialogue systems is to model long dialogue context and external knowledge information. Some neural dialogue systems are modularized. Although they are known to be stable and easy to interpret, they usually require expensive human labels for each component, and have unwanted module dependencies. On the other hand, end-to-end approaches learn the hidden dialogue representation automatically and directly retrieve/generate system responses. They require much less human involvement, especially in the dataset construction. However, most of the existing models suffer from incorporating too much information into end-to-end learning frameworks.

In this thesis, we focus on learning task-oriented dialogue systems with deep learning models, which is an important research direction in natural language processing. We leverage the neural copy mechanism and memory-augmented neural networks to address the existing challenge of modeling and optimizing information in conversation. We show the effectiveness of our strategy by achieving state-of-the-art performance in multi-domain dialogue state tracking, retrieval-based dialogue systems, and generation-based dialogue systems. 

We first improve the performance of a dialogue state tracking module, which is the core module in modularized dialogue systems. Unlike most of the existing dialogue state trackers, which are over-dependent on domain ontology and lacking knowledge sharing across domains, our proposed model, the transferable dialogue state generator, leverages its copy mechanism to get rid of ontology, share knowledge between domains, and memorize the long dialogue context. We also evaluate our system on a more advanced setting, unseen domain dialogue state tracking. We empirically show that TRADE enables zero-shot dialogue state tracking and can adapt to new few-shot domains without forgetting the previous domains. 

Second, we utilize two memory-augmented neural networks, the recurrent entity network and dynamic query memory network, to improve end-to-end retrieval-based dialogue learning. They are able to capture dialogue sequential dependencies and memorize long-term information. We also propose a recorded delexicalization copy strategy to simplify the problem by replacing real entity values with ordered entity types. Our models are shown to surpass other retrieval baselines, especially when the conversation has a large amount of turns.

Lastly, we tackle end-to-end generation-based dialogue learning with two successive proposed models, the memory-to-sequence model (Mem2Seq) and global-to-local memory pointer network (GLMP). Mem2Seq is the first model to combine multi-hop memory attention with the idea of the copy mechanism, which allows an agent to effectively incorporate knowledge base information into a generated response. It can be trained faster and outperforms other baselines in three different task-oriented dialogue datasets, including human-human dialogues. Moreover, GLMP is an extension of Mem2Seq, which further introduces the concept of response sketching and double pointers copying. We empirically show that GLMP surpasses Mem2Seq in terms of both automatic evaluation and human evaluation, and achieves the state-of-the-art performance.

%% file: chapter1.tex
\chapter{Introduction}

\section{Motivation and Research Problems} 
Dialogue systems, known as conversational agents or chatbots, can communicate with human via natural language to assist, inform and entertain people. 
They have become increasingly important in both research and industrial communities.
Such systems can be split into two categories: chit-chat conversational systems and task-oriented dialogue systems, shown in Table~\ref{TB:DIALOGUE_EXP}, with the former designed to keep users company and engage them with a wide range of topics, and the latter designed to accomplish specific tasks, such as restaurant reservation or ticket booking.
Task-oriented dialogue systems are required to understand user requests, ask for clarification, provide related information, and take actions.
Unlike chit-chat systems, these systems also usually involve tracking the intentions of users, retrieving information from external databases, and planning for multi-turn conversations. 
In this thesis, we focus on task-oriented dialogue systems.

Usually, task-oriented dialogue systems have been built modularly, with modules for spoken language understanding (SLU)~\cite{raymond2007generative,deng2012use}, dialogue state tracking (DST)~\cite{williams2007partially,thomson2010bayesian,henderson2014robust}, dialogue management (DM)~\cite{rudnicky1999agenda,young2006using,young2013pomdp}, and natural language generation (NLG)~\cite{busemann1998flexible}. 
The SLU module performs semantic decoding, and passes the information to the DST module to maintain the states of the dialogue. The DM module subsequently takes dialogue states and produces dialogue actions for the next utterance. The dialogue action is then passed to the NLG module to obtain the final system response.

These components traditionally are designed with a large amount of hand written rules and templates by experts. 
Motivated by the need for a data-driven framework to reduce the cost of laboriously hand-crafting dialogue managers and to provide robustness against the errors created by speech recognition, statistical dialogue systems were first introduced to include an explicit Bayesian model of uncertainty and optimize the policy, e.g., the partially observable Markov decision processes (POMDPs)~\cite{young2013pomdp}.
Next, neural networks and deep learning, a specific set of algorithms for function approximation, are now transforming the natural language processing (NLP) field, e.g, neural statistical dialogue systems have considerable attention in recent years. 
With labelled data, a model can learn to minimize loss function through iterative gradient update (back-propagation) of its parameters.

Even though these neural statistical modularized systems are known to be stable and easy to interpret by combining domain-specific knowledge and a slot-filling technique, they usually have the following drawbacks:
1) Complicated human annotated labels are required. For example, SLU and DST need labels for every domain and slot, DM requires dialogue experts to label dialogue actions and slot information, and NLG needs comprehensive language templates or human rule;
2) Dependencies between modules are complex, which may result in serious error propagation. In addition, the interdependent modules in modularized systems may result in performance mismatch, e.g., the update of the down-stream module may cause other upper-stream modules to be sub-optimal.
3) Generalization ability to new domains or new slots is limited. With too specific domain knowledge in each module, it is difficult to extend the modularized architecture to a new setting or transfer the learned knowledge to a new scenario.
4) Knowledge base (KB) interpretation requires additional human defined rule. There is no neural memory architectures that are designed to learn and represent the database information.

\begin{table}[t]
\caption{Dialogue examples for chit-chat and task-oriented dialogue systems.}
\label{TB:DIALOGUE_EXP}
\resizebox{\linewidth}{!}{
\begin{tabular}{|r|c|c|}
\hline
\multicolumn{1}{|c|}{} & \textbf{Chit-Chat Dialogue Systems} & \textbf{Task-oriented Dialogue Systems} \\ \hline
\textit{System} & Hello, how are you today? & Hello, what can I help with you today? \\
\textit{User} & I got the job promotion. & Can you help me book a dinner at 6? \\
\textit{System} & That is awesome! Congrats! & No problem. Where should it be? \\
\textit{User} & Thank you! Gonna celebrate it at bar tonight. & Nearby Taipei city for 4. \\
\textit{System} & Definitely! Don't be too drunk haha. & What cuisine do you prefer? \\
\textit{User} & But it is boring to drink alone. & Let's try Japanese food... Sushi maybe. \\
\textit{System} & Ask more friends to join! & What do you think about this option: Sushi Bar? \\
\textit{User} & They don't want to come out tonight... & Sounds good! Please make a reservation. \\
\textit{System} & Why? Today is Sunday why not? & Done. Anything else I can help? \\
\textit{User} & Because they gonna watch Game of Thrones. & No thank you. \\
\textit{System} & Right people are crazy about that! & You are welcome. See you next time. \\ \hline
\end{tabular}
}
\end{table}

End-to-end neural approaches are an alternative to the traditional modularized solutions for task-oriented dialogue systems. These approaches train the model directly on text transcripts of dialogues, learn a distributed vector representation of the dialogue states automatically and retrieve or generate the system response in the end.
Everything in an end-to-end model is learned together with the joint objective functions.
In this way, the models make no assumption on the dialogue state structure and additional human labels, gaining the advantage of easily scaling up.
Specifically, using recurrent neural networks (RNNs) is an attractive solution, where the latent memory of the RNN represents the dialogue states. 
However, existing end-to-end approaches in task-oriented dialogue systems still suffer from the following problems:
1) They struggle to effectively incorporate dialogue history and external KB information into the RNN hidden states since RNNs are known to be unstable over long sequences. Both of them are essential because dialogue history includes information about users goal and external KB has the information that need to be provided (as shown in Table~\ref{TB:knowledge_base}).
2) Processing long sequences using RNNs is very time-consuming, especially when encoding the whole dialogue history and external KB using an attention mechanism.
3) Correct entities are hard to generate from the predefined vocabulary space, e.g., restaurant names or addresses. Additionally, these entities are relatively important compared to the chit-chat scenario because it is usually the expected information in the system response. For example, a driver expects to get the correct address of the gas station rather than a random place, such as a gym.

\begin{table}[t]
\caption{A knowledge base example for task-oriented dialogue systems.}
\label{TB:knowledge_base}
\centering
\resizebox{0.8\linewidth}{!}{
\begin{tabular}{|c|c|c|c|c|}
\hline
\textbf{Point-of-Interest} & \textbf{Distance} & \textbf{Traffic Info} & \textbf{POI Type} & \textbf{Address} \\ \hline
Maizuru & 5 miles & moderate traffic & japanese restaurant & 329 El Camino Real \\ \hline
Round Table & 4 miles & no traffic & pizza restaurant & 113 Anton Ct \\ \hline
World Gym & 10 miles & heavy traffic & gym and sports & 256 South St \\ \hline
Mandarin Roots & 5 miles & no traffic & chinese restaurant & 271 Springer Street \\ \hline
Palo Alto Cafe & 4 miles & moderate traffic & coffee or tea place & 436 Alger Dr \\ \hline
Dominos & 6 miles & heavy traffic & pizza restaurant & 776 Arastradero Rd \\ \hline
Sushi Bar & 2 miles & no traffic & japanese restaurant & 214 El Camino Real \\ \hline
Hotel Keen & 2 miles & heavy traffic & rest stop & 578 Arbol Dr \\ \hline
Valero & 3 miles & no traffic & gas station & 45 Parker St \\ \hline
\end{tabular}
}
\end{table}

We propose to augment neural networks with external memory and a neural copy mechanism to address the challenges of modeling long dialogue context and external knowledge information in task-oriented dialogue learning.
Memory-augmented neural networks (MANNs)~\cite{graves2014neural,weston2014memory,sukhbaatar2015end} can be leveraged to maintain long-term memory, enhance reasoning ability, speed up the training process, and strengthen the neural copy mechanism, which are all desired features to achieve better memorization of information and better conversational agents.
A MANN writes external memory into its memory modules and uses a memory controller to read and write memories repeatedly. 
This approach can memorize external information and rapidly encode long sequences since it usually does not require auto-regressive (sequential) encoding. 
Moreover, a MANN usually includes a multi-hop attention mechanism, which has been empirically shown to be essential in achieving high performance on reasoning tasks, such as machine reading comprehension and question answering. A copy mechanism, meanwhile, allows a model to memorize words and directly copy them from input to output, which is crucial to successfully generate correct entities. 
It can not only reduce the generation difficulty but is also more like human behavior. 
Intuitively, when humans want to tell others the address of a restaurant, for example, they need to ``copy'' the information from the internet or their own memory to their response. 

In this thesis, we focus on \textit{neural task-oriented dialogue learning} that can effectively \textit{incorporate long dialogue context} and \textit{external knowledge} information. We fist demonstrate how to memorize long dialogue context in dialogue state tracking tasks, including single-domain, multi-domain, and unseen-domain settings. Then we show how to augment neural networks with memory and copy mechanism to memorize long dialogue context and external knowledge for both retrieval-based and generation-based dialogue systems.

\section{Thesis Outline}
The rest of the thesis is organized as:
\begin{itemize}[leftmargin=*]
    \item Chapter 2 introduces the background and related work on task-oriented dialogue systems, sequence text generation, copy mechanisms, and memory-augmented neural networks. 
     \item Chapter 3 presents the transferable dialogue state generator to effectively generate dialogue states with a copy mechanism. We further extend the model to multi-domain dialogue state tracking and unseen domain dialogue state tracking. 
    \item Chapter 4 presents two memory-augmented neural networks, a recurrent entity network and dynamic query memory network, with recorded delexicalization copying for end-to-end retrieval-based dialogue learning. These two models are able to memorize long-term sequential dependency in dialogue.
    \item Chapter 5 introduces two proposed state-of-the-art dialogue generation models: memory-to-sequence and global-to-local memory pointer network. These two models combine multi-hop attention mechanisms with the idea of the copy mechanism, which allows us to effectively incorporate long context information. 
    \item Chapter 6 summarizes this thesis and discusses possible future research directions.
\end{itemize}

%% file: chapter2.tex
\chapter{Background and Related Work}

In task-oriented dialogue systems, there are several terms that will be used frequently: dialogue history, domains, intentions, slots, slot values, states, and external knowledge base (KB). Dialogue history does not mean the whole conversational history between chatbots and users, instead it is the whole dialogue context in the current conversation. Domains are the topics of the current conversation, for example, restaurant domain is about restaurant reservation and taxi domain is about booking a taxi to somewhere. Intentions are the goals of each user utternace, for example, the intention of saying ``\textit{Is it going to rain tomorrow?}'' is to check weather, and the intention of saying ``\textit{Schedule a party at 7 on Friday}'' is to add schedule. Slots are the predefined variables in the dialogue that can be filled with all kinds of slot values. For example, a location slot can have its slot values such as Hong Kong and Taipei. Dialogue states are the semantic decoding results such as slot-value pairs. Tracking the states can be viewed as the Markov decision process. Lastly, external KB is the stored information that could be provided to users, which is large and dynamic. For example, there could be many possible restaurants that meet the criteria, and the weather at each city might change everyday.

In this chapter, we first review the existing modularized and end-to-end task-oriented dialogue systems. 
We then introduce sequence generation using recurrent neural networks in natural language processing, and attention-based copy mechanisms.
Lastly, we cover the general idea of memory-augmented neural networks and end-to-end memory networks in detail.

\section{Modularized Task-Oriented Dialogue Systems}

\begin{figure}[t]
\centering
\includegraphics[width=\linewidth]{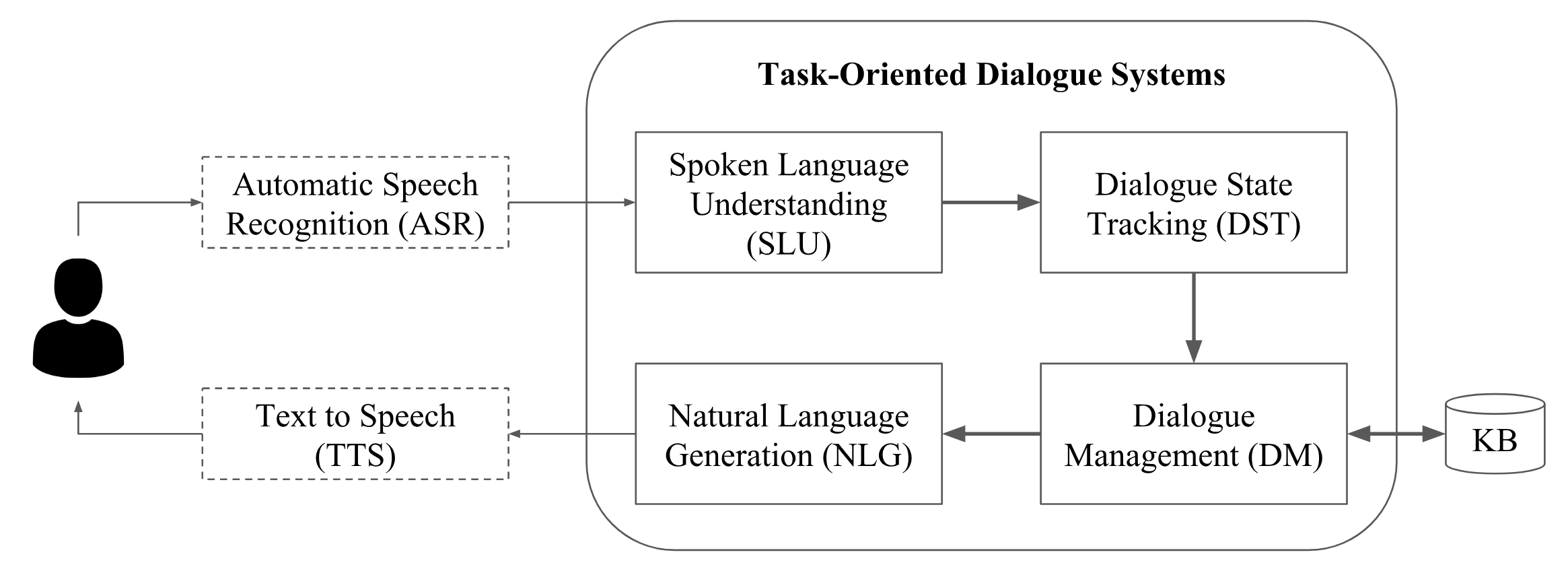} 
\caption{Block diagram of modularized task-oriented dialogue systems.}
\label{FIG:TODSblock}
\end{figure}

The block diagram of a modularized task-oriented dialogue systems is shown in Figure~\ref{FIG:TODSblock}. The automatic speech recognition (ASR) module first transcribes users' speech into text format. The speech transcriptions are passed to the spoken language understanding module, which usually performs domain classification, intention detection, and slot-filling. The information extracted by the SLU module is next passed to the dialogue state tracking module, known as belief tracking, to maintain the states of the dialogue. The dialogue management module subsequently takes dialogue states and retrieved information from KB as input and produces dialogue actions for the next utterance. The dialogue action is then passed to the natural language generation module to obtain the natural language system response. Lastly, the text-to-speech (TTS) module transforms the text into speech and replies to users. In this thesis, we focus on all the modules except the ASR and TTS; that is, we focus on natural language processing in task-oriented dialogue systems.

\subsection{Spoken Language Understanding}
The SLU module~\cite{raymond2007generative,deng2012use,yao2014spoken,guo2014joint,zhang2016joint} typically involves identifying users intentions and extracting semantic components from the user utterances.
Intention detection ~\cite{tur2012towards,chen2016zero} can be framed as a semantic utterance classification problem. 
Given a sequence of words, the goal is to predict an intent class from intent candidates. 
Slot-filling~\cite{nguyen2007comparisons,mesnil2015using,kurata2016leveraging} extracts semantic components by searching inputs to fill in values for predefined slots, which is a sequence labeling task that assigns a semantic label to each word.
Slots are basically variables in utterances, which can have their own values but are by default empty.
A slot and its value together can be viewed as a slot-value pair to represent the dialogue semantics. 

\subsection{Dialogue State Tracking}
The DST module is a crucial component in task-oriented dialogue systems, where the goal is to extract user goals/intentions during a conversation as compact dialogue states.
Given the extracted results from the SLU module, the DST module is usually expressed by a list of goal slots and the probability distribution of the corresponding candidate values.
Early systems for dialogue state tracking~\cite{williams2007partially,thomson2010bayesian,henderson2014robust} relied on hand-crafted features and a complex domain-specific lexicon and domain ontology.
To reduce human effort and overcome the limitations of modeling uncertainties, learning-based dialogue state trackers have been proposed in the literature ~\cite{williams2016dialog, NBT,rastogi2017scalable,P18-1134SpanPtr,MDBT,zhong2018global}.

\subsection{Dialogue Management}
The DM module~\cite{rudnicky1999agenda,young2006using,young2013pomdp} generates dialogue actions based on the dialogue states from the DST module and the retrieved information from the external KBs. 
The dialogue actions usually consist of a speech action (e.g., inform, request) and grounding information represented by slots and values.
Moreover, several approaches use reinforcement learning methods to learn the policy for predicting the actions given by states ~\cite{li2009reinforcement,lipton2018bbq,gao2018neural}.

\subsection{Natural Language Generation}
The most widely used method in NLG is template-based~\cite{busemann1998flexible}, which the response sentences are manually designed to be the output for all the possible dialogue actions in DM.
The main advantages include the ease with which developers can control the system.
However, building such a system is expensive when there are many dialogue actions, and it cannot handle the ones that not previously designed. 
To address these problems, machine learning-based NLG~\cite{wen2015semantically,press2017language} has been actively researched, where a corpus consisting of only dialogue actions and utterances.

\section{End-to-End Task-Oriented Dialogue Systems}
\subsection{Overview}
Unlike modularized systems that use a pipeline to connect each module, end-to-end neural systems are designed to train directly on text transcripts and the KB information.
Inspired by the success of end-to-end training of chit-chat dialogue systems~\cite{shang-lu-li:2015:ACL-IJCNLP,fung2016zara,fung2016towards,fung2018empathetic}, researchers have recently started exploring end-to-end solutions for task-oriented dialogue systems.
RNNs, such as long short-term memory networks~\cite{hochreiter1997long} (LSTMs) and gated recurrent units~\cite{chung2014empirical} (GRUs), play important roles in this direction due to their ability to create latent representations, avoiding the need for artificial labels.
\cite{wen2016network} proposed an end-to-end trainable neural dialogue model in which each system component is connected, and each module is trained separately.
However, each component is pre-trained separately with different objective functions.
\cite{bordes2016learning} and \cite{perez2016gated} used end-to-end memory networks (MNs)~\cite{sukhbaatar2015end} to learn utterance representations, and applied them to the next-utterance prediction task.
Although the performance is similar to existing retrieval approaches, these models do not model the utterance dependencies in the memory.
\cite{seo2016query}, meanwhile, considered utterances as a sequence of state-changing triggers and reduced the memory query to a more informed query through time.
\cite{williams2017hybrid} proposed hybrid code networks (HCNs) that combine RNNs with domain-specific human rule, and \cite{lei2018sequicity} incorporated explicit dialogue state tracking into a delexicalized sequence generation.

Most related to this thesis, \cite{eric-manning:2017:EACLshort} is the first work that applied a neural copy mechanism into the sequence-to-sequence architecture for the task-oriented dialogue task. It showed that the performance of its generation model with the copy mechanism was competitive with other retrieval models on a simulated dialogue dataset. However, the authors encoded all the information, including the dialogue history and external KB, using a single RNN, which makes the encoding process time-consuming and makes it hard to learn good representations. This drawback is the motivations for our proposed architectures, which will be introduced in the following chapters on end-to-end dialogue response generation.


\subsection{Recurrent Neural Networks}
\begin{figure}[h]
\centering
\includegraphics[width=0.8\linewidth]{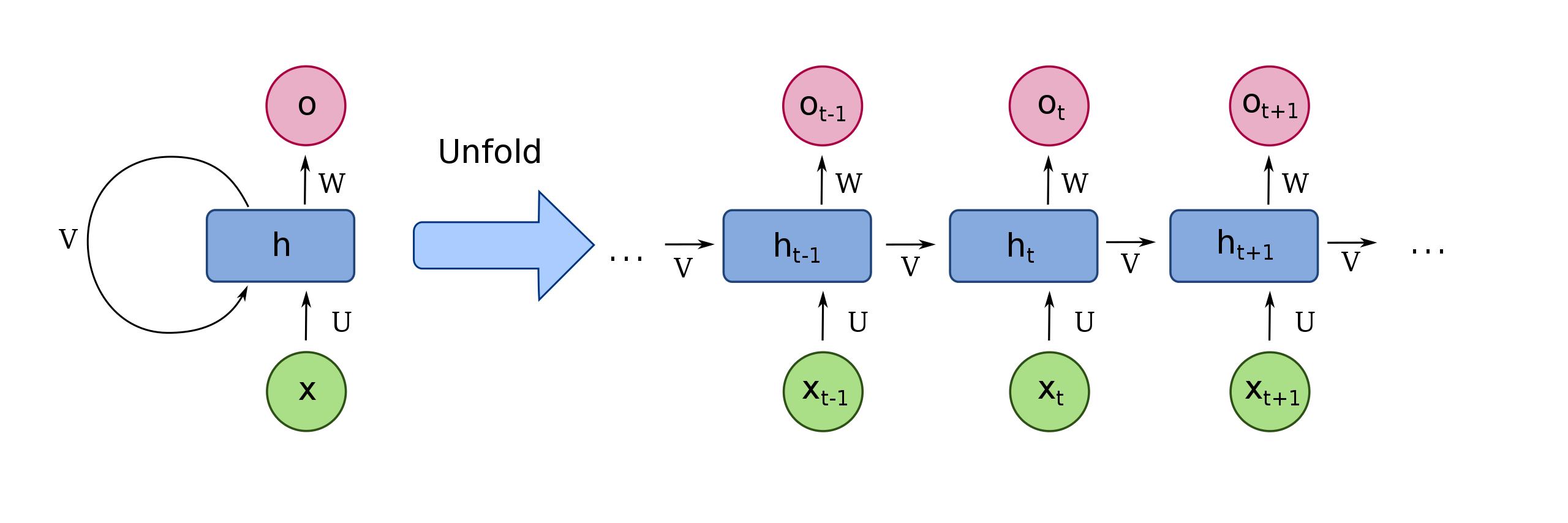} 
\caption{Unfolded basic recurrent neural networks.}
\label{FIG:RNN}
\end{figure}

A recurrent neural network (RNN), shown in Figure~\ref{FIG:RNN}, is a class of artificial neural network where connections between nodes form a directed graph along a temporal sequence. This allows it to exhibit dynamic temporal behavior. Differently from feed-forward neural networks, RNNs can use their internal state (memory) to process sequences of inputs. A widely used RNN architecture is the Elman type RNN~\cite{schuster1997bidirectional}. The RNN feeds the hidden layer output at time $t - 1$ back to the same hidden layer at time $t$ via recurrent connections. Thus, information stored in the hidden layer can be viewed as a summary of input sequence up till the current time. Then a non-linear and differentiable activation function is applied to the weighted sum of the input vector and the previous hidden state. The hidden state at time $t$ can thus be expressed as:
\begin{equation}
    h_t = \sigma(U x_t + V h_{t-1} + b),
\end{equation}
where $\sigma$ is the non-linear activation function. In RNN training, parameters in the network can be optimized based on loss functions derived using maximum likelihood. Network parameters are updated using back-propagation through time (BPTT) method considering influence of past states through recurrent connections. Error from the output layer is back-propagated to the hidden layers through the recurrent connections backwards in time. The weight matrices are updated after each training sample or mini-batch.

\subsubsection{Long Short-Term Memory and Gated Recurrent Units}
Long short-term memory (LSTM)~\cite{hochreiter1997long} and gated recurrent units (GRUs)~\cite{Cho2014LearningPR} are variances of original recurrent neural network. 
The LSTM was followed by the Gated Recurrent Unit (GRU), and both have the same goal of tracking long-term dependencies effectively while mitigating the vanishing/exploding gradient problems. 
The LSTM does so via input, forget, and output gates: the input gate regulates how much of the new cell state to keep, the forget gate regulates how much of the existing memory to forget, and the output gate regulates how much of the cell state should be exposed to the next layers of the network. 
On the other hand, the GRU operates using a reset gate and an update gate. The reset gate sits between the previous activation and the next candidate activation to forget previous state, and the update gate decides how much of the candidate activation to use in updating the cell state.
In this thesis, we utilize GRUs in most of the experiments because it has fewer parameters than LSTM and could be trained faster.

\subsection{Sequence-to-Sequence Models}
\begin{figure}[t]
\centering
\includegraphics[width=\linewidth]{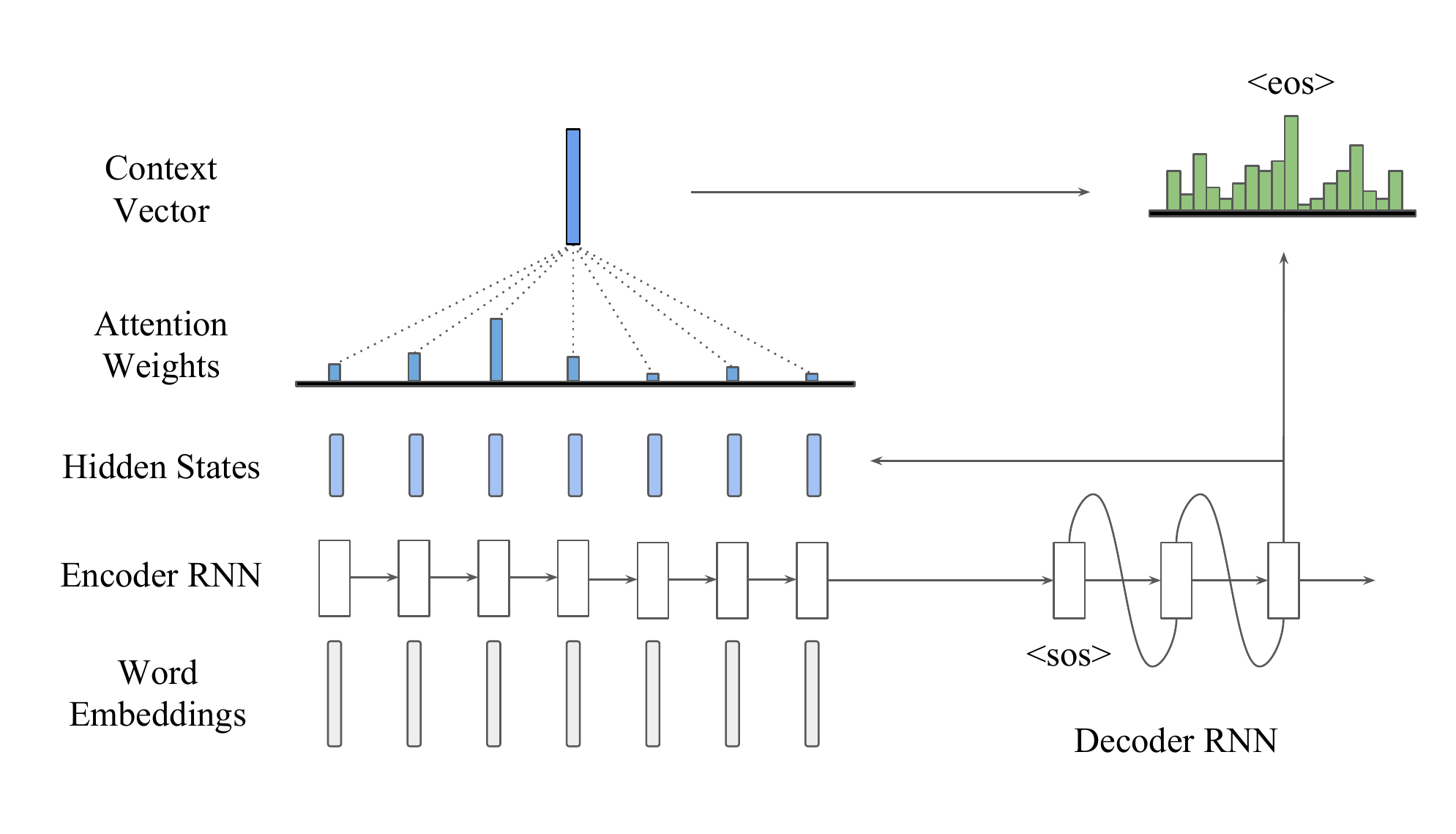} 
\caption{A general view of encoder-decoder structure with attention mechanism.}
\label{FIG:EncDec}
\end{figure}

As shown in Figure~\ref{FIG:EncDec}, the most common and powerful conditional generative model for natural language is the sequence-to-sequence (Seq2Seq) model~\cite{Cho2014LearningPR, serban2017hierarchical}, which is a type of encoder-decoder models.
Seq2Seq models the target word sequence $Y$ conditioned on given word sequence $X$, that is, $P(Y|X)$.
The basic Seq2Seq uses an encoder $\text{RNN}^{enc}$ to encode input $X$, and a decoder $\text{RNN}^{dec}$ to predict words in output $Y$.
Note that the encoder and decoder are not necessarily RNNs. Different kinds of sequence modeling approaches are possible alternatives.
Let $w^x_i$ and $w^y_j$ be the $i^{th}$ and $j^{th}$ words in the source and target sequences, respectively.
In most machine learning-based natural language applications, instead of representing words using one-hot vectors, words in vocabulary are usually represented by word embeddings, fixed-length vectors with real numbers.
Embeddings can either be randomly initialized and learned through loss optimization or be pre-trained using a distributional semantics hypothesis~\cite{mikolov2013distributed,xu2018emo2vec}.
Afterwards, the source sequence is encoded by recursively applying:
\begin{gather}
    h^{enc}_0 = 0, \\
    h^{enc}_i = \text{RNN}^{enc}(w^x_i, h^{enc}_{i-1}).
\label{}
\end{gather}
Then the last hidden state of the encoder $h^{enc}_{|X|}$ is viewed as the representation of $X$ to initialize the decoder hidden state.
The decoder then predicts the words in target $Y$ sequentially via:
\begin{gather}
    h^{dec}_j = \text{RNN}^{dec}(w^y_j, h^{dec}_{j-1}), \\
    o_j = \text{Softmax}(W h^{dec}_j + b),
\label{}
\end{gather}
where $o_j$ is the output probability for every word in the vocabulary at time step $j$.
In addition, in order to make the model predict the first word and to terminate the prediction, special SOS (start-of-sentence) and EOS (end-of-sentence) tokens are usually padded at the beginning and the end of the target sequence.

\subsection{Attention Mechanism}
Although the standard Seq2Seq model is able to learn the long-term dependency in theory, it often struggles to deal with long-term information in practice.
An attention mechanism ~\cite{bahdanau2014neural, Luong2015EffectiveAT} is an important extension of the Seq2Seq models, mimicking the word alignment in statistical machine translation. 
The intuitive idea is that instead of solely depending on a fixed-length encoded vector, the attention mechanism allows the decoder to create dynamic encoded representations for each decoding time step by the weighted-sum encoded vector.
Let $H^{enc}=\{h^{enc}_1,\dots,h^{enc}_{|X|}\}$ be the hidden states of the encoder $\text{RNN}^{enc}$. Then at each decoding time step the decoder predicts the output distribution by
\begin{gather}
    h^{dec}_j = \text{RNN}^{dec}(w^y_j, h^{dec}_{j-1}), \\
    u^i_j = \text{Match}(h^{enc}_i, h^{dec}_j), \\
    \alpha_j = \text{Softmax}(u_j), \\
    c_j = \Sigma_i^{|X|} (\alpha^i_j h^{enc}_i), \label{eq:context} \\
    o_j = \text{Softmax}(W [h^{dec}_j; c_j] + b). \label{eq:p_vocab}
\end{gather}
The $\alpha$ vector is the attention score (probability distribution) computed by a matching function, which can be simple cosine similarity, linear mapping, or a neural network, as
\begin{gather}
\text{Match}(h_i, h_j) = 
\left\{
     \begin{array}{lr}
     h_i h_j^\top, \quad \text{(dot)} &  \\
     h_i W h_j^\top, \quad \text{(general)} &  \\
     \text{tanh}(W[h_i; h_j]). \quad \text{(concat)} &  \\
     \end{array}
\right.
\end{gather}
The $c_j$ in Eq. (\ref{eq:context}) is the context vector in decoding time step $j$, which is the weighted-sum of the encoder hidden states based on the attention weights.

\subsection{Copy Mechanism}
A copy mechanism is a recently proposed extension of attention mechanisms.
Intuitively, it encourages the decoder to learn how to ``copy'' words from the input sequence $X$.
The main advantage of the copy mechanism is its ability to handle rare words or out-of-vocabulary (OOV) words.
There are three common strategies to perform the copy mechanism: index-based, hard-gate, and soft-gate. 
Index-based copying usually produces the start and end positional indexes, and copies the corresponding text from the input source; hard-gate and soft-gate copying usually have two distributions, one over the vocabulary space and the other over the source text. Hard-gate copying uses a learned gating function to switch between two distributions, while soft-gate copying, on the other hand, combines two distributions into one with a learned scalar.

Pointer networks~\cite{Vinyals2015PointerN} were the first model to perform an index-based copy mechanism, which directly generates indexes corresponding to positions in the input sequence via:
\begin{gather}
    o_j = \text{Softmax}(u_j).
\label{}
\end{gather}
In this way, the output distribution is the attention distribution, and the output word is the input word that has highest probability.

\begin{figure}[h]
\centering
\includegraphics[width=\linewidth]{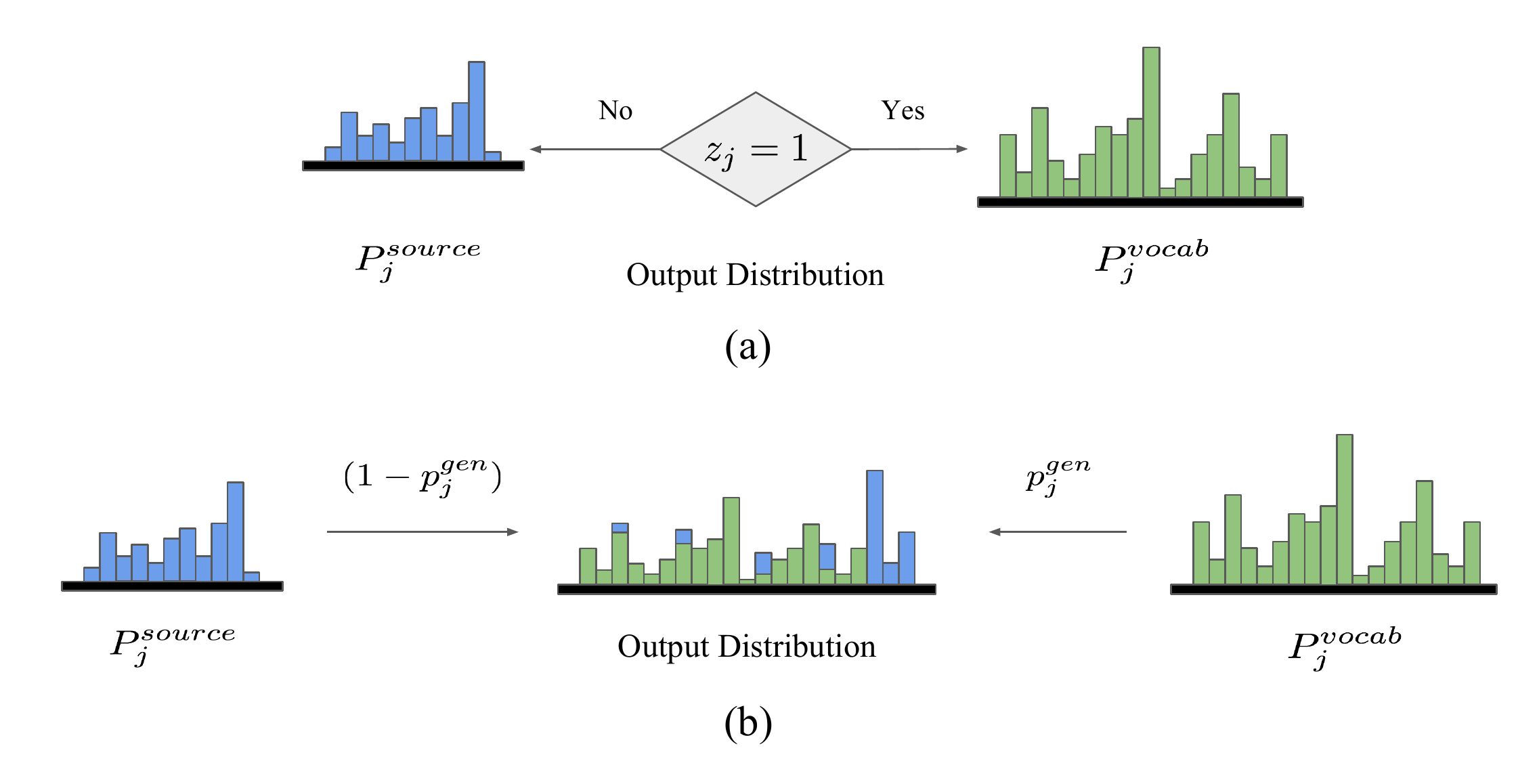} 
\caption{Copy mechanisms: (a) hard-gate and (b) soft-gate. The controllers $z_j$ and $p^{gen}_j$ are context-dependent parameters.}
\label{FIG:GATING}
\end{figure}

During decoding, hard-gate copying~\cite{gulcehre2016pointing, Dehghani:2017:LAC:3132847.3133010}, shown in Figure~\ref{FIG:GATING}(a) uses a switch to select distributions, and picks up the output word from the selected distribution.
The switching probability $z_j$ is modeled as a multi-layer perceptron with a binary output.
The concept is similar to pointer networks but the decoder retains the ability to generate output words from the predefined vocabulary distribution:
\begin{gather}
    P^{vocab}_j = \text{Softmax}(W [h^{dec}_j; c_j] + b), \\
    P^{source}_j = \text{Softmax}(u_j), \\
    z_j = 
    \begin{cases} 
        1 &\mbox{if } \text{Sigmoid}(f(w^{dec}_{j}, h^{dec}_{j-1})) > 0.5, \\ 
        0 &\mbox{otherwise},
    \end{cases}, \\
    o_j = 
    \begin{cases} 
        P^{vocab}_j &\mbox{if } z_j=1, \\ 
        P^{source}_j &\mbox{otherwise}.
        \label{eq:hard_gate}
    \end{cases}
\end{gather}
As shown in Eq. (\ref{eq:hard_gate}), the output distribution is dependent on $z_j$ to switch between $P^{vocab}_j$ and $P^{source}_j$.
In this way, the model is able to generate an unknown word in the vocabulary by directly copying a word from the input to output. 
Usually, there are multiple objective functions combined to learn the output generation, at least one for the vocabulary space supervision and one for the gating function supervision.

The soft-gate copy mechanism~\cite{Merity2016PointerSM,Gu2016IncorporatingCM,McCann2018TheNL,he-EtAl:2017:Long1}, shown in Figure~\ref{FIG:GATING}(b), on the other hand, combines the two distributions into one output distribution and generates words.
Usually a context-dependent scalar $p^{gen}_j$ is learned to weighted-sum the distributions:
\begin{gather}
    p^{gen}_j = \text{Sigmoid}(W [h^{dec}_t; w^{dec}_t; c_j]), \\
    o_j = p^{gen}_j \times P^{vocab}_j + (1-p^{gen}_j) \times P^{source}_j.
\label{eq:soft-copy}
\end{gather}
In this way, the output distribution is weighted by the source distribution, given a higher probability that the word appears in the input.
Note that words generated by the soft-gate copy mechanism are not constrained by the predefined vocabulary, i.e., the input unknown words will be concatenated with the vocabulary space.
In this thesis, we adopt the hard-gate and soft-gate copying strategies.

\section{Memory-Augmented Neural Networks}
\label{SEC:MANN}

\begin{figure}[h]
\centering
\includegraphics[width=0.9\linewidth]{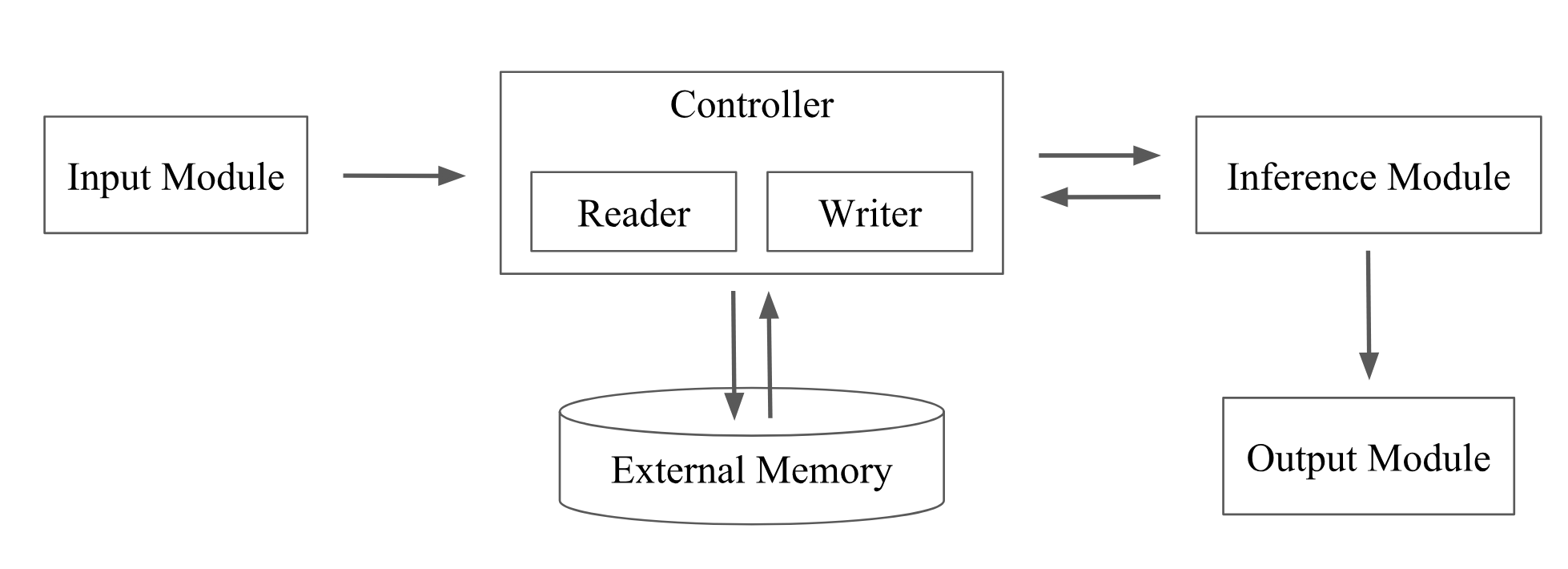} 
\caption{Block diagram of general memory-augmented neural networks.}
\label{FIG:MANN}
\end{figure}

\subsection{Overview}
Although the recurrent approaches using LSTM or GRU have been successful in most cases, they may suffer from two main problems: 
1) They struggle to effectively incorporate external KB information into the RNN hidden states~\cite{sukhbaatar2015end}, and they are known to be unstable over long sequences.
2) Processing long sequences one-by-one is very time-consuming, especially when using the attention mechanism.
Therefore, in this section, we will introduce the general concept of memory-augmented neural networks (MANNs) ~\cite{graves2014neural,weston2014memory,sukhbaatar2015end,kumar2016ask,graves2016hybrid,wang-EtAl:2016:EMNLP20161,vaswani2017attention,DBLP:journals/corr/KaiserNRB17}.

In Figure~\ref{FIG:MANN}, we show the general block diagram of MANNs. The key difference compared to standard neural networks is that a MANN usually has an external memory that a controller can interact with. The memory can be bounded or unbounded, flat or hierarchical, read-only or read-write capable, and contains implicit or explicit information. The overall computation can be summarized as follows: The input module first receives the input and sends the encoded input to the controller. A controller reads relevant information from the memory, or does computation to store some information in the memory by writing. The controller sends the result of the computation to the inference module. Finally, the inference module does high-level computations and sends the results to the output module for general output. 

\subsection{End-to-End Memory Networks}

\begin{figure}[h]
\centering
\includegraphics[width=\linewidth]{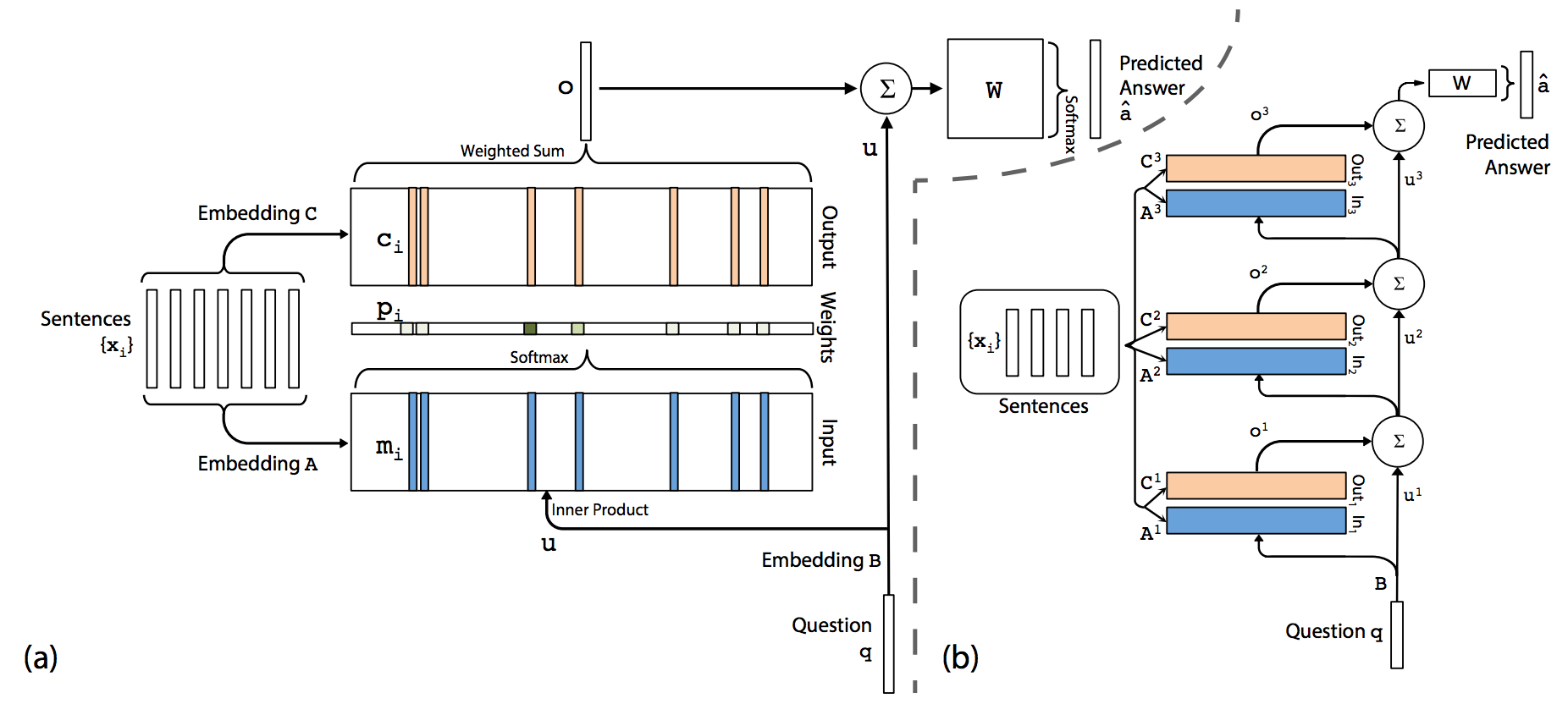} 
\caption{The architecture of end-to-end memory networks}
\label{FIG:MemNN}
\end{figure}

Most related to this thesis, we now introduce end-to-end memory networks (MNs) ~\cite{sukhbaatar2015end} in detail. In Figure~\ref{FIG:MemNN},~\footnote{The figure is from the original paper~\cite{sukhbaatar2015end}.} the left-hand side (a) shows how the model reads from and writes to the memory, and how the process can be repeated multiple times (``hops''), as shown in the right-hand side (b).

The memories of memory networks are represented by a set of trainable embedding matrices $E = \{$ $A^1, C^1$,$\dots$,$A^{K}, C^{K}\}$, where each $A^k$ or $C^k$ maps tokens to vectors and $K$ is the number of hops. 
There are two common ways of weight tying within the model, adjacent and layer-wise.
In adjacent weight tying, the output embedding for one layer is the input embedding for the one above, i.e., $A^{k+1} = C^{k}$; meanwhile, in layer-wise weight tying, $A^{k} = A^{k+1}$ and $C^{k} = C^{k+1}$, so it is more RNN-like.
In the remainder of this section, we use adjacent weight tying as the setting because it empirically outperforms the other setting.

A query vector $q^k$ is used as a reading head. The model loops over $K$ hops and it first computes the attention weights at hop $k$ for each memory $i$ using:
\begin{equation}
  p^k_i = \text{Softmax}((q^k)^T C^k_i),
  \label{Eq:mnattention}
\end{equation}
where $C^k_i$ is the memory content in position $i$ that is represented from the embedding matrix $C^k$. Here, $p^k$ is a soft memory selector that decides the memory relevance with respect to the query vector $q^k$. 
The model reads out the memory $o^k$ by the weighted-sum over $C^{k+1}$ (using $C^{k+1}$ from adjacent weighted tying),
\begin{equation}
  o^k = \sum_i p^k_i C^{k+1}_i.
  \label{eq3}
\end{equation}
Then the query vector is updated for the next hop using 
\begin{equation}
  q^{k+1} = q^{k} + o^{k}.
  \label{Eq:qk+1}
\end{equation}

Another essential part is what to allocate in the memory. In \cite{sukhbaatar2015end} and \cite{bordes2016learning}, each memory slot is represented as a sentence either from the predefined facts or utterances in dialogues, and the word embeddings in the same sentence are summed to be one single embedding for the memory slot. In~\cite{bordes2016learning}, to let the model recognize the speaker information, the authors add the speaker embeddings to the corresponding memory slots, in order to distinguish which utterances are from a user and which are from a system.

\begin{figure}[h]
\centering
\includegraphics[width=0.75\linewidth]{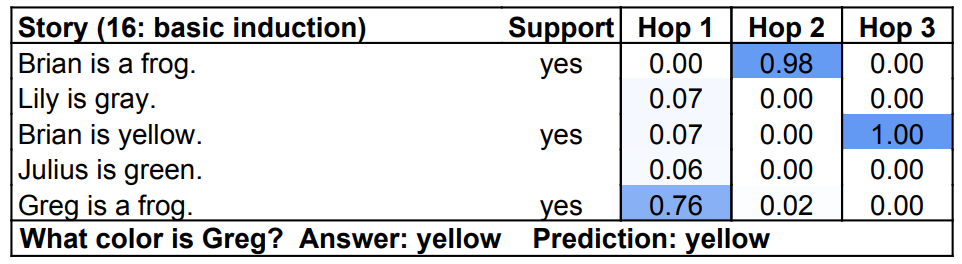}
\caption{Example prediction on the simulated question-answering tasks.}
\label{FIG:QAMultiHop}
\end{figure}

This processing can be repeated several times, which is usually called multiple hops reasoning. It has been empirically proven that multiple hops are useful in several question-answering tasks.
For example, in the multi-hop prediction example in Figure~\ref{FIG:QAMultiHop} using the bAbI dataset~\cite{weston2015towards}, there are five sentences represented as memory, and the query vector (in this case a question) is ``\textit{What color is Greg?}''
As shown in the attention weights, in the first hop, the model focuses on the memory slot of ``\textit{Greg is a frog.}'' After memory readout from hop one, the model pays attention to ``\textit{Brian is a frog.}'' In the end, the model is able to predict that Greg's color is yellow because of the attention on ``\textit{Brian is yellow},'' at the third hop.

%% file: chapter3.tex
\chapter{Copy-Augmented Dialogue State Tracking}
In this chapter, we focus on improving the core of pipeline dialogue systems, i.e., dialogue state tracking.
To effectively track the states, the model needs to memorize long dialogue context and be able to detect whether there is any slot is triggered, also what are its corresponding values.
Traditionally, state tracking approaches based on the assumption that ontology is defined in advance, where all slots and their values are known.
Having a predefined ontology can simplify DST into a classification problem and improve performance. 
However, there are two major drawbacks to this approach: 
1) A full ontology is hard to obtain in advance~\cite{P18-1134SpanPtr}.
In the industry, databases are usually accessed through an external API only, which is owned and maintained by others. It is not feasible to gain access to enumerate all the possible values for each slot.
2) Even if a full ontology exists, the number of possible slot values could be large and variable. 
For example, a restaurant name or a train departure time can contain a large number of possible values. 
Therefore, many of the previous work~\cite{williams2016dialog, NBT,MDBT,zhong2018global} that are based on neural classification models may not be applicable in a real scenario.

The copy mechanism, therefore, could be essential in dialogue state tracking via copying slot values from a dialogue history to extracted states. 
A dialogue state tracker with copy ability can detect unknown slot values in an ontology. 
Here, we propose a dialogue state tracker, the \textbf{tra}nsferable \textbf{d}ialogue stat\textbf{e} generator (TRADE), which is a novel end-to-end architecture without SLU module to perform state tracking based on generative models~\cite{wu2019TRADE}.
It includes an utterance encoder, a slot gate, and a state generator. It leverages its context-enhanced slot gate and copy mechanism to properly track slot values mentioned anywhere in a dialogue history.

\section{Model Description}
\begin{figure*}[t]
\centering
\includegraphics[width=\linewidth]{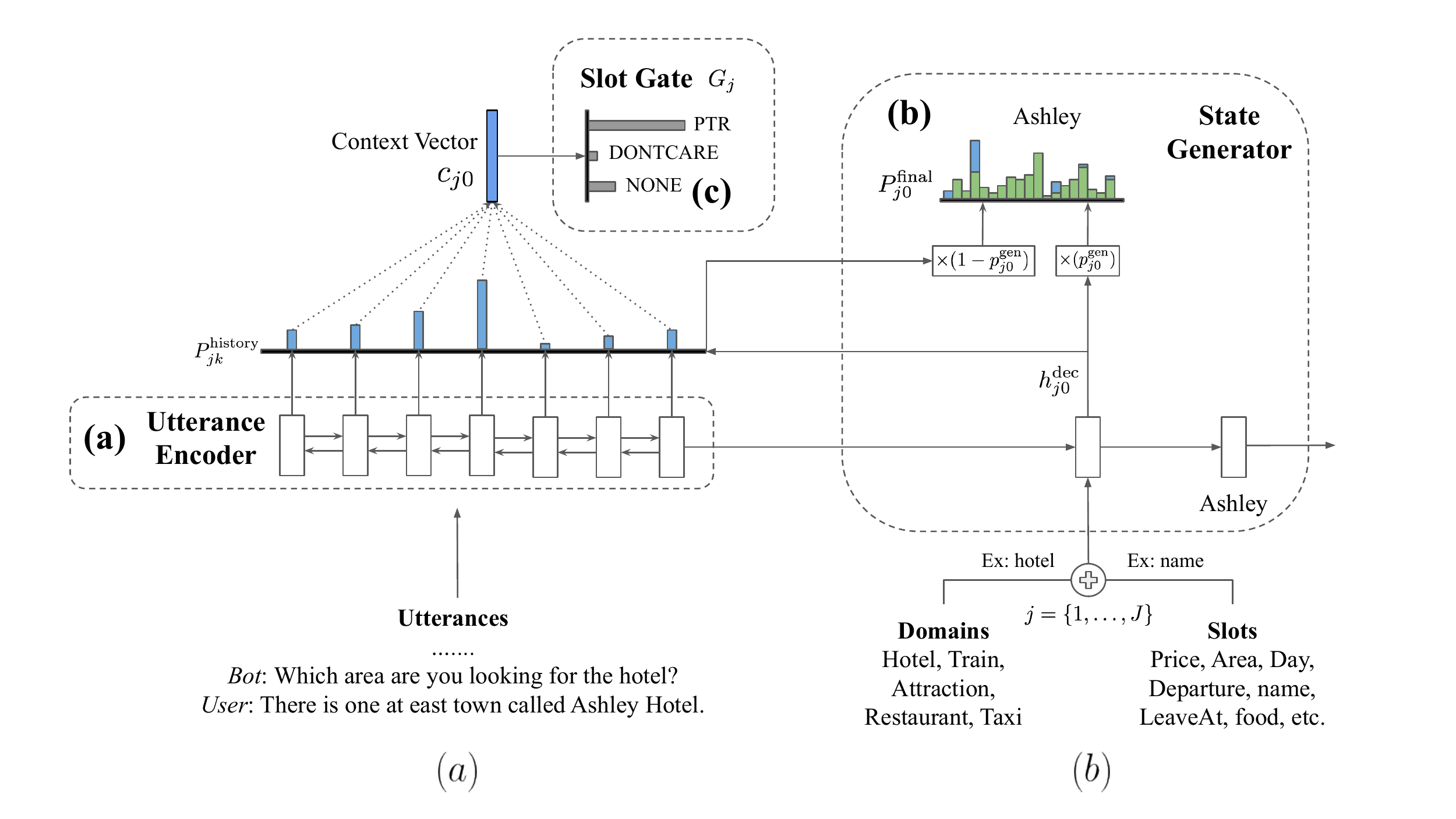}
\caption{The architecture of the proposed TRADE model, which includes (a) an utterance encoder, (b) a state generator, and (c) a slot gate, all of which are shared among domains.}
\label{FIG:TRADE_model}
\end{figure*}

The proposed TRADE model in Fig.~\ref{FIG:TRADE_model} comprises three components: an utterance encoder, a slot gate, and a state generator. 
Instead of predicting the probability of every predefined ontology term, this model directly generates slot values using the sequence decoding strategy.
Similar to~\cite{Q17-1024} for multilingual neural machine translation, we share all the model parameters, and the state generator starts with a different start-of-sentence token for each \textit{(domain, slot)} pair.

\subsection{Architecture}

The (a) utterance encoder encodes dialogue utterances into a sequence of fixed-length vectors.
The (b) state generator decodes multiple output tokens for all \textit{(domain, slot)} pairs independently to predict their corresponding values.
To determine whether any of the \textit{(domain, slot)} pairs are mentioned, the context-enhanced (c) slot gate is used with the state generator. 
The context-enhanced slot gate predicts whether each of the pairs is actually triggered by the dialogue via a three-way classifier. 
Assuming that there are $J$ possible \textit{(domain, slot)} pairs in the setting.

\subsubsection{(a) Utterance Encoder}
Note that the utterance encoder can be any existing encoding model.
We use s bi-directional GRU to encode the dialogue history.
The input to the utterance encoder is the concatenation of all words in the dialogue history, and the model infers the states across a sequence of turns. 
We use the dialogue history as the input of the utterance encoder, rather than the current utterance only. 

\subsubsection{(b) State Generator}
To generate slot values using text from the input source, a copy mechanism is required.
We employ soft-gated pointer-generator copying to combine a distribution over the vocabulary and distribution over the dialogue history into a single output distribution.
We use a GRU as the decoder of the state generator to predict the value for each \textit{(domain, slot)} pair, as shown in Fig.~\ref{FIG:TRADE_model}.
The state generator decodes $J$ pairs independently. 
We simply supply the summed embedding of the domain and slot as the first input to the decoder.

At decoding step $k$ for the $j$-th \textit{(domain, slot)} pair, the generator GRU takes a word embedding $w_{jk}$ as its input and returns a hidden state $h^{\text{dec}}_{jk}$. 
The state generator first maps the hidden state $h^{\text{dec}}_{jk}$ into the vocabulary space $P^{\text{vocab}}_{jk}$ using the trainable embedding $E \in \mathbb{R}^{|V| \times d_{hdd}}$, where $|V|$ is the vocabulary size and $d_{hdd}$ is the hidden size. 
At the same time, the $h^{\text{dec}}_{jk}$ is used to compute the history attention $P^{\text{history}}_{jk}$ over the encoded dialogue history $H_{t}$: 
\begin{gather}
        P^{\text{vocab}}_{jk}= \text{Softmax}(E  (h^{\text{dec}}_{jk})^\top) \in \mathbb{R}^{|V|}, \\
        P^{\text{\text{history}}}_{jk} = \text{Softmax}(H_{t} (h^{\text{dec}}_{jk})^\top) \in \mathbb{R}^{|X_t|}.
\label{prob}
\end{gather}
The final output distribution $P^{\text{final}}_{jk}$ is the weighted-sum of two distributions,
\begin{gather}
        P^{\text{final}}_{jk} 
        = p^{\text{gen}}_{jk} \times P^{\text{vocab}}_{jk} 
        +  (1-p^{\text{gen}}_{jk}) \times P^{\text{\text{history}}}_{jk} \in \mathbb{R}^{|V|}.
\end{gather}
The scalar $p^{\text{gen}}_{jk}$ is trainable to combine the two distributions, which is computed by
\begin{gather}
    p^{\text{gen}}_{jk} = \text{Sigmoid}(W_1  [h^{\text{dec}}_{jk}; w_{jk}; c_{jk}]) \in \mathbb{R}^{1}, \\
    c_{jk} = P^{\text{\text{history}}}_{jk} H_{t} \in \mathbb{R}^{d_{hdd}}.
\label{p_gen}
\end{gather}
The soft-gate copy mechanism is the same as Eq.~(\ref{eq:soft-copy}), but here we repeat it $J$ times to get different distributions for every \textit{(domain, slot)} pair.

\subsubsection{(c) Slot Gate}
The context-enhanced slot gate $G$ is a simple three-way classifier that maps a context vector taken from the encoder hidden states $H_{t}$ to a probability distribution over \textit{ptr}, \textit{none}, and \textit{dontcare} classes.
For each \textit{(domain, slot)} pair, if the slot gate predicts \textit{none} or \textit{dontcare}, we ignore the values generated by the decoder and fill the pair as ``not-mentioned'' or ``does not care''.
Otherwise, we take the generated words from our state generator as its value.
With a linear layer parameterized by $W_g\in \mathbb{R}^{3 \times d_{hdd}}$, the slot gate for the $j$-th \textit{(domain, slot)} pair is defined as 
\begin{equation}
    G_j = \text{Softmax}(W_g\cdot  (c_{j0})^\top) \in \mathbb{R}^{3},
\label{gate}
\end{equation}
where $c_{j0}$ is the context vector computed in Eq~(\ref{p_gen}) using the first decoder hidden state.

\subsection{Optimization}
During training, we optimize for both the slot gate and the state generator. 
For the former, the cross-entropy loss $L_{g}$ is computed between the predicted slot gate $G_{j}$ and the true one-hot label $y^{\text{gate}}_j$, 
\begin{equation}
    L_g = \sum_{j=1}^{J} - \log(G_j\cdot  (y^{\text{gate}}_j)^\top).
\end{equation}
For the latter, another cross-entropy loss $L_v$ between $P^{\text{final}}_{jk}$ and the true words $Y_{j}^{\text{label}}$ is used. We define $L_v$ as 
\begin{equation}
L_v = \sum_{j=1}^{J} \sum_{k=1}^{|Y_j|} - \log(P^{\text{final}}_{jk}\cdot (y^{\text{value}}_{jk})^\top).
\end{equation}
$L_v$ is the sum of losses from all the \textit{(domain, slot)} pairs and their decoding time steps. 
We optimize the weighted-sum of these two loss functions using hyper-parameters $\alpha$ and $\beta$,
\begin{equation}
\label{eq:loss}
L = \alpha L_g + \beta L_v . 
\end{equation}

\section{Multiple Domain DST}
\begin{figure}[t]
\centering
\includegraphics[width=0.8\linewidth]{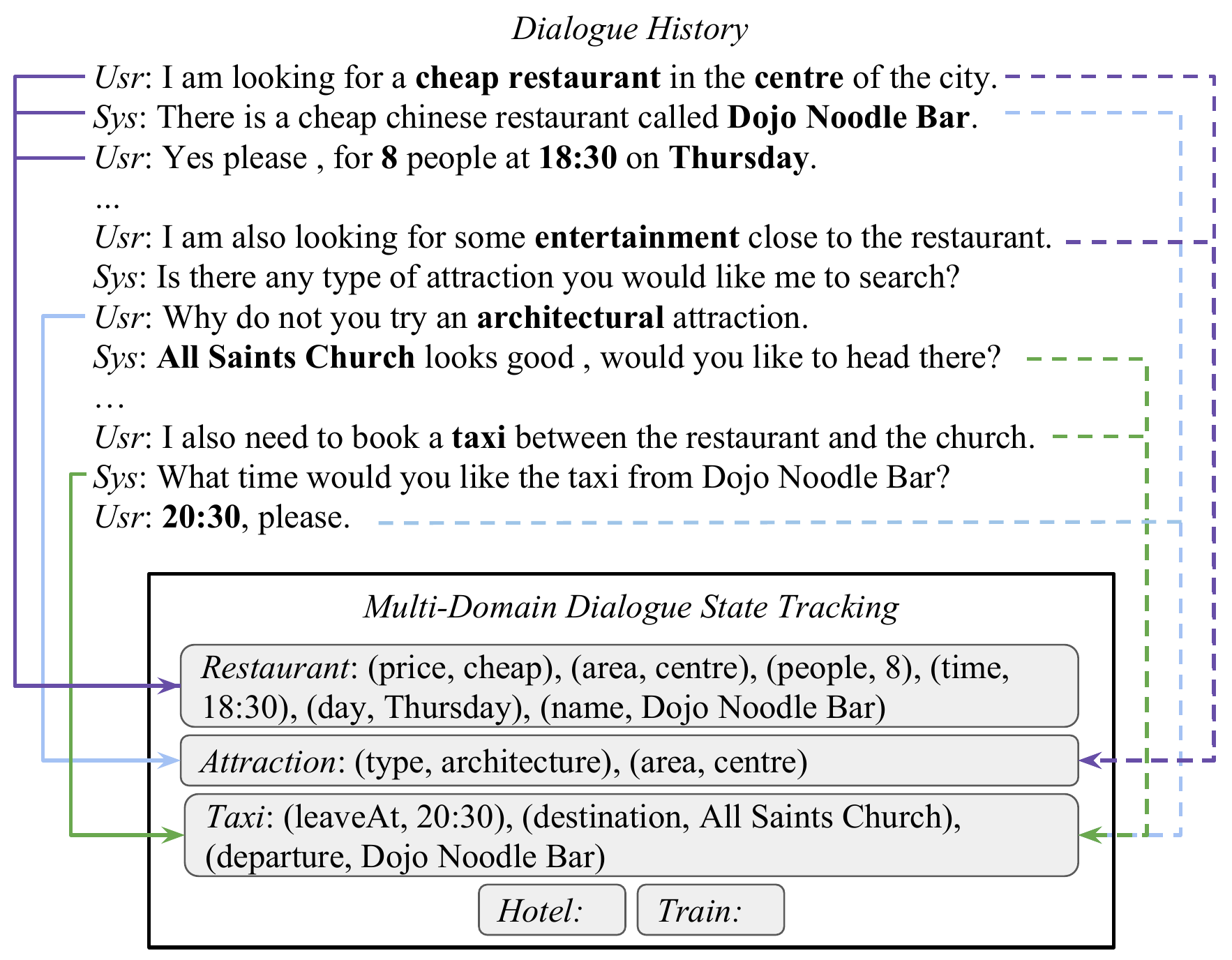}
\caption{An example of multi-domain dialogue state tracking in a conversation. The solid arrows on the left are the single-turn mapping, and the dot arrows on the right are multi-turn mapping. The state tracker needs to track slot values mentioned by the user for all the slots in all the domains.}
\label{FIG:multiwoz_example}
\end{figure}

In a single-task multi-domain dialogue setting, as shown in Fig.~\ref{FIG:multiwoz_example}, a user can start a conversation by asking to reserve a restaurant, then request information regarding an attraction nearby, and finally ask to book a taxi. 
In this case, the DST model has to determine the corresponding domain, slot, and value at each turn of dialogue, which contains a large number of combinations in the ontology.
For example, single-domain DST problems usually have only a few slots that need to be tracked, four slots in WOZ~\cite{wen2016network} and eight slots in DSTC2~\cite{henderson2014dstc2}, but there are 30 \textit{(domain, slot)} pairs and over 4,500 possible slot values in MultiWOZ~\cite{multiwoz}, a multi-domain dialogue dataset.
Another challenge in the multi-domain setting comes from the need to perform multi-turn mapping.
Single-turn mapping refers to the scenario where the \textit{(domain, slot, value)} triplet can be inferred from a single turn (the solid line in the figure), while in multi-turn mapping, it may need to be inferred from multiple turns which happen in different domains (the dotted line in the figure).
For instance, the \textit{(area, centre)} pair from the \textit{attraction} domain in Fig.~\ref{FIG:multiwoz_example} can be predicted from the \textit{area} information in the \textit{restaurant} domain, which is mentioned in the preceding turns.

To tackle these challenges, we emphasize that DST models should share tracking knowledge across domains. There are many slots among different domains that share all or some of their values.
For example, the \textit{area} slot can exist in many domains, e.g., \textit{restaurant}, \textit{attraction}, and \textit{taxi}. Moreover, the~\textit{name} slot in the \textit{restaurant} domain can share the same value with the \textit{departure} slot in the \textit{taxi} domain.
Additionally, to enable the DST model to track slots in unseen domains, transferring knowledge across multiple domains is imperative. 
We expect DST models can learn to track some slots in zero-shot domains by learning to track the same slots in other domains. For example, if the model learns how to track the ``departure'' slot in the \textit{bus} domain, then it could transfer the knowledge to track the same slot in \textit{taxi} domain.

\subsection{Experimental Setup}
\subsubsection{Dataset}
Multi-domain Wizard-of-Oz (MultiWOZ) is the largest existing human-human conversational corpus spanning over seven domains, containing 8438 multi-turn dialogues, with each dialogue averaging 13.68 turns.
Different from existing standard datasets like WOZ~\cite{wen2016network} and DSTC2~\cite{henderson2014dstc2}, which contain less than 10 slots and only a few hundred values, MultiWOZ has 30 \textit{(domain, slot)} pairs and over 4,500 possible values.
We use the DST labels from the original training, validation and testing dataset.
Only five domains (\textit{restaurant}, \textit{hotel}, \textit{attraction}, \textit{taxi}, \textit{train}) are used in our experiment because the other two domains (\textit{hospital}, \textit{police}) have very few dialogues (10\% compared to others) and only appear in the training set.
The slots in each domain and the corresponding data size are reported in Table~\ref{MultiWOZ-DATASET-TABLE}.

\begin{table}[t!]
\begin{center}
\caption{The dataset information of MultiWOZ on five different domains: hotel, train, attraction, restaurant, and taxi.}
\label{MultiWOZ-DATASET-TABLE}
\begin{tabular}{r|c|c|c|c|c}
\hline
 & \textbf{Hotel} & \textbf{Train} & \textbf{Attraction} & \textbf{Restaurant} & \textbf{Taxi} \\ \hline
\textit{Slots} & \begin{tabular}[c]{@{}c@{}}price,\\ type,\\ parking,\\ stay,\\ day,\\ people,\\ area,\\ stars,\\ internet,\\ name\end{tabular} & \begin{tabular}[c]{@{}c@{}}destination,\\ departure,\\ day,\\ arrive by,\\ leave at,\\ people\end{tabular} & \begin{tabular}[c]{@{}c@{}}area,\\ name,\\ type\end{tabular} & \begin{tabular}[c]{@{}c@{}}food,\\ price,\\ area,\\ name,\\ time,\\ day,\\ people\end{tabular} & \begin{tabular}[c]{@{}c@{}}destination,\\ departure,\\ arrive by,\\ leave by\end{tabular} \\ \hline
\textit{Train} & 3381 & 3103 & 2717 & 3813 & 1654 \\
\textit{Valid} & 416 & 484 & 401 & 438 & 207 \\
\textit{Test} & 394 & 494 & 395 & 437 & 195 \\ \hline
\end{tabular}
\end{center}
\end{table}

\subsubsection{Training}
The model is trained end-to-end using the Adam optimizer ~\citep{kingma2014adam} with a batch size of 32. 
The learning rate annealing is in the range of $[0.001, 0.0001]$ 
with a dropout ratio of 0.2. Both $\alpha$ and $\beta$ in Eq~(\ref{eq:loss}) are set to one. 
All the embeddings are initialized by concatenating Glove embeddings~\cite{pennington2014glove} and character embeddings~\cite{hashimoto2016joint}, where the dimension is 400 for each vocabulary word.
A greedy search decoding strategy is used for our state generator since the generated slot values are usually short in length and contain simple grammar.
In addition, to increase model generalization and simulate an out-of-vocabulary setting, a word dropout is utilized with the utterance encoder by randomly masking a small number of input tokens.

\subsubsection{Evaluation Metrics}
Two evaluation metrics, joint goal accuracy and slot accuracy, are used to evaluate the performance on multi-domain DST. 
The joint goal accuracy compares the predicted dialogue states to the ground truth at each dialogue turn, and the output is considered correct if and only if all the predicted values exactly match the ground truth values.
The slot accuracy, on the other hand, individually compares each (domain, slot, value) triplet to its ground truth label. 

\subsection{Baseline Models}
We make a comparison with the following existing models: MDBT~\cite{MDBT}, GLAD~\cite{zhong2018global}, GCE~\cite{Nouri2018TowardSN}, and SpanPtr~\cite{P18-1134SpanPtr}, and we briefly describe these baseline models below:
\begin{itemize}
  \item MDBT: A model that jointly identifies the domain and tracks the belief states corresponding to that domain is proposed. Multiple bi-LSTMs are used to encode system and user utterances. The semantic similarity between utterances and every predefined ontology term is computed separately. Each ontology term is triggered if the predicted score is greater than a threshold. The model was tested on multi-domain dialogues.

  \item GLAD: 
  This model uses self-attentive RNNs to learn a global tracker that shares data among slots and a local tracker that tracks each slot. The model takes previous system actions and the current user utterance as input and computes semantic similarity with predefined ontology terms. The authors show that it can generalize on rare slot-value pairs with few training examples. The model was tested on single-domain dialogues. 
  
  \item GCE: 
  This is the current state-of-the-art model on the single-domain WOZ dataset~\cite{wen2016network} and DSTC2~\cite{henderson2014dstc2}; it is a simplified and sped up version of GLAD, without the slot-specific RNNs that compute for each utterance. It reduces the latency in training and inference times by 35\% on average, while preserving performance of state tracking. The model was tested on single-domain dialogues.
  
  \item SpanPtr: This is the first model that applies pointer networks ~\cite{Vinyals2015PointerN} to the DST problem, which generates both start and end pointers to perform index-based copying. The authors claim that their model based on the pointer network can effectively extract unknown slot values while still obtains promising accuracy on the DSTC2~\cite{henderson2014dstc2} benchmark. The model was tested on single-domain dialogues.
\end{itemize}

\begin{table}[t!]
\begin{center}
\caption{The multi-domain DST evaluation on MultiWOZ and its single \textit{restaurant} domain.}
\label{JOINT-TABLE}
\begin{tabular}{r|cc|cc}
\hline
\multicolumn{1}{l|}{} & \multicolumn{2}{c|}{\textbf{MultiWOZ}} & \multicolumn{2}{c}{\textbf{\begin{tabular}[c]{@{}c@{}}MultiWOZ\\ (Only Restaurant)\end{tabular}}} \\ \cline{2-5} 
 & \textit{\textbf{Joint}} & \textit{\textbf{Slot}} & \textit{\textbf{Joint}} & \textit{\textbf{Slot}} \\ \hline
\textit{MDBT} & 15.57 & 89.53 & 17.98 & 54.99 \\ \hline
\textit{GLAD} & 35.57 & 95.44 & 53.23 & 96.54 \\ \hline
\textit{GCE} & 36.27 & 98.42 & 60.93 & 95.85 \\ \hline
\textit{SpanPtr} & 30.28 & 93.85 & 49.12 & 87.89 \\ \hline
\textit{TRADE} & \textbf{48.62} & 96.92 & \textbf{65.35} & 93.28 \\
\hline
\end{tabular}
\end{center}
\end{table}

\subsection{Results and Discussion}

As shown in Table~\ref{JOINT-TABLE}, TRADE achieves the highest performance, 48.62\% on joint goal accuracy and 96.92\% on slot accuracy, on multiWOZ.
For comparison with the performance on a single domain, the results on the \textit{restaurant} domain of MultiWOZ are reported as well.
The performance difference between SpanPtr and our model mainly comes from the limitation of index-based copying.
For examples, if the true label for the price range slot is \textit{cheap}, the relevant user utterance describing the restaurant may actually be, for example, \textit{economical}, \textit{inexpensive}, or \textit{cheaply}. 
Note that the MDBT, GLAD, and GCE models each need a predefined domain ontology to perform binary classification for each ontology term, which hinders their DST tracking performance. 


\begin{figure}[t]
\begin{subfigure}{0.5\textwidth}
  \centering
  \includegraphics[width=0.9\linewidth]{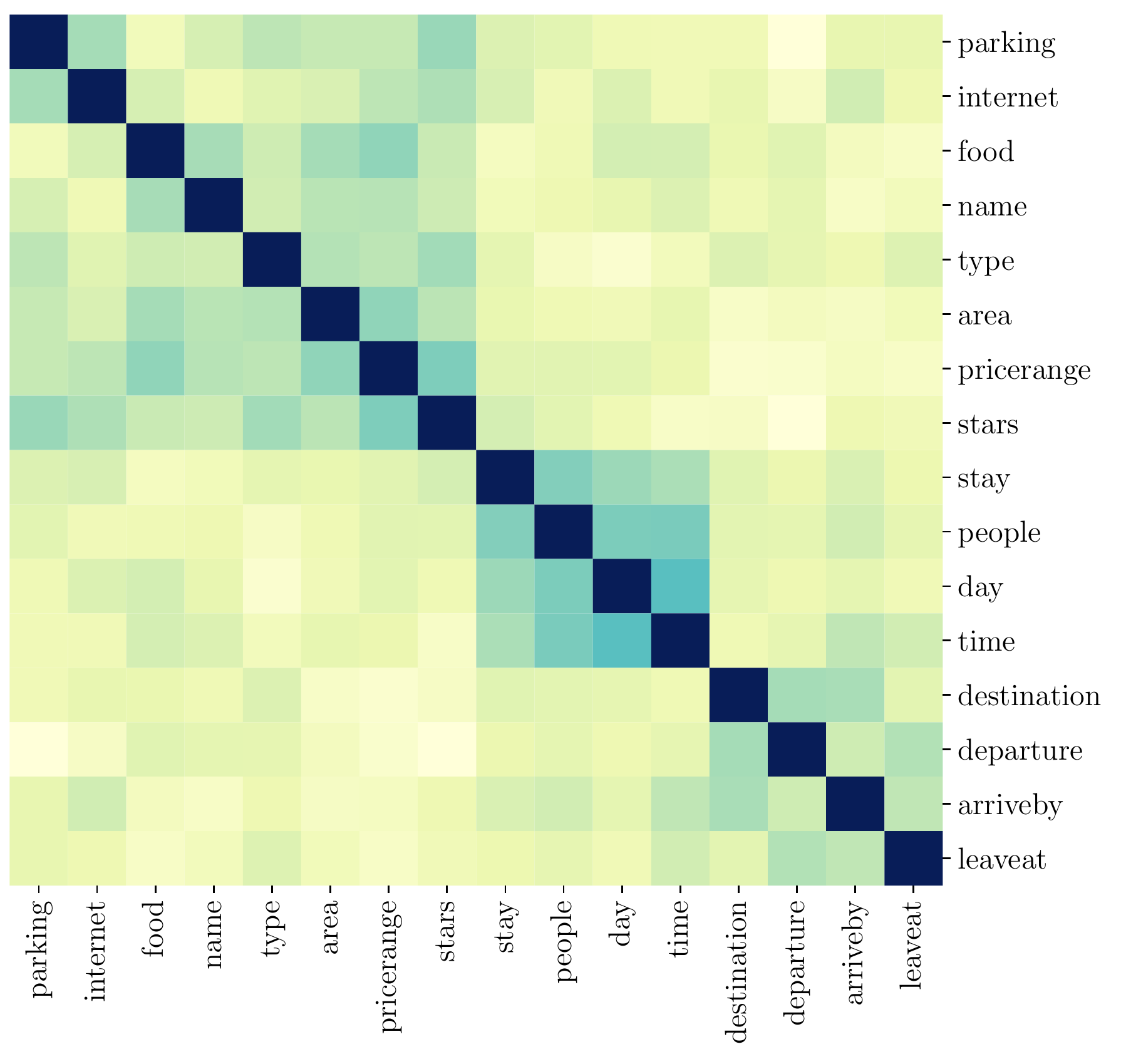}
  \caption{}
\end{subfigure}
\begin{subfigure}{0.5\textwidth}
  \centering
  \includegraphics[width=0.9\linewidth]{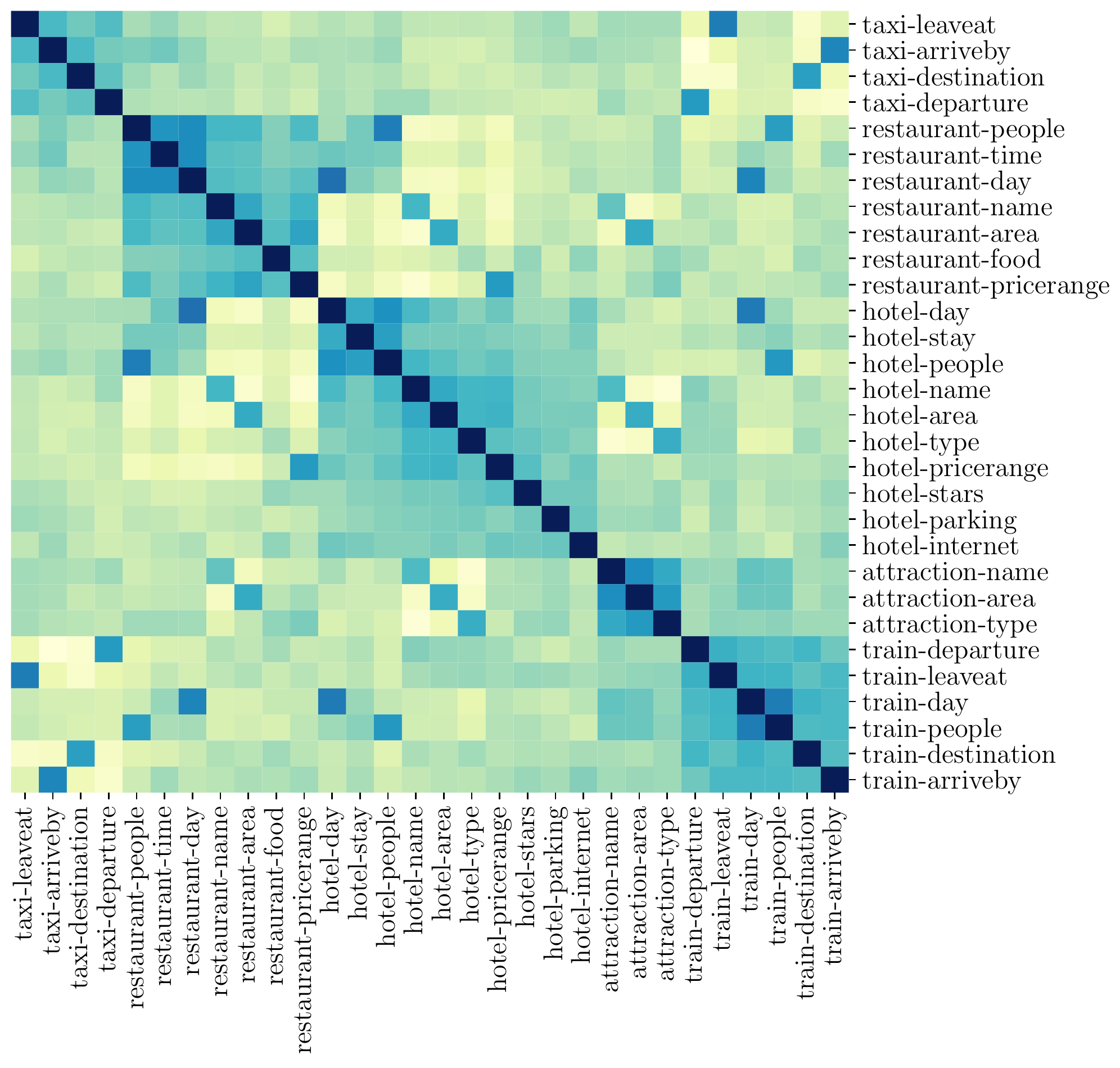}
  \caption{}
\end{subfigure}%
\caption{The embeddings cosine similarity visualization for (a) slots and (b) \textit{(domain, slot)} pairs. }
\label{FIG:slot-viz}
\end{figure}

We visualize the cosine similarity matrix for all the possible slot embeddings in Figure~\ref{FIG:slot-viz} (a). 
Most of the slot embeddings are not close to each other, which is expected because the model only depends on these features as start-of-sentence embeddings to distinguish different slots.
Note that some slots are relatively close because either the values they track may share similar semantic meanings or the slots are correlated. 
For example,~\textit{destination} and~\textit{departure} track names of cities, while ~\textit{people} and~\textit{stay} track numbers. 
On the other hand,~\textit{price range} and \textit{star} in the hotel domain are correlated because high-star hotels are usually expensive.
The visualization of all the possible \textit{(domain, slot)} embeddings is shown in Figure~\ref{FIG:slot-viz}(b).

\section{Unseen Domain DST}
In this section, we focus on the ability of TRADE to generalize to an unseen domain by considering zero-shot transferring and few-shot domain expansion.
In the zero-shot setting, we assume we have no training data in the new domain, while in the few-shot case, we assume just 1\% of the original training data in the unseen domain is available (around 20 to 30 dialogues). 
One of the motivations to perform unseen domain DST is because collecting a large-scale task-oriented dataset for a new domain is expensive and time-consuming ~\cite{multiwoz}, and there are a large number of domains in realistic scenarios.

\subsection{Zero-shot DST}
Zero-shot applications in dialogue learning have been used for intention classifiers~\cite{chen2016zero}, slot-filling~\cite{bapna2017towards}, dialogue policy~\cite{gavsic2014gaussian}, and language generation~\cite{Q17-1024,zhao2018zero}.
In the DST task, ideally based on the slots already learned, a model is able to directly track those slots that are present in a new domain. 
For example, if the model is able to track the \textit{departure} slot in the \textit{train} domain, then that ability may transfer to the \textit{taxi} domain, which uses similar slots.
Note that generative DST models take the dialogue context/history $X$, the domain $D$, and the slot $S$ as input and then generate the corresponding values $Y^{\text{value}}$.
Let $(X, D_{\text{source}}, S_{\text{source}},Y^{\text{value}}_{\text{source}})$ be the set of samples seen during the training phase and $(X, D_{\text{\text{target}}}, S_{\text{target}},Y^{\text{value}}_{\text{target}})$ the samples which the model was not trained to track. 
A zero-shot DST model should be able to generate the correct values of $Y^{\text{value}}_{\text{target}}$ given the context $X$, domain $D_{\text{target}}$, and slot $S_{\text{target}}$, without using any training samples. 
The same context $X$ may appear in both source and target domains but the pairs ($D_{\text{target}}, S_{\text{target}}$) are unseen. 
This setting is extremely challenging if no slot in $S_{\text{target}}$ appears in $S_{\text{source}}$, since the model has never been trained to track such a slot.  

\subsubsection{Results and Discussion}

\begin{table}[h]
\begin{center}
\caption{Zero-shot experiments on an unseen domain. We held-out one domain each time to simulate the setting.}
\label{ZeroShot-TABLE}
\begin{tabular}{r|cc|cc}
\hline
\multicolumn{1}{l|}{\multirow{2}{*}{}} & \multicolumn{2}{c|}{\textbf{Trained Single}} & \multicolumn{2}{c}{\textbf{Zero-Shot}} \\ \cline{2-5} 
\multicolumn{1}{c|}{} & 
\multicolumn{1}{l}{\textit{Joint}} & 
\multicolumn{1}{c|}{\textit{Slot}} & 
\multicolumn{1}{c}{\textit{Joint}} & 
\multicolumn{1}{c}{\textit{Slot}} \\ \hline
\textit{Hotel} & 55.52 & 92.66 & 13.70 & 65.32 \\ \hline
\textit{Train} & 77.71 & 95.30 & 22.37 & 49.31 \\ \hline
\textit{Attraction} & 71.64 & 88.97 & 19.87 & 55.53 \\ \hline
\textit{Restaurant} & 65.35 & 93.28 & 11.52 & 53.43 \\ \hline
\textit{Taxi} & 76.13 & 89.53 & \textbf{60.58} & 73.92 \\ \hline
\end{tabular}
\end{center}
\end{table}

We run zero-shot experiments by excluding one domain from the training set.
As shown in Table~\ref{ZeroShot-TABLE}, the \textit{taxi} domain achieves the highest zero-shot performance, 60.58\% on joint goal accuracy, which is close to the result achieved by training on all the \textit{taxi} domain data (76.13\%).
Although performances on the other zero-shot domains are not especially promising, they still achieve around 50 -- 65\% slot accuracy without using any in-domain samples. 
The reason why the zero-shot performance on the \textit{taxi} domain is high is that all four slots share similar values with the corresponding slots in the~\textit{train} domain.

\subsection{Expanding DST for Few-shot Domain}
In this section, we assume that only a small number of samples from the new domain $(X, D_{\text{target}}$, $S_{\text{target}}$, $Y^{\text{value}}_{\text{target}})$ are available, and the purpose is 1) to evaluate the ability of our DST model to transfer its learned knowledge to the new domain without forgetting previously learned domains; and 2) to expand the DST model to an unseen domain without expensive cost.
There are two advantages to performing few-shot domain expansion: 
1) being able to quickly adapt to new domains and obtain decent performance with only a small amount of training data;
2) not requiring retraining with all the data from previously learned domains, since the data may no longer be available and retraining is often very time-consuming.

Firstly, we consider a straightforward naive baseline, i.e., fine-tuning on the target domain with no constraints. Then, we employ two specific continual learning techniques: elastic weight consolidation (EWC)~\cite{kirkpatrick2017overcoming} and gradient episodic memory (GEM)~\cite{lopez2017gradient} to fine-tune our model.
We define $\Theta_{S}$ as the model's parameters trained in the source domain, and $\Theta$ indicates the current optimized parameters according to the target domain data. 

EWC uses the diagonal of the Fisher information matrix $F$ as a regularizer for adapting to the target domain data. This matrix is approximated using samples from the source domain.
The EWC loss is defined as
\begin{equation}
    L_{ewc}(\Theta) = L(\Theta) + \sum_i \frac{\lambda}{2} F_i (\Theta_i - \Theta_{S,i})^2,
\end{equation}
where $\lambda$ is a hyper-parameter.
Different from EWC, GEM keeps a small number of samples $K$ from the source domains, and, while the model learns the new target domain, a constraint is applied on the gradient to prevent the loss on the stored samples from increasing. 
The training process is defined as: 
\begin{align}
\begin{split}
    &\text{Minimize}_{\Theta} \ L(\Theta) \\
    & \text{Subject to} \  L(\Theta, K) \leq L(\Theta_S, K),
\label{contrain}
\end{split}
\end{align}
where $L(\Theta, K)$ is the loss value of the $K$ stored samples. 
\cite{lopez2017gradient} show how to solve the optimization problem in Eq (\ref{contrain}) with quadratic programming if the loss of the stored samples increases.

\subsubsection{Experimental Setup}
We follow the same procedure as in the joint training section, and we run a small grid search for all the methods using the validation set. For EWC, we set different values of $\lambda$ for all the domains, and the optimal value is selected using the validation set. Finally, in GEM, we set the memory size $K$ to 1\% of the source domains.

\subsubsection{Results and Discussion}
In this setting, the TRADE model is pre-trained on four domains and a \textit{withheld} domain is reserved for domain expansion to perform fine-tuning. 
After fine-tuning on the new domain, we evaluate the performance of TRADE on 1) the four pre-trained domains, and 2) the new domain. We experiment with different fine-tuning strategies.
In Table~\ref{1_alldomain}, the first row is the base model that is trained on the four domains. The second row is the results on the four domains after fine-tuning on 1\% new domain data using three different strategies. One can find that GEM outperforms naive and EWC fine-tuning in terms of catastrophic forgetting on the four domains. Then we evaluate the results on a new domain for two cases: training from scratch and fine-tuning from the base model. Results show that fine-tuning from the base model usually achieves better results on the new domain compared to training from scratch.
In general, GEM outperforms naive and EWC fine-tuning by far in terms of overcoming catastrophic forgetting. We also find that pre-training followed by fine-tuning outperforms training from scratch on the single domain.

\begin{table*}[t]
\resizebox{\linewidth}{!}{
\begin{tabular}{cr|cc|cc|cc|cc|cc}
\hline
\multicolumn{1}{l}{} &  & \textbf{Joint} & \textbf{Slot} & \textbf{Joint} & \textbf{Slot} & \textbf{Joint} & \textbf{Slot} & \textbf{Joint} & \textbf{Slot} & \textbf{Joint} & \textbf{Slot} \\
\multicolumn{2}{c|}{\textbf{Evaluation on 4 Domains}} & \multicolumn{2}{l|}{\textit{Except Hotel}} & \multicolumn{2}{l|}{\textit{Except Train}} & \multicolumn{2}{l|}{\textit{Except Attraction}} & \multicolumn{2}{l|}{\textit{Except Restaurant}} & \multicolumn{2}{l}{\textit{Except Taxi}} \\ \hline
\multicolumn{2}{c|}{\begin{tabular}[c]{@{}c@{}}Base Model (BM)\\ training on 4 domains\end{tabular}} & 58.98 & 96.75 & 55.26 & 96.76 & 55.02 & 97.03 & 54.69 & 96.64 & 49.87 & 96.77 \\ \hline
\multicolumn{1}{c|}{\multirow{3}{*}{\begin{tabular}[c]{@{}c@{}}Fine-tuning BM\\ on 1\% new domain\end{tabular}}} & \textit{Naive} & 36.08 & 93.48 & 23.25 & 90.32 & 40.05 & 95.54 & 32.85 & 91.69 & 46.10 & 96.34 \\
\multicolumn{1}{c|}{} & \textit{EWC} & 40.82 & 94.16 & 28.02 & 91.49 & 45.37 & 84.94 & 34.45 & 92.53 & \textbf{46.88} & 96.44 \\
\multicolumn{1}{c|}{} & \textit{GEM} & \textbf{53.54} & \textbf{96.27} & \textbf{50.69} & \textbf{96.42} & \textbf{50.51} & \textbf{96.66} & \textbf{45.91} & \textbf{95.58} & 46.43 & \textbf{96.45} \\ 
\hline \hline
\multicolumn{2}{c|}{\textbf{Evaluation on New Domain}} & \multicolumn{2}{c|}{\textit{Hotel}} & \multicolumn{2}{c|}{\textit{Train}} & \multicolumn{2}{c|}{\textit{Attraction}} & \multicolumn{2}{c|}{\textit{Restaurant}} & \multicolumn{2}{c}{\textit{Taxi}} \\ \hline
\multicolumn{2}{c|}{Training 1\% New Domain} & 19.53 & 77.33 & 44.24 & 85.66 & \textbf{35.88} & \textbf{68.60} & 32.72 & 82.39 & 60.38 & 72.82 \\ \hline
\multicolumn{1}{c|}{\multirow{3}{*}{\begin{tabular}[c]{@{}c@{}}Fine-tuning BM\\ on 1\% new domain\end{tabular}}} & \textit{Naive} & 19.13 & 75.22 & \textbf{59.83} & \textbf{90.63} & 29.39 & 60.73 & \textbf{42.42} & \textbf{86.82} & \textbf{63.81} & \textbf{79.81} \\
\multicolumn{1}{c|}{} & \textit{EWC} & 19.35 & 76.25 & 58.10 & 90.33 & 32.28 & 62.43 & 40.93 & 85.80 & 63.61 & 79.65 \\
\multicolumn{1}{c|}{} & \textit{GEM} & \textbf{19.73} & \textbf{77.92} & 54.31 & 89.55 & 34.73 & 64.37 & 39.24 & 86.05 & 63.16 & 79.27 \\ \hline
\end{tabular}
}
\caption{Domain expanding DST for different few-shot domains.}
\label{1_alldomain}
\end{table*}

Fine-tuning TRADE with GEM maintains higher performance on the original four domains.
Take the \textit{hotel} domain as an example, the performance on the four domains after fine-tuning with GEM only drops from 58.98\% to 53.54\% (-5.44\%) on joint accuracy, whereas naive fine-tuning deteriorates the tracking ability, dropping joint goal accuracy to 36.08\% (-22.9\%). Expanding TRADE from four domains to a new domain achieves better performance than training from scratch on the new domain. 
This observation underscores the advantages of transfer learning with the proposed TRADE model.
For example, our TRADE model achieves 59.83\% joint accuracy after fine-tuning using only 1\% of \textit{Train} domain data, outperforming training the\textit{Train} domain from scratch, which achieves 44.24\% using the same amount of new-domain data.

Finally, when considering \textit{hotel} and \textit{attraction} as a new domain, fine-tuning with GEM outperforms the naive fine-tuning approach on the new domain.
To elaborate, GEM obtains 34.73\% joint accuracy on the \textit{attraction} domain, but naive fine-tuning on that domain can only achieve 29.39\%.
This implies that in some cases learning to keep the tracking ability (learned parameters) of the learned domains helps to achieve better performance for the new domain.

\section{Short Summary}
We introduce a transferable dialogue state generator for multi-domain dialogue state tracking, which can better memorize the long dialogue context and track the states efficiently.
Our model learns to track states without any predefined domain ontology, which can handle unseen slot values using a copy mechanism.
TRADE shares all of its parameters across multiple domains and achieves state-of-the-art joint goal accuracy and slot accuracy on the MultiWOZ dataset for five different domains.
Moreover, domain sharing enables TRADE to perform zero-shot DST for unseen domains.
With the help of existing continual learning algorithms, our model can quickly adapt to few-shot domains without forgetting the learned ones.

%% file: chapter4.tex
\chapter{Retrieval-Based Memory-Augmented Dialogue Systems}

In the previous chapter, we discussed how we can memorize long dialogue context via copy mechanism, and how to leverage multiple domains to further improve state tracking performance and enable unseen domain DST.
In the remaining parts of this thesis, instead of solely optimizing the DST component, we view the whole dialogue system as a black box and train the system end-to-end. The inputs of the system are the long dialogue history/context and external knowledge base (KB) information, and the output is the system response for the next coming turn.

In this chapter, we first introduce one of the aspects of end-to-end dialogue learning, the retrieval-based dialogue systems.
Given the dialogue history and knowledge base information, machine learning models are required to predict/select the correct system response from a predefined response candidates. This task is usually suitable for small dataset training or for dialogue systems that required regular but not diverse system behavior. We propose a delexicalization strategy to simplify the retrieval problem, then we introduce two memory-augmented neural networks, a recurrent entity networks (REN)~\cite{henaff2016tracking} and dynamic query memory network (DQMN)~\cite{dqmem8461426}, for task-oriented dialogue learning. Lastly, we evaluate the models on simulated bAbI dialogue~\cite{bordes2016learning}, and also its more challenging OOV setting.


\section{Recorded Delexicalization Copying}
\begin{figure}[h]
\centering
\centerline{\includegraphics[width=\linewidth]{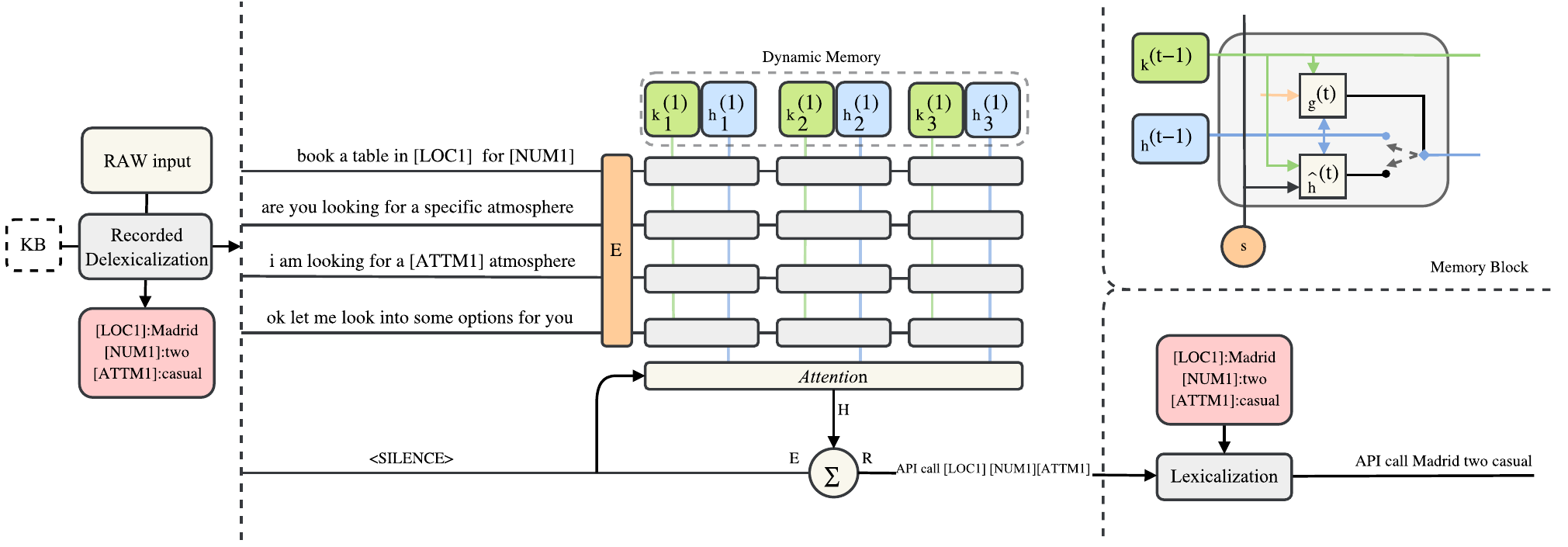}}
\caption{Entity-value independent recurrent entity network for goal-oriented dialogues. The graphic on the top right shows the detailed memory block.}
\label{fig:REN}
\end{figure}

There are a large amount of entities in the ontology, e.g., names of restaurants. It is hard for a retrieval-based model to distinguish the minor difference in the response candidates.
To simply overcome the weak entity identification problem, we propose a practical strategy, recorded delexicalization copying (RDC), to replace each real entity value with its entity type and the order appearance in the dialogue. 
We also build a lookup table to record the mapping, e.g., the first user utterance in Figure~\ref{fig:REN}, ``\textit{Book a table in Madrid for two,}" will be transformed into ``\textit{Book a table in [LOC-1] for [NUM-1].}" 
At the same time, \textit{[LOC-1]} and \textit{[NUM-1]} are stored in a lookup table as \textit{Madrid} and \textit{two}, respectively. 
Lexicalization is the reverse, copying the real entity values stored in the table to the output template. 
For example, when the output ``\textit{api-call [LOC-1] [NUM-1] [ATTM-1]}" is predicted, we will copy \textit{Madrid}, \textit{two} and \textit{casual} to fill in the blanks.
Last, we build the action template candidates by all the possible delexicalization system responses.
RDC is similar to delexicalization and the entity indexing strategy in~\cite{williams2017hybrid} and \cite{zhao2017generative}.
It not only decreases the learning complexity but also makes our system scalable to OOV settings.

\section{Model Description}

\subsection{Recurrent Entity Networks}

The REN model was first used in question answering tasks and has empirically shown its effectiveness \cite{henaff2016tracking, madotto2017question}. 
It is one kind of memory-augmented neural networks that is equipped with a dynamic long-term memory, which allows it to maintain and update a representation of the state of the world as it receives new data.
We first proposed to utilize REN in learning retrieval-based dialogue systems~\cite{wu2017end} in the 6th Dialogue System Technology Challenge (DSTC6)~\cite{perez2017dialog}. 
We treat every incoming utterance as the new received data, and store the dialogue history and external KB in the dynamic long-term memory to represent the state of the world.

REN has three main components: an input encoder, dynamic memory, and output module. 
The input encoder transforms the set of sentences $s_{t}$ and the question $q$ (we set the last user utterance as the question.) into vector representations by using multiplicative masks. We first look up word embeddings for each word in the sentences, and then apply the learned multiplicative masks, $f^{(s)}$ and $f^{(q)}$, to each word in a sentence. The final encoding vector of a sentence is defined as
\begin{equation}
s_{t} = \sum_{i} s_{t}^{i} \odot f_i^{(s)}, \quad \quad q= \sum_{i} q^{i} \odot f_i^{(q)}.
\end{equation} 

The dynamic memory stores long-term information, which is very similar to a GRU with a hidden state divided into blocks. The blocks ideally represent an entity type (e.g., \textit{LOC, PRICE,} etc.), and store relevant facts about it. Each block $i$ is made of a hidden state $h_i$ and a key $k_i$. The dynamic memory module is made up of a set of blocks, which can be represented with a set of hidden states $\{ h_1,\dots,h_z\}$ and their corresponding set of keys $\{ k_1,\dots,k_z\}$. The equations used to update a generic block $i$ are the following: 
\begin{gather}
g_i^{(t)} = \text{Sigmoid}(s_t^\top h_i^{(t-1)} + s_t^\top k_i^{(t-1)}) \\  
\hat h_i^{(t)} = \text{ReLU}(U h_i^{(t-1)} + V k_i^{(t-1)} + W s_t ) \\ 
h_i^{(t)} = h_i^{(t-1)} + g_i^{(t)} \odot \hat h_i^{(t)}  \\
h_i^{(t)} = h_i^{(t)}/\| h_i^{(t)}, \|  
\end{gather}
where $g_i^{(t)}$ is the gating function which determines how much of the $i$th memory should be updated, and $\hat h_i^{(t)}$ is the new candidate value of the memory to be combined with the existing $h_i^{(t-1)}$. The matrices $U$, $V$, and $W$ are shared among different blocks, and are trained together with the key vectors. 

The output module creates a probability distribution over the memories and hidden states using the question $q$. Thus, the hidden states are summed, using the probability as weight, to obtain a single vector representing all the inputs. Finally, the network output is obtained by combining the final state with the question to predict the new utterance. 
The model is trained using a cross-entropy loss, and it outputs the next dialogue utterance by choosing among action templates. The lexicalization step simply copies entities in the table and replaces delexicalized elements in the action template to obtain the final response.

\subsection{Dynamic Query Memory Networks}

\begin{figure}[h]
\centering
\includegraphics[width=\linewidth]{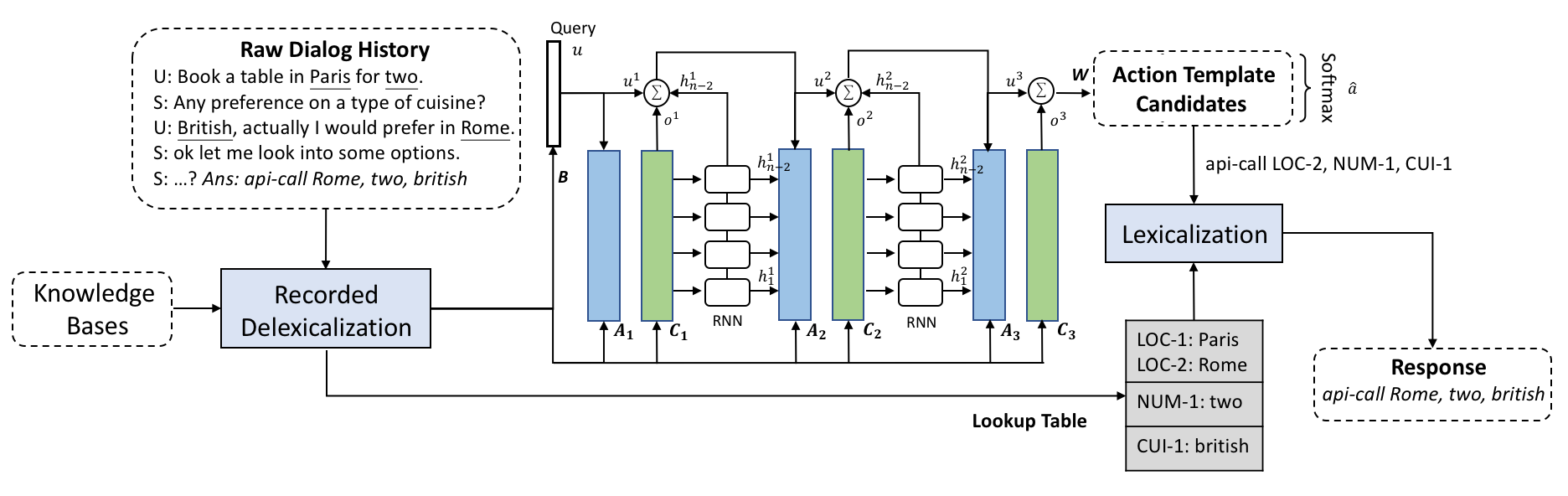}
\caption{Dynamic query memory networks with recorded delexicalization copying}
\label{fig:dqmem}
\end{figure}

One major drawback of end-to-end memory networks is that they are insensitive to representing temporal dependencies between memories. To mitigate the problem, we propose a novel architecture called a dynamic query memory network (DQMN) to capture time step information in dialogues, by utilizing RNNs between memory layers to represent latent dialogue states and the dynamic query vector. We adopt the idea from ~\cite{henaff2016tracking}, whose model can be seen as a bank of gated RNNs, and hidden states correspond to latent concepts and attributes. Therefore, to obtain a similar behavior, DQMN adds a recurrent architecture between memory hops in the original memory networks. We use the memory cells as the inputs of an GRU, based on the utterance order appearing in the dialogue history. 
The final hidden state of the GRU is added to the query $u_k$: 
\begin{equation}
  u^{k+1} = u^{k} + o^{k} + h_{N}^{k},
  \label{eq4}
\end{equation}
where $h_{N}^{k}$ is the last GRU hidden state at the hop $k$. 
In this way, compared to Eq.~(\ref{Eq:qk+1}), DQMN is able to capture the global attention over memory cells $o^{k}$, and also the internal latent representation of the dialogue state $h_{N}^{k}$. 

In addition, motivated by the query-reduction networks in~\cite{seo2016query}, we use each hidden state of the corresponding time step to query the next memory cells separately. 
That is, the next hop query vector is not generic over all the memory cells but customized. 
Each cell has its unique query vector 
\begin{equation}
  q_{i}^{k+1} = u^{k+1} + h_{i}^{k},
  \label{eq5}
\end{equation}
which is then sent to the attention computation in Eq.~(\ref{Eq:mnattention}).
DQMN considers the previous hop memory cells as a sequence of query-changing triggers, which trigger the GRU to generate more dynamically informed queries. 
Therefore, it can effectively alleviate temporal problems with the dynamic query components.

\section{Experimental Setup}
\label{SEC:bAbI-Dataset}

\subsection{Dataset}

\begin{table}[h]
\centering
\caption{Statistics of bAbI dialogue dataset.}
\label{tab:bAbI}
\begin{tabular}{r|ccccc}
\hline
\textbf{Task} & \textbf{1} & \textbf{2} & \textbf{3} & \textbf{4} & \textbf{5} \\ \hline
\textit{Avg. User turns} & 4 & 6.5 & 6.4 & 3.5 & 12.9  \\
\textit{Avg. Sys turns} & 6 & 9.5 & 9.9 & 3.5 & 18.4 \\
\textit{Avg. KB results} & 0 & 0 & 24 & 7 & 23.7  \\ 
\textit{Avg. Sys words} & 6.3 & 6.2 & 7.2 & 5.7 & 6.5  \\ \hline
\textit{Vocabulary} & \multicolumn{5}{c}{3747}  \\
\textit{Train dialogs} & \multicolumn{5}{c}{1000}  \\
\textit{Val dialogs} & \multicolumn{5}{c}{1000}  \\
\textit{Test dialogs} & \multicolumn{5}{c}{1000 + 1000 OOV} \\ \hline
\end{tabular}
\end{table}

We use the bAbI dialogue dataset~\cite{bordes2016learning} to evaluate the performance. 
It is one of the standard benchmarks since it evaluates all the desirable features of an end-to-end task-oriented dialogue system. 
The dataset is divided into five different tasks, each of which has its own training, validation, and test set. Tasks 1--4 are issuing an API call, refining an API call, recommending options, and providing additional information, respectively. 
Task 5 (full dialogues) is a union of tasks 1--4 and includes more conversational turns. 
\begin{itemize}
    \item Task 1 Issuing API calls: A user request implicitly defines a query that can contain the required fields. The bot must ask questions for filling the missing fields and eventually generate the correct corresponding API call. The bot asks for information in a deterministic order, making prediction possible.
    
    \item Task 2 Updating API calls: Starting by issuing an API call as in Task 1, users then ask to update their requests between 1 and 4 times. The order in which fields are updated is random. The bot must ask users if they are done with their updates and issue the updated API call.
    
    \item Task 3 Displaying options: Given a user request, the KB is queried using the corresponding API call and add the facts resulting from the call to the dialogue history. The bot must propose options to users by listing the restaurant names sorted by their corresponding rating (from higher to lower) until users accept.
    
    \item Task 4 Providing extra information: Given a user request, a restaurant is sampled and start the dialogue as if users had agreed to book a table there. We add all KB facts corresponding to it to the dialogue. Users then ask for the phone number of the restaurant, its address or both. The bot must learn to use the KB facts correctly to answer.
    
    \item Task 5 Conducting full dialogues: Tasks 1--4 are combined to generate full dialogues.
\end{itemize}
There are two test sets for each task: one follows the same distribution as the training set and the other has OOV words from a different KB, i.e., the slot values that do not appear in the training set. Dataset statistics are reported in Table~\ref{tab:bAbI}.

\subsection{Training}
\subsubsection{Recurrent Entity Networks}
For each task, we fix the number of the memory block to five because there are five different slots, and we use Adam \cite{kingma2014adam} optimizer with learning rate 0.1 and dropout ratio 0.3. The weights are initialized randomly from a Gaussian distribution with zero mean and $\sigma=0.1$. The gradient is clipped to a maximum of 40 to avoid gradient explosion. We try a small grid search over hyper-parameters such as batch size and embedding size. Then, the setting that achieves the highest accuracy in validation is selected for the final evaluation.

\subsubsection{Dynamic Query Memory Network}
All experiments used a $K=3$ hops model with the adjacent weight sharing scheme, e.g., ${\bf C}^{k}={\bf A}^{k+1}$ for $k=1,2$. For sentence representation, we use position encoding bag-of-words as in \cite{sukhbaatar2015end} to capture words order. Our model is using a learning rate of 0.01, annealing half every 25 epochs until 100 epochs are reached. The weights are initialized randomly from a Gaussian distribution with zero mean and $\sigma=0.1$. 
All training uses a batch size of 16 (but the cost is not averaged over a batch), and gradients with a $L2$ norm greater than 40 are clipped. 

\subsection{Evaluation Metrics}
\paragraph{Per-response/dialogue Accuracy} 
Per-response accuracy is considered correct only if the predicted system response is exactly the same as the gold system response. 
In addition, per-dialogue accuracy takes every turn in one dialogue into account.
It is considered correct if all the system responses are exactly the same as the true dialogue. 
These are strict evaluation metrics and are highly dependent on the system behavior, which may only be suitable for a simulated dialogue dataset, where all the responses are regular and standardized.

\section{Results and Discussion}

\subsection{Quantitative Results}

\begin{table}[h]
\centering
\caption{Per-response accuracy and per-dialogue accuracy (in parentheses) on bAbI dialogue dataset using REN and DQMN.}
\label{tab:bAbI-result-dqmn}
\resizebox{\linewidth}{!}{
\begin{tabular}{r|cc|cc|cc}
\hline
\textit{\textbf{Task}} & \textbf{MN} & \textbf{GMN} & \textbf{REN} & \textbf{DQMN} & \textbf{REN+RDC} & \textbf{DQMN+RDC} \\ \hline
\textit{T1} & 99.9 (99.6) & 100 (100) & 100 (100) & 100 (100) & 100 (100) & 100 (100) \\
\textit{T2} & 100 (100) & 100 (100) & 100 (100) & 100 (100) & 100 (100) & 100 (100) \\
\textit{T3} & 74.9 (2.0) & 74.9 (3.0) & 74.9 (0) & 74.9 (2.0) & 91.4 (40.3) & 98.7 (90.8) \\
\textit{T4} & 59.5 (3.0) & 57.2 (0) & 57.2 (0) & 57.2 (0) & 100 (100) & 100 (100) \\
\textit{T5} & 96.1 (49.4) & 96.3 (52.5) & 97.3 (46.3) & 99.2 (88.7) & 97.5 (62.5) & 99.9 (98.3) \\ \hline
\textit{T1-OOV} & 72.3 (0) & 82.4 (0) & 81.8 (0) & 82.5 (0) & 100 (100) & 100 (100) \\
\textit{T2-OOV} & 78.9 (0) & 78.9 (0) & 78.9 (0) & 78.9 (0) & 100 (100) & 100 (100) \\
\textit{T3-OOV} & 74.4 (0) & 75.3 (0) & 75.3 (0) & 74.9 (0) & 89.6 (29.8) & 98.7 (90.4) \\
\textit{T4-OOV} & 57.6 (0) & 57.0 (0) & 57.0 (0) & 57.0 (0) & 100 (100) & 100 (100) \\
\textit{T5-OOV} & 65.5 (0) & 66.7 (0) & 65.1 (0) & 72.0 (0) & 96.0 (46.3) & 99.4 (91.6) \\ \hline
\end{tabular}
}
\end{table}

In Table~\ref{tab:bAbI-result-dqmn}, we report the results obtained on the bAbI dialogue test sets (including the OOV). 
We compare our proposed models with and without RDC to the original end-to-end memory networks (MN)~\cite{sukhbaatar2015end} and gated memory networks (GMN) \cite{perez2016gated}, which are both retrieval-based models.

First, we discuss the model performance without the RDC strategy. On the full dialogue task (T5), REN and DQMN outperform memory network and GMN, and DQMN achieves the highest, 99.2\% per-response accuracy and 88.7\% per-dialogue accuracy. This result shows that the dynamic query components in DQMN allow the memory network to learn a more complex dialogue policy. Task 5 includes long conversational turns, and it requires a stronger dialogue state tracking ability. Although there is no performance difference between our models and the other baselines for T1 to T4, we can still observe a better generalization ability of our models on the OOV test set. For example, our model DQMN achieves 72.0\% per-response accuracy in the T5-OOV setting, which is 7\% better than others.

Next, we show the effectiveness of the RDC strategy by applying it to both REN and DQMN. On T3 the restaurant recommendation task, REN with RDC improves the performance by 16.5\% on per-response accuracy, and DQMN with RDC improves by 23.8\%. On T4, the providing additional information task, both REN and DQMN with RDC can achieve perfect performance. On T5 full dialogue task, DQMN with RDC achieves 99.9\% per-response accuracy and 98.3\% per-dialogue accuracy. Note that with RDC, the DQMN model can achieve almost perfect per-response accuracy, even on Task5-OOV, which also confirms our initial assumption that using RDC strongly decreases the learning complexity. This strategy leads to an overall accuracy improvement, which is particularly useful when the network needs to learn how to work with abstract OOV entities. 


\subsection{Visualization}

\begin{figure}[h]
\begin{subfigure}{0.45\textwidth}
  \centering
  \includegraphics[width=\linewidth]{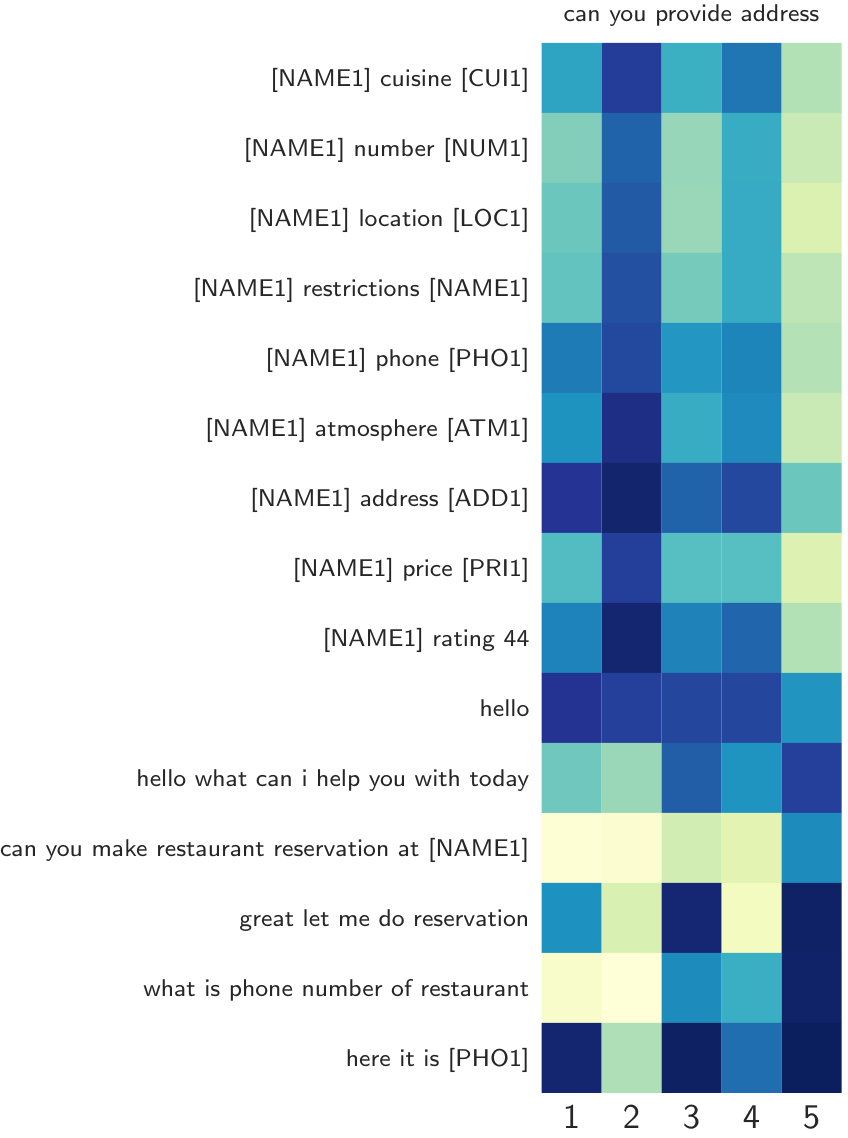}
  \caption{}
\end{subfigure}
\begin{subfigure}{0.55\textwidth}
  \centering
  \includegraphics[width=0.9\linewidth]{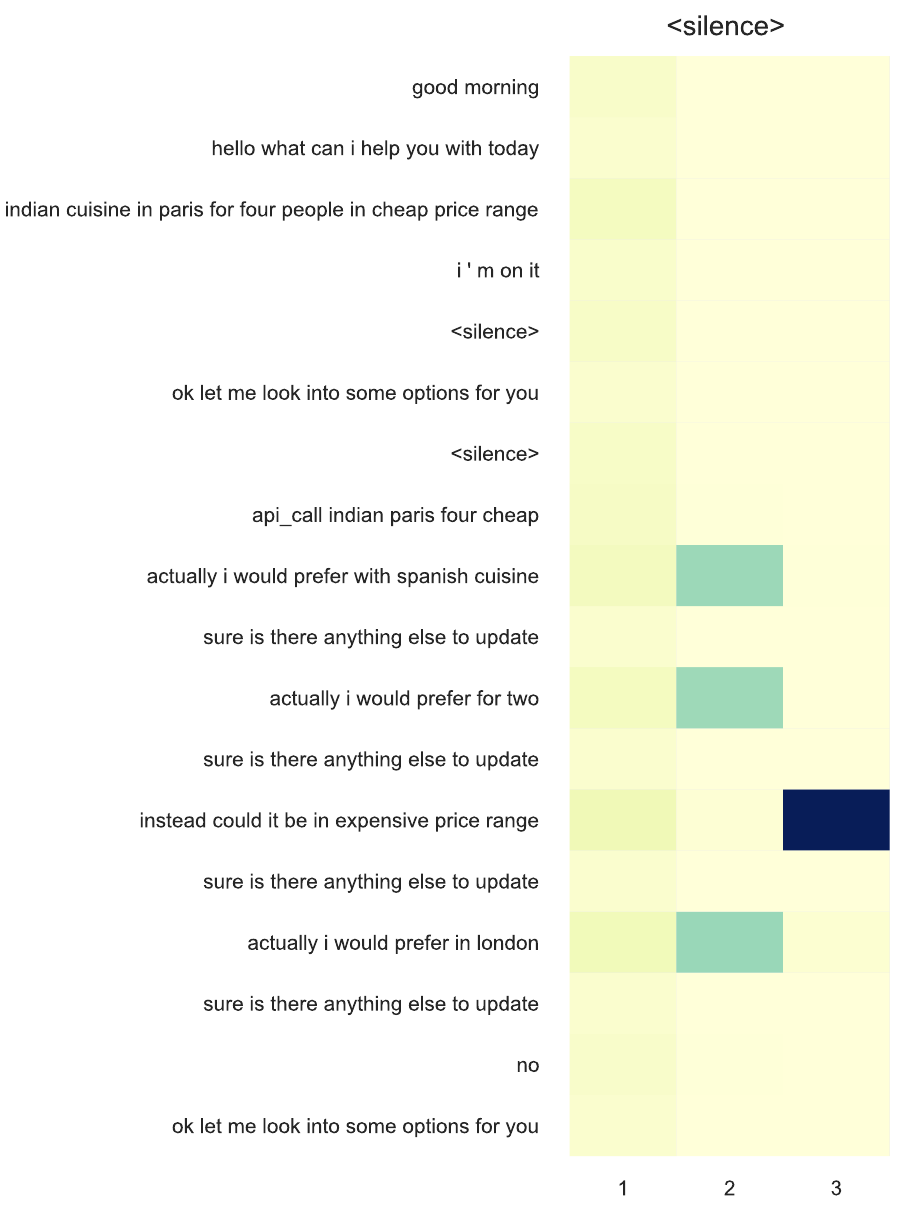}
  \caption{}
\end{subfigure}%
\caption{Heatmap representation of the (a) gating function for each memory block in the REN model and (b) memory attention for each hop in DQMN.}
\label{fig:REN_VIZ}
\end{figure}

To better understand the REN behavior, we visualize the gating activation function in Figure~\ref{fig:REN_VIZ}(a). The output of this function decides how much and what we store in each memory cell. We take the model trained on T4 (i.e., providing additional information) for the visualization. We plot the activation matrix of the gate function and observe how REN learns to store relevant information. As we can see in the figure, the model opens the memory gate once a useful information appears as input, and closes the gate for other useless sentences. Different memory blocks may focus on different information. For example, block 5 stores more information from the discourse rather than explicit KB knowledge; and block 2 opens its gate fully when the address and rating information is provided. In this case, the last user utterance (question) is ``\textit{Can you provide address?},", we can get the correct prediction because the latent address feature is represented in those memory blocks that open during the utterance ``\textit{[NAME1] address [ADD1]}".

We visualize the memory attentions of different hops in DQMN in Figure~\ref{fig:REN_VIZ}(b). One can observe that in the first hop, the model usually pays attention to almost every memory slot, which intuitively means that the model is ``understanding'' the general dialogue flow. In the second hop, the model focuses on three different slots (Spanish cuisine, two people, and London location) because the user has changed his/her mind to modify the intention. In the third hop, the model becomes very sharp on the utterance that is related to the price slot. In the end, DQMN gets all of the information it needs and predicts the output response ``\textit{api\_call Spanish London two expensive}''.

\section{Short Summary}
REN and DQMN are two memory-augmented frameworks for retrieval-based task-oriented dialogue systems, which are designed to model long dialogue context and external knowledge more efficiently.
They are designed to overcome the drawbacks in retrieval-based dialogue applications that they are hard to capture long-term dependencies. 
A recorded delexicalization copy mechanism is utilized to reduce the learning complexity and also alleviate out-of-vocabulary entity problems. 
The experimental results show that our models outperform other memory networks, especially on a task with longer dialogue turns.

%% file: chapter5.tex
\chapter{Generation-Based Memory-Augmented Dialogue Systems}

In the previous chapter, we discussed how to effectively incorporate long dialogue context and external knowledge base (KB) into retrieval-based dialogue systems. However, although they may be one of the most robust dialogue systems, they have two main drawbacks:
1) Retrieved responses are too regular and limited. While facing the real users, the system is not able to reply if they say something out-of-domain, which is not predefined in the response candidates.
2) Although using record delexicalization strategy can simplify the problem, the model does not deal with the real entity values in this case and might lose some information implied. For example, when users ask for a French restaurant, they may imply that the price of the dinner they want might not be cheap. This information is missing when the values are replaced by the slot type, i.e., replace ``French'' with ``CUISINE-1''.

In this chapter, on the other hand, we cope with another challenging approach of end-to-end dialogue learning, which is system response generation problem.
Given the dialogue history and KB information, machine learning models are required to generate the system response word-by-word using recurrent structures.
Compared to solely doing retrieval from the response candidates, generated responses can be more diverse, human-like, and have the potential to generalize to an unseen scenario.
We propose two models, a memory-to-Sequence (Mem2Seq) model ~\cite{mem2seq} and a global-to-local memory pointer network (GLMP) ~\cite{wu2019global}, to effectively incorporate long dialogue context and external knowledge into end-to-end learning.

A multi-turn dialogue between a driver and an agent is shown in Table~\ref{TB:EXAMPL_DIALOG_1}. 
The upper part of the table is the KB information available, which includes different points-of-interest (POIs), and their corresponding addresses, types, traffic information, and distances.
One can find that the system responses include multiple entities that are existed in the table, e.g., \textit{Palo Alto Cafe}, \textit{4 miles}, and \textit{436 Alger Dr}.
Therefore, the ability to reason over the KB information and copy the entities from the KB to the response is essential.

\begin{table}[t]
\centering
\caption{Multi-turn dialogue example for an in-car assistant in the navigation domain.}
\label{TB:EXAMPL_DIALOG_1}
\resizebox{0.8\linewidth}{!}{
\begin{tabular}{|c|c|c|c|c|}
\hline
\textbf{Distance} & \textbf{Traffic\_info} & \textbf{Poi\_type}  & \textbf{Address}    & \textbf{Poi}          \\ \hline
5 miles           & moderate traffic       & rest stop           & 329 El Camino Real  & The Westin            \\ \hline
4 miles           & no traffic             & pizza restaurant    & 113 Anton Ct        & Round Table           \\ \hline
5 miles           & no traffic             & chinese restaurant  & 271 Springer Street & Mandarin Roots        \\ \hline
4 miles           & moderate traffic       & coffee or tea place & 436 Alger Dr        & Palo Alto Cafe        \\ \hline
6 miles           & heavy traffic          & pizza restaurant    & 776 Arastradero Rd  & Dominos               \\ \hline
6 miles           & no traffic             & hospital            & 214 El Camino Real  & Stanford Express Care \\ \hline
2 miles           & heavy traffic          & rest stop           & 578 Arbol Dr        & Hotel Keen            \\ \hline
\end{tabular}
}
\resizebox{0.8\linewidth}{!}{
\begin{tabular}{l|rl}
\hline
\multirow{2}{*}{\textit{1st Turn}} & \textbf{DRIVER} & Where can I get tea? \\ \cline{2-3} 
 & \textbf{System} & \begin{tabular}[c]{@{}l@{}}Palo Alto Cafe is 4 miles away and serves coffee and tea. \\ Do you want the address?\end{tabular} \\ \hline
\multirow{2}{*}{\textit{2nd Turn}} & \textbf{DRIVER} & Yes. \\ \cline{2-3} 
 & \textbf{System} & Palo Alto is located at 436 Alger Dr. \\ \hline
\end{tabular}
}
\end{table}

\section{Memory-to-Sequence}
As mentioned in Section~\ref{SEC:MANN}, existing LSTM or GRUs usually are unable to incorporate external knowledge into end-to-end learning.
We present a novel architecture called Mem2Seq to learn task-oriented dialogues in an end-to-end manner. 
This model augments the existing memory network framework with a sequential generative architecture, using global multi-hop attention mechanisms to copy words directly from dialogue history or KBs. 
Mem2Seq is the first model to combine multi-hop attention mechanisms with the idea of pointer networks, which allows us to effectively incorporate KB information. 
It also learns how to generate dynamic queries to control memory access, and we visualize and interpret the model dynamics among hops for both the memory controller and the attention.
Lastly, Mem2Seq can be trained faster and achieves state-of-the-art results on several task-oriented dialogue datasets.

\subsection{Model Description}
\begin{figure}[t]
\centering
\includegraphics[width=\linewidth]{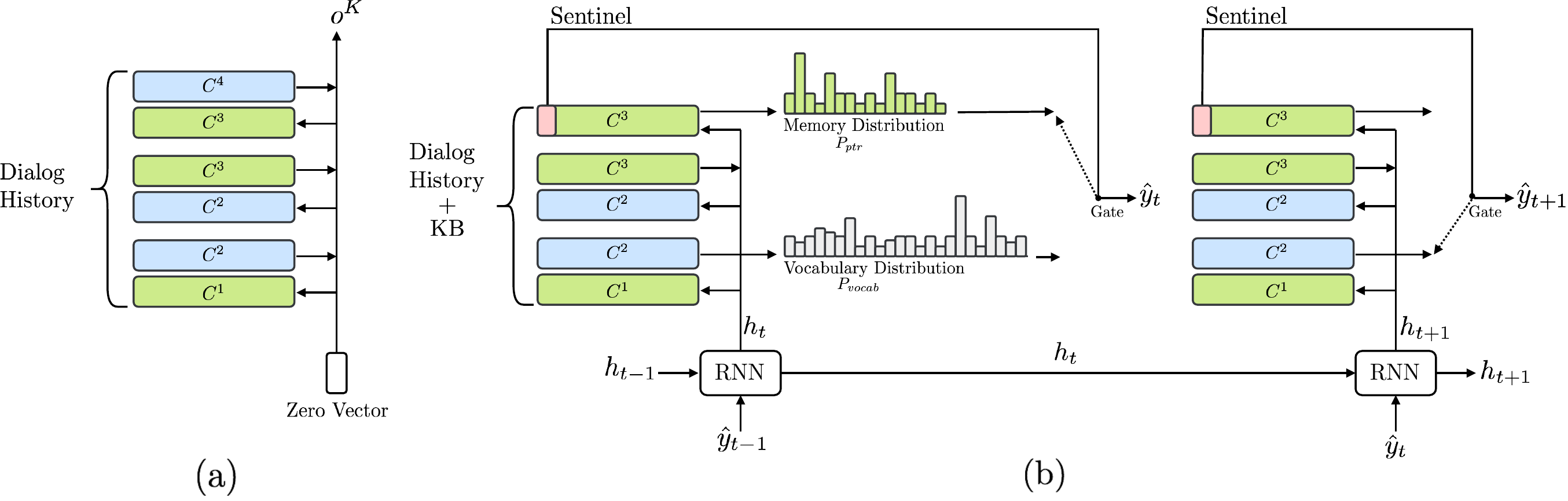}
\caption{The proposed Mem2Seq architecture for task-oriented dialogue systems. (a) Memory encoder with three hops and (b) memory decoder over two-step generation.}
\label{FIG:Mem2Seq}
\end{figure}

Mem2Seq~\footnote{The code is available at \url{https://github.com/HLTCHKUST/Mem2Seq}} comprises of two components: an MN encoder and a memory decoder, as shown in Figure~\ref{FIG:Mem2Seq}.
The memory network encoder creates a vector representation of the dialogue history. Then the memory decoder reads and copies the memory to generate a response.

\subsubsection{Memory Content}
We store word-level content in the memory module. 
Similar to ~\cite{bordes2016learning}, we add temporal information and speaker information to capture the sequential dependencies. For example, ``\textit{hello $t1$ $\$u$}'' means ``\textit{hello}'' at time step 1 spoken by a user.
On the other hand, to store the KB information, we follow the works \cite{miller-EtAl:2016:EMNLP2016} and \cite{ericKVR2017}, which use a \textit{(subject, relation, object)} representation. 
For example, we represent the information of \textit{Dominos} in Table~\ref{TB:EXAMPL_DIALOG_1}: \textit{(Dominos, Distance, 6 miles)}. 
Then we sum word embeddings of the subject, relation, and object to obtain each KB memory representation. 
During the decoding stage, the object part is used as the generated word for copying. 
For instance, when the KB triplet \textit{(Dominos, Distance, 6 miles)} is pointed to, our model copies ``\textit{6 miles}'' as an output word.

\subsubsection{Memory Encoder}
Mem2Seq uses a standard memory network with adjacent weighted tying as an encoder. 
The input of the encoder is the dialogue history only because we believe that the encoder does not require KB information to track dialogue states. 
The memories of MemNN are represented by a set of trainable embedding matrices, and a query vector is used as a reading head. The model loops over $K$ hops and it computes the attention weights at hop $k$ for each memory.
Mem2Seq will return a soft memory selector that decides the memory relevance with respect to the query vector. 
The model also reads out the memory by the weighted-sum of embeddings.
The result from the encoding step is the memory readout vector, which will become the initial state of the decoder RNN.  

\subsubsection{Memory Decoder}
The decoder uses an RNN and MN. The MN is loaded with both the dialogue history and external knowledge since we use both to generate a proper system response. 
A GRU is used as a dynamic query generator for the MN. 
At each decoding step $t$, the GRU gets the previously generated word and the previous query as input, and it generates the new query vector.
Then the query is passed to the MN, which will produce the token. 
At each time step, two distributions are generated: one over all the words in the vocabulary ($P_{vocab}$) and the other over the memory contents ($P_{ptr}$). 
The first, $P_{vocab}$, is generated by concatenating the first hop memory readout and the current query vector: 
\begin{equation}
P_{vocab}(\hat{y_t}) = \text{Softmax}(W [h^{dec}_t; o^1]).
\end{equation}
On the other hand, $P_{ptr}$ is generated using the attention weights at the last MN hop of the decoder. 
The decoder generates tokens by pointing to the input words in the memory, which is a similar mechanism to the attention used in pointer networks~\cite{Vinyals2015PointerN}.

If the expected word does not appear in the memories, $P_{ptr}$ is trained to produce the sentinel token $\$$. 
To sum up, once the sentinel is chosen, our model generates the token from $P_{vocab}$; otherwise, it takes the memory content using the $P_{ptr}$ distribution. 
Basically, the sentinel token is used as a hard gate to control which distribution to use at each time step. 
A similar approach has been used in~\cite{Merity2016PointerSM} to control a soft gate in a language modeling task. 
With this method, the model does not need to learn a gating function separately as in~\cite{gulcehre2016pointing}, and is not constrained by a soft gate function as in~\cite{see-liu-manning:2017:Long}. 

We designed our architecture in this way because we expect the attention weights in the first and the last hop to show a ``looser'' and ``sharper'' distribution, respectively. 
To elaborate, the first hop focuses more on retrieving memory information and the last tends to choose the exact token leveraging the pointer supervision. 
Hence, during training, all the parameters are jointly learned by minimizing the sum of two standard cross-entropy losses: one between $P_{vocab}$ and the true response word for the vocabulary distribution, and one between $P_{ptr}$ and the true memory position for the memory distribution.

\subsection{Experimental Setup}

\subsubsection{Dataset}
We use three public multi-turn task-oriented dialogue datasets to evaluate our model: the bAbI dialogue~\cite{bordes2016learning}, Dialogue State Tracking Challenge 2 (DSTC2)~\cite{henderson2014second} and In-Car Assistant~\cite{ericKVR2017}.
The training/validation/test sets of these three datasets have been split in advance by the providers.
The dataset statistics are reported in Table~\ref{Tab:statistic-Mem2Seq}. 

The bAbI dialogue dataset was introduced in Section~\ref{SEC:bAbI-Dataset}. 
For the dialogues extracted from the DSTC2 we used the refined version from~\cite{bordes2016learning}, which ignores the dialogue state annotations. 
The main difference from the bAbI dialogue dataset is that this dataset is extracted from real human-bot dialogues, which are noisier and harder since the bots make mistakes due to speech recognition errors or misinterpretations. The In-Car Assistant dataset, meanwhile, is a human-human, multi-domain dialogue dataset collected from Amazon Mechanical Turk. 
It has three distinct domains: calendar scheduling, weather information retrieval, and point-of-interest navigation. 
This dataset has shorter conversation turns, but the user and system behaviors are more diverse. 
In addition, the system responses are variant and the KB information is much more complicated.
Hence, this dataset requires a stronger ability to interact with KBs, rather than dialogue state tracking.

\begin{table}[t]
\centering
\caption{Dataset statistics for three different datasets, bAbI dialogue, DSTC2, and In-Car Assistant.}
\label{Tab:statistic-Mem2Seq}
\begin{tabular}{r|ccccc|c|c}
\hline
\textbf{Task} & \textbf{1} & \textbf{2} & \textbf{3} & \textbf{4} & \textbf{5} & \textbf{DSTC2} & \textbf{In-Car} \\ \hline
\textit{Avg. User turns} & 4 & 6.5 & 6.4 & 3.5 & 12.9 & 6.7 & 2.6 \\
\textit{Avg. Sys turns} & 6 & 9.5 & 9.9 & 3.5 & 18.4 & 9.3 & 2.6 \\
\textit{Avg. KB results} & 0 & 0 & 24 & 7 & 23.7 & 39.5 & 66.1 \\ 
\textit{Avg. Sys words} & 6.3 & 6.2 & 7.2 & 5.7 & 6.5 & 10.2 & 8.6 \\ 
\textit{Max. Sys words} & 9 & 9 & 9 & 8 & 9 & 29 & 87 \\
\textit{Pointer Ratio} & .23 & .53 & .46 & .19 & .60 & .46 & .42\\ \hline
\textit{Vocabulary} & \multicolumn{5}{c|}{3747} & 1229 & 1601 \\
\textit{Train dialogues} & \multicolumn{5}{c|}{1000} & 1618 & 2425 \\
\textit{Val dialogues} & \multicolumn{5}{c|}{1000}  & 500 & 302 \\
\textit{Test dialogues} & \multicolumn{5}{c|}{1000 + 1000 OOV} & 1117 & 304 \\ \hline
\end{tabular}
\end{table}

\subsubsection{Evaluation Metrics}
\paragraph{Per-response/dialogue Accuracy}
For per-response accuracy, a generated response is correct only if it is exactly the same as the gold response; for per-dialogue accuracy, a dialogue is considered correct only if every system response in the dialogue is exactly the same as the gold response.
These evaluation metrics are only suitable for the bAbI dialogue dataset because it is a simulated dataset with very regular system behavior.
Note that \cite{bordes2016learning} tests their model by selecting the system response from predefined response candidates; that is, their system solves a multi-class classification problem. 
On the other hand, Mem2Seq generates each token individually so evaluating with these metrics is more challenging.

\paragraph{BLEU} 
BLEU~\cite{papineniBLEU2002} is a measure commonly used for machine translation systems, but it has also been used in evaluating dialogue systems~\cite{eric-manning:2017:EACLshort,zhao2017generative} and chat-bots~\cite{li-EtAl:2016:N16-11}. 
Moreover, the BLEU score is a relevant measure on task-oriented dialogues as there is little variance between the generated answers, unlike open-domain generation. 
\paragraph{Entity F1}
We micro-average over the entire set of system responses and compare the entities in plain text. 
The entities in each gold system response are selected by a predefined entity list. This metric evaluates the ability to generate relevant entities from the provided KBs and to capture the semantics of the dialogue~\cite{eric-manning:2017:EACLshort,ericKVR2017}.

\subsection{Results and Discussion}
We compare Mem2Seq with hop 1, 3, and 6 hops with several existing models: an end-to-end memory networks (MN), gated end-to-end memory networks (GMN), and dynamic query memory networks (DQMN). 
We also implement the following baseline models: a sequence-to-sequence (Seq2Seq) model with and without attention~\cite{Luong2015EffectiveAT}, and pointing to unknown (Ptr-Unk)~\cite{gulcehre2016pointing}. 

\subsubsection{Quantitative Study}

\begin{table}[t]
\centering
\resizebox{\linewidth}{!}{
\begin{tabular}{r|ccc|ccc|ccc}
\hline
\textbf{Task} & MN & GMN & DQMN & Seq2Seq & Seq2Seq+Attn & Ptr-Unk & \textbf{\begin{tabular}[c]{@{}c@{}}Mem2Seq K1\end{tabular}} & \textbf{\begin{tabular}[c]{@{}c@{}}Mem2Seq K3\end{tabular}} & \textbf{\begin{tabular}[c]{@{}c@{}}Mem2Seq K6\end{tabular}} \\ \hline
\textit{T1} & 99.9 (99.6) & 100 (100) & 100 (100) & 100 (100) & 100 (100) & 100 (100) & 100 (100) & 100 (100) & 100 (100) \\
\textit{T2} & 100 (100) & 100 (100) & 100 (100) & 100 (100) & 100 (100) & 100 (100) & 100 (100) & 100 (100) & 100 (100) \\
\textit{T3} & 74.9 (2.0) & 74.9 (0) & 74.9 (2.0) & 74.8 (0) & 74.8 (0) & 85.1 (19.0) & 87.0 (25.2)   & 94.5 (59.6)   & \textbf{94.7 (62.1)} \\
\textit{T4} & 59.5 (3.0) & 57.2 (0)  & 57.2 (0) & 57.2 (0) & 57.2 (0) & \textbf{100 (100)} & 97.6 (91.7) & \textbf{100 (100)} & \textbf{100 (100)} \\
\textit{T5} & 96.1 (49.4) & 96.3 (52.5) & 99.2 (88.7) & 98.8 (81.5) & 98.4 (87.3) & \textbf{99.4 (91.5)} & 96.1 (45.3) & 98.2 (72.9) & 97.9 (69.6) \\ \hline
\textit{T1-OOV} & 72.3 (0) &    82.4 (0) & 82.5 (0) &   79.9 (0) &    81.7 (0) &    92.5 (54.7) &    93.4 (60.4) &    91.3 (52.0) &    \textbf{94.0 (62.2)} \\
\textit{T2-OOV} & 78.9 (0) &    78.9 (0) & 78.9 (0) &   78.9 (0) &    78.9 (0) &    83.2 (0) &    81.7 (1.2) &    84.7 (7.3) &    \textbf{86.5 (12.4)} \\
\textit{T3-OOV} & 74.4 (0) &    75.3 (0) & 74.9 (0)&   74.3 (0) &    75.3 (0) &    82.9 (13.4) &    86.6 (26.2) &    \textbf{93.2 (53.3)} &    90.3 (38.7) \\
\textit{T4-OOV} & 57.6 (0) &    57.0 (0) & 57.0 (0) &   57.0 (0) &    57.0 (0) &    \textbf{100 (100)} &    97.3 (90.6) &    \textbf{100 (100)} &    \textbf{100 (100)} \\
\textit{T5-OOV} & 65.5 (0) &    66.7 (0) & 72.0 (0) &  67.4 (0) &    65.7 (0) &    73.6 (0) &    67.6 (0) &    78.1 (0.4) &    \textbf{84.5 (2.3)} \\ \hline
\end{tabular}}
\caption{Mem2Seq evaluation on simulated bAbI dialogues. Generation methods, especially with copy mechanism, outperform other retrieval baselines.}
\label{TB:babi-dialog-Mem2Seq}
\end{table}

\paragraph{bAbI Dialogue}
As shown in Table~\ref{TB:babi-dialog-Mem2Seq}, Mem2Seq with 6 hops achieves 84.5\% per-response accuracy for the OOV test set, which surpasses existing methods by far.
This indicates that our model can generalize well, with only 13.4\% performance loss for test OOV data, while the others have around a 25--35\% drop. 
This performance gain on the OOV data can be mainly attributed to the use of the copy mechanism. 
In addition, the effectiveness of the hops is demonstrated in tasks 3--5, since they require reasoning ability over the KB information. 
Note that the MN, GMN, and DQMN view bAbI dialogue tasks as classification problems, which are easier to solve compared to our generative methods.
Finally, one can find that the Seq2Seq and Ptr-Unk models are also strong baselines, which further confirms that generative methods can achieve good performance in task-oriented dialogue systems~\cite{eric-manning:2017:EACLshort}.    

\begin{table}[h]
\centering
\caption{Mem2Seq evaluation on human-robot DSTC2. We make a comparison based on entity F1 score, and per-response/dialogue accuracy is low in general.}
\label{TB:DSTC2-Mem2Seq}
\resizebox{0.55\linewidth}{!}{
\begin{tabular}{r|c|c|c|c}
\hline
\multicolumn{1}{l|}{\textbf{}} & \textbf{Ent. F1} & \textbf{BLEU} &  \textbf{Per-Resp.} & \textbf{Per-dial.}  \\ \hline
\textit{Rule-Based} & - & - & 33.3 & - \\ 
\textit{QRN} & - & - & 43.8 & - \\
\textit{MN} & - & - & 41.1 & 0.0 \\
\textit{GMN} & - & - & \textbf{47.4} & 1.4 \\ 
\textit{Seq2Seq} & 69.7 & 55.0 & 46.4 & \textbf{1.5} \\
\textit{+Attn} & 67.1  & \textbf{56.6} & 46.0 & 1.4 \\
\textit{+Copy} & 71.6 & 55.4 & 47.3 & 1.3 \\ \hline
\textit{\textbf{Mem2Seq K1}} & 72.9 & 53.7  & 41.7 & 0.0 \\
\textit{\textbf{Mem2Seq K3}} & \textbf{75.3} & 55.3  & 45.0 & 0.5 \\
\textit{\textbf{Mem2Seq K6}} & 72.8 & 53.6  & 42.8 & 0.7\\ \hline
\end{tabular}
}
\end{table}

\paragraph{DSTC2}
In Table~\ref{TB:DSTC2-Mem2Seq}, the results compared to the Seq2Seq models from~\cite{eric-manning:2017:EACLshort} and the rule-based model from~\cite{bordes2016learning} are reported. 
Mem2Seq achieves the highest, 75.3\%m BLEU score among the other baselines.
Note that the per-response accuracy for every model is less than 50\% since the dataset is quite noisy and it is hard to generate a response that is exactly the same as the gold response. 

\paragraph{In-Car Assistant}
In Table~\ref{TB:INCAR-Mem2Seq}, our model can achieve highest the highest BLEU score, or 12.6, and a 33.4\% entity F1 score, which surpasses other models. 
Note that baselines such as Seq2Seq and Ptr-Unk have especially poor performances on this dataset since it is very inefficient for RNN methods to encode longer KB information. 
This is the advantage of Mem2Seq.

Furthermore, we observe the interesting phenomenon that humans can easily achieve a high entity F1 score with a low BLEU score. This implies that stronger reasoning ability over entities (hops) is crucial, but the results may not be similar to the golden answer. 
We believe humans could produce good answers even with a low BLEU score since there are different ways to express the same concept. Therefore, Mem2Seq shows the potential to successfully choose the correct entities.

\begin{table}[t]
\centering
\caption{Mem2Seq evaluation on human-human In-Car Assistant dataset.}
\label{TB:INCAR-Mem2Seq}
\resizebox{0.65\linewidth}{!}{
\begin{tabular}{r|c|c|ccc}
\hline
\multicolumn{1}{l|}{\textbf{}} & \textbf{BLEU}  & \textbf{Ent. F1} & \textbf{Sch. F1} & \textbf{Wea. F1} & \textbf{Nav. F1} \\ \hline
\textit{Human} & 13.5 & 60.7 & 64.3 & 61.6 & 55.2  \\
\textit{Rule-Based} & 6.6 & 43.8 & 61.3 & 39.5 & 40.4 \\ \hline
\textit{Seq2Seq} & 8.4 & 10.3 & 09.7 & 14.1 & 07.0  \\
\textit{+Attn} & 9.3 & 19.9 & 23.4 & 25.6 & 10.8  \\
\textit{Ptr-Unk}  & 8.3 & 22.7 & 26.9 & 26.7 & 14.9  \\ \hline
\textit{\textbf{Mem2Seq H1}} & 11.6 & 32.4 & 39.8 & \textbf{33.6} & \textbf{24.6}   \\
\textit{\textbf{Mem2Seq H3}} & \textbf{12.6} & \textbf{33.4} & \textbf{49.3}    & 32.8 & 20.0  \\
\textit{\textbf{Mem2Seq H6}} & 9.9   & 23.6    & 34.3 & 33.0    & 4.4 \\
\hline
\end{tabular}}
\end{table}  

\begin{figure}[H]
\centering
\includegraphics[width=0.6\linewidth]{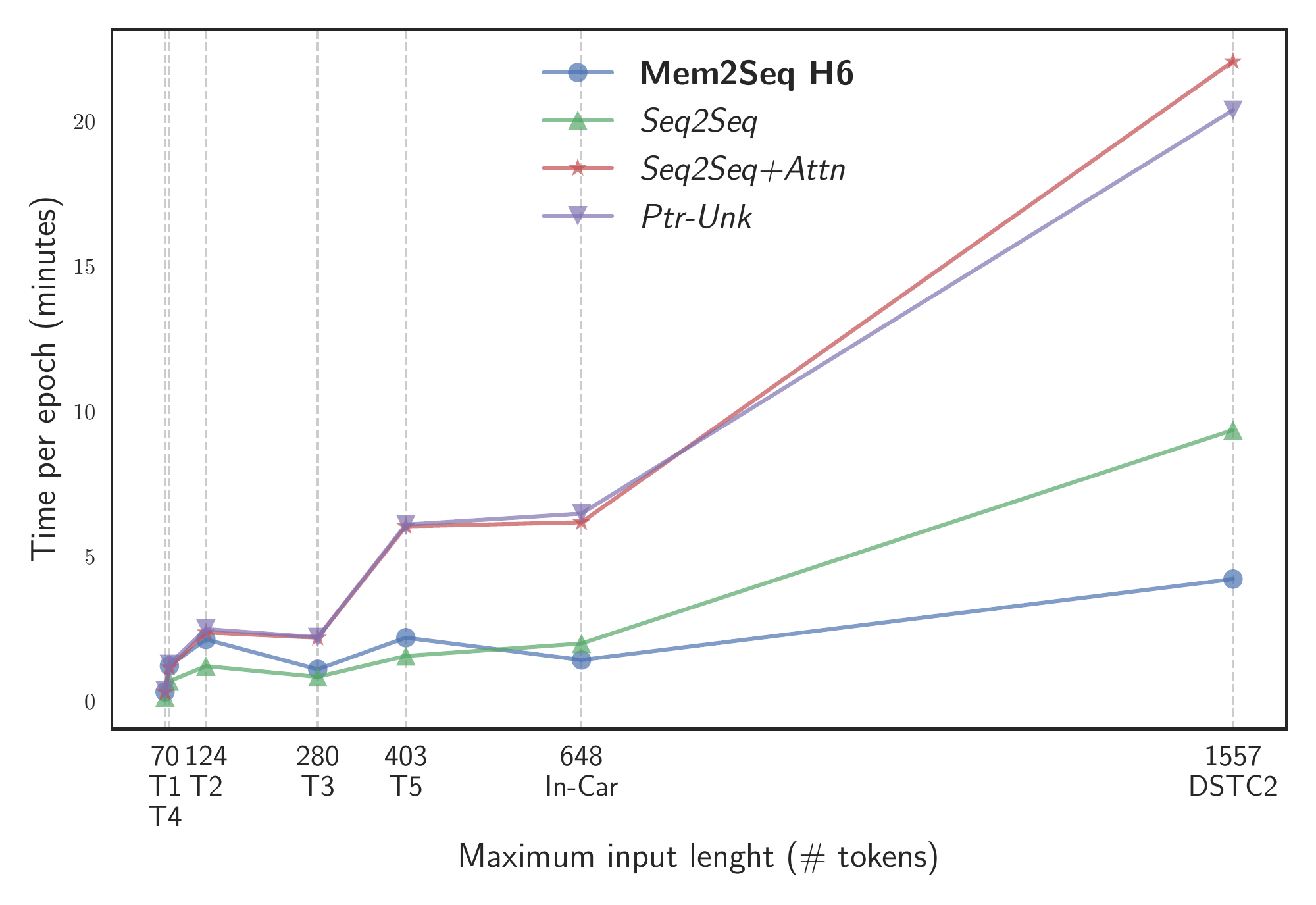}
\caption{Training time per-epoch for different tasks.}
\label{FIG:speed-Mem2Seq}
\end{figure}

\paragraph{Time Per-Epoch} 
We also compare the training times, as shown in Figure~\ref{FIG:speed-Mem2Seq}, using an Intel(R) Core(TM) i7-3930K CPU@3.20GHz, with a GeForce GTX 1080 Ti. 
The experiments are set with batch size 16, and we report each model with the hyper-parameter that can achieve the highest performance. 
One can observe that the training time is not that different for short input lengths (bAbI dialogue tasks 1--4) and the gap becomes larger as the maximal input length increases. 
Mem2Seq is around five times faster on In-Car Assistant and DSTC2 compared to Seq2Seq with attention. 
This difference in training efficiency is mainly attributed to the fact that Seq2Seq models have input sequential dependencies which limit any parallelization. 
Moreover, it is unavoidable for Seq2Seq models to encode KBs, while Mem2Seq only encodes with dialogue history.

\subsubsection{Qualitative Study}
Table~\ref{TB:EXAMPL_DIALOG_1_MEM2SEQ}, Table~\ref{TB:EXAMPL_DIALOG_2_MEM2SEQ}, Table~\ref{TB:EXAMPL_DIALOG_3_MEM2SEQ} and  show the generated responses of different models on the three test set samples from the In-Car Assistant dataset.
Seq2Seq generally cannot produce related information, and sometimes fails in language modeling. Using attention instead helps with this issue, but it still rarely produces the correct entities. 
For example, Seq2Seq with attention generated ``\textit{5 miles}'' but the correct response is ``\textit{4 miles}''. In addition, Ptr-Unk often cannot copy the correct token from the input, as shown by ``PAD''. 
On the other hand, Mem2Seq is able to produce the correct responses in these two examples. 
In particular, in the navigation domain, Mem2Seq produces a different but still correct utterance. 

\vspace{30pt}

\begin{table}[h]
\centering
\caption{Example of generated responses for the In-Car Assistant on the navigation domain.}
\label{TB:EXAMPL_DIALOG_1_MEM2SEQ}
\resizebox{0.9\linewidth}{!}{
\begin{tabular}{|c|c|c|c|c|}
\hline
\textbf{Distance} & \textbf{Traffic\_info} & \textbf{Poi\_type}  & \textbf{Address}    & \textbf{Poi}          \\ \hline
5 miles           & moderate traffic       & rest stop           & 329 El Camino Real  & The Westin            \\ \hline
4 miles           & no traffic             & pizza restaurant    & 113 Anton Ct        & Round Table           \\ \hline
5 miles           & no traffic             & chinese restaurant  & 271 Springer Street & Mandarin Roots        \\ \hline
4 miles           & moderate traffic       & coffee or tea place & 436 Alger Dr        & Palo Alto Cafe        \\ \hline
6 miles           & heavy traffic          & pizza restaurant    & 776 Arastradero Rd  & Dominos               \\ \hline
6 miles           & no traffic             & hospital            & 214 El Camino Real  & Stanford Express Care \\ \hline
2 miles           & heavy traffic          & rest stop           & 578 Arbol Dr        & Hotel Keen            \\ \hline
\end{tabular}
}
\resizebox{0.85\linewidth}{!}{
\begin{tabular}{l|rl}
\hline
\multirow{6}{*}{\rotatebox{90}{\textit{1th Turn}}} & \textbf{DRIVER} & Where can I get tea? \\ \cline{2-3} 
 & \textit{Seq2Seq} & I have a away from away would you like the address \\ 
 & \textit{+Attn} & I have a listing for a place that serves tea that is 5 miles away \\ 
 & \textit{Ptr-Unk} & There is a away you like would you like more info \\ \cline{2-3} 
 & \textit{\textbf{Mem2Seq}} & \begin{tabular}[c]{@{}l@{}}The nearest tea shop is Palo Alto Cafe located 4 miles away, \\ would you like directions there?\end{tabular} \\ \cline{2-3} 
 & \textbf{GOLD} & \begin{tabular}[c]{@{}l@{}}Palo Alto Cafe is 4 miles away and serves coffee and tea. \\ Do you want the address?\end{tabular} \\ \hline
\multirow{6}{*}{\rotatebox{90}{\textit{2th Turn}} } & \textbf{DRIVER} & Yes. \\ \cline{2-3} 
 & \textit{Seq2Seq} & The is at at \\ 
 & \textit{+Attn} & \begin{tabular}[c]{@{}l@{}}The address is 329 El Camino Real and it's 3 miles away \\ and there is no traffic\end{tabular} \\ 
 & \textit{Ptr-Unk} & Palo Alto Cafe is 4 miles away and PAD is no traffic \\ \cline{2-3} 
 & \textit{\textbf{Mem2Seq}} & Palo Alto Cafe is 4 miles away at 436 Alger Drive \\ \cline{2-3} 
 & \textbf{GOLD} & Palo Alto is located at 436 Alger Dr. \\ \hline
\end{tabular}
}
\end{table}

\begin{table}[H]
\centering
\caption{Example of generated responses for the In-Car Assistant on the scheduling domain.}
\label{TB:EXAMPL_DIALOG_2_MEM2SEQ}
\resizebox{0.7\linewidth}{!}{\textit{}
\begin{tabular}{|c|c|c|c|c|c}
\hline
\textbf{room} & \textbf{time} & \textbf{date} & \textbf{party} & \textbf{event} \\ \hline
conference room 100 & 3pm & Thursday & sales team & conference \\ \hline
 - & 11am & Tuesday & - & tennis activity \\ \hline
- & 7pm & Wednesday & - & dentist appointment \\ \hline
conference room 100 &  1pm & Tuesday & HR & meeting \\ \hline
- & 7pm & Friday & - & doctor appointment \\ \hline
- &  11am & Tuesday & - & lab appointment \\ \hline
\end{tabular}
}
\resizebox{0.6\linewidth}{!}{
\begin{tabular}{r|l}
\hline
\textbf{DRIVER}      & When's my tennis match?                    \\ \hline
\textit{Seq2seq}     & Your lab appointment is on Monday at       \\
\textit{+Att}        & Your lab appointment is on Tuesday at 11am \\
\textit{Ptr-Unk}      & Your tennis match is on PAD at 1pm         \\ \hline
\textit{Mem2Seq} & Your tennis is on Tuesday at 11am          \\ \hline
\textbf{GOLD}        & Your tennis match is Tuesday at 11am      \\  \hline
\end{tabular}
}
\end{table}

\begin{table}[H]
\centering
\caption{Example of generated responses for the In-Car Assistant on the weather domain.}
\label{TB:EXAMPL_DIALOG_3_MEM2SEQ}
\resizebox{\linewidth}{!}{
\begin{tabular}{|c|c|c|c|}
\hline
\textbf{location}                & \textbf{monday}                      & \textbf{tuesday}                  & \textbf{wednesday}                    \\ \hline
grand rapids                     & hot, low of 50F, high of 70F         & raining, low of 60F, high of 80F  & rain, low of 20F, high of 30F         \\ \hline
new york                         & misty, low of 30F, high of 50F       & snow, low of 70F, high of 80F     & cloudy, low of 20F, high of 30F       \\ \hline
boston                           & hail, low of 90F, high of 100F       & overcast, low of 60F, high of 70F & rain, low of 50F, high of 60F         \\ \hline
durham                           & hot, low of 90F, high of 100F        & dry, low of 60F, high of 80F      & misty, low of 60F, high of 80F        \\ \hline
san francisco                    & rain, low of 60F, high of 70F        & cloudy, low of 30F, high of 40F   & overcast, low of 90F, high of 100F    \\ \hline
carson                           & raining, low of 70F, high of 80F     & humid, low of 90F, high of 100F   & frost, low of 40F, high of 60F        \\ \hline
san jose                         & blizzard, low of 40F, high of 50F    & snow, low of 90F, high of 100F    & overcast, low of 60F, high of 80F     \\ \hline
\textbf{thursday}                    & \textbf{friday}                  & \textbf{saturday}                     & \textbf{sunday}                   \\ \hline
clear skies, low of 60F, high of 70F & warm, low of 70F, high of 90F    & foggy, low of 50F, high of 60F        & overcast, low of 50F, high of 60F \\ \hline
rain, low of 80F, high of 100F       & rain, low of 40F, high of 60F    & cloudy, low of 30F, high of 50F       & snow, low of 20F, high of 40F     \\ \hline
dew, low of 20F, high of 30F         & cloudy, low of 90F, high of 100F & overcast, low of 50F, high of 70F     & overcast, low of 80F, high of 90F \\ \hline
misty, low of 90F, high of 100F      & hot, low of 70F, high of 90F     & hail, low of 30F, high of 40F         & rain, low of 60F, high of 80F     \\ \hline
stormy, low of 90F, high of 100F     & cloudy, low of 90F, high of 100F & clear skies, low of 90F, high of 100F & foggy, low of 30F, high of 50F    \\ \hline
dry, low of 50F, high of 60F         & foggy, low of 80F, high of 100F  & overcast, low of 50F, high of 70F     & overcast, low of 60F, high of 70F \\ \hline
rain, low of 30F, high of 50F        & rain, low of 80F, high of 100F   & dry, low of 50F, high of 60F          & dry, low of 90F, high of 100F     \\ \hline
\end{tabular}
}
\begin{tabular}{r|l}
\hline
\textbf{DRIVER}      & Is it gonna snow in Durham during the week?                \\ \hline
\textit{Seq2seq}     & there is no snow in the forecast for this week             \\
\textit{+Att}        & there is no snow in the forecast for durham for durham     \\
\textit{Ptr-Unk}      & yes there is no snow in durham on week                     \\ \hline
\textit{Mem2Seq} & no snow, the forecast does not predict any snow on Durham  \\ \hline
\textbf{GOLD}        & Durham does not have any snow predicted this week          \\ \hline
\end{tabular}
\end{table}

\begin{table}[H]
\centering
\caption{Example of generated responses for the In-Car Assistant on the navigation domain.}
\label{TB:EXAMPL_DIALOG_4_MEM2SEQ}
\resizebox{0.7\linewidth}{!}{
\begin{tabular}{|c|c|c|c|c|}
\hline
\textbf{distance} & \textbf{traffic\_info} & \textbf{poi\_type} & \textbf{address}   & \textbf{poi}   \\ \hline
3 miles           & no traffic             & chinese restaurant & 593 Arrowhead Way  & Chef Chu's     \\ \hline
1 miles           & no traffic             & chinese restaurant & 669 El Camino Real & P.F. Changs    \\ \hline
5 miles           & road block nearby      & gas station        & 200 Alester Ave    & Valero         \\ \hline
3 miles           & no traffic             & pizza restaurant   & 528 Anton Ct       & Pizza My Heart \\ \hline
4 miles           & heavy traffic          & rest stop          & 753 University Ave & Comfort Inn    \\ \hline
5 miles           & heavy traffic          & chinese restaurant & 842 Arrowhead Way  & Panda Express  \\ \hline
2 miles           & heavy traffic          & pizza restaurant   & 704 El Camino Real & Pizza Hut      \\ \hline
6 miles           & no traffic             & friends house      & 864 Almanor Ln     & jacks house    \\ \hline
\end{tabular}
}
\resizebox{0.7\linewidth}{!}{
\begin{tabular}{r|l}
\hline
\textbf{DRIVER} & Find me the nearest gas station. \\\hline
\textit{Seq2seq} & the nearest parking garage is from at away \\
\textit{+att} & The is one is away away \\
\textit{PTRUNK} & PAD nearest gas station is is is \\ \hline
\textit{Mem2Seq} & The nearest gas station is gas station which is you you \\ \hline
\textbf{GOLD} & The nearest gas station is located 5 miles away. Need more info? \\ \hline
\end{tabular}
}
\end{table}

\begin{table}[H]
\centering
\caption{Example of generated responses for the In-Car Assistant on the navigation domain.}
\label{TB:EXAMPL_DIALOG_5_MEM2SEQ}
\resizebox{0.7\linewidth}{!}{
\begin{tabular}{|c|c|c|c|c|}
\hline
\textbf{Poi}             & \textbf{Distance} & \textbf{Traffic\_info} & \textbf{Poi\_type} & \multicolumn{1}{c|}{\textbf{Address}} \\ \hline
The Westin               & 4 miles           & no traffic             & rest stop          & 329 El Camino Real                    \\ \hline
Palo Alto Garage R       & 2 miles           & moderate traffic       & parking garage     & 481 Amaranta Ave                      \\ \hline
Mandarin Roots           & 4 miles           & moderate traffic       & chinese restaurant & 271 Springer Street                   \\ \hline
Sigona Farmers Market    & 4 miles           & no traffic             & grocery store      & 638 Amherst St                        \\ \hline
Whole Foods              & 2 miles           & heavy traffic          & grocery store      & 819 Alma St                           \\ \hline
Stanford Shopping Center & 2 miles           & moderate traffic       & shopping center    & 773 Alger Dr                          \\ \hline
toms house               & 1 miles           & no traffic             & friends house      & 580 Van Ness Ave                      \\ \hline
\end{tabular}}
\resizebox{0.7\linewidth}{!}{
\begin{tabular}{r|l}
\hline
\textbf{DRIVER}        & Give me directions to the closest grocery store \\
\textbf{CAR}           & \begin{tabular}[c]{@{}l@{}}There are Whole Foods 2 miles away and \\ Sigona Farmers Market 4 miles away. Where do we go?\end{tabular}         \\
\textbf{DRIVER}        & I need the one that avoids all heavy traffic \\ \hline
\textit{Seq2seq}       & he is is is is away at \\
\textit{+Att}           & Chef Chu s is located at 593 Arrowhead way \\
\textit{Ptr-Unk}        & PAD is the is PAD the to \\ \hline
\textit{Mem2Seq}   & \begin{tabular}[c]{@{}l@{}}Sigona Farmers Market is the only option i is 4 miles \\ away with no traffic, setting navigation now\end{tabular} \\ \hline
\textbf{GOLD} & \begin{tabular}[c]{@{}l@{}}Sigona Farmers Market is located 4 miles away with no traffic. \\ It's located at 638 Amherst St.\end{tabular}     \\ \hline
\end{tabular}}
\end{table}

\begin{figure}[H]
\centering
\includegraphics[width=0.75\linewidth]{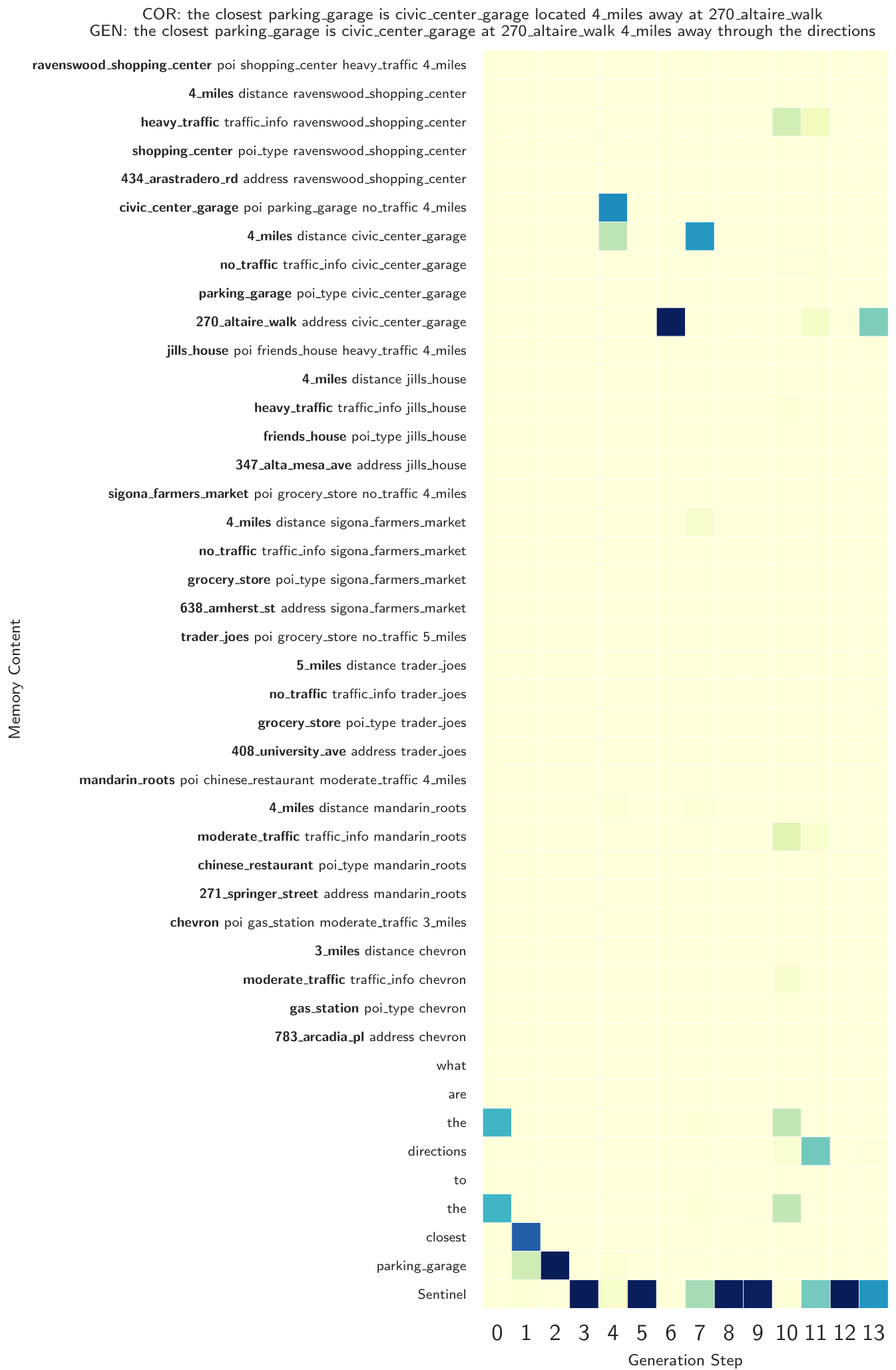}
\caption{Mem2Seq memory attention visualization of last hop. Y-axis is the concatenation of KB information and dialogue history, and x-axis is the decoding step.}
\label{FIG:VIZ-MEM2SEQ}
\end{figure}

\subsubsection{Visualization}
\paragraph{Memory Attention} Analyzing the attention weights has been frequently used to show the memory read-out since it is an intuitive way to understand the model dynamics. 
Figure~\ref{FIG:VIZ-MEM2SEQ} shows the attention vector at the last hop for each generated token. 
Each column represents the $P_{ptr}$ vector at the corresponding generation step. 
Our model has a sharp distribution over the memory, which implies that it is able to select the right token from the memory. 
For example, the KB information ``\textit{270\_altarie\_walk}'' was retrieved at the sixth step, which is an address for ``\textit{civic\_center\_garage}''. On the other hand, if the sentinel is triggered, then the generated word comes from vocabulary distribution $P_{vocab}$. For instance, the third generation step triggers the sentinel, and ``is'' is generated from the vocabulary, as the word is not present in the dialogue history. 

\begin{figure}[t]
\centering
\includegraphics[width=0.75\linewidth]{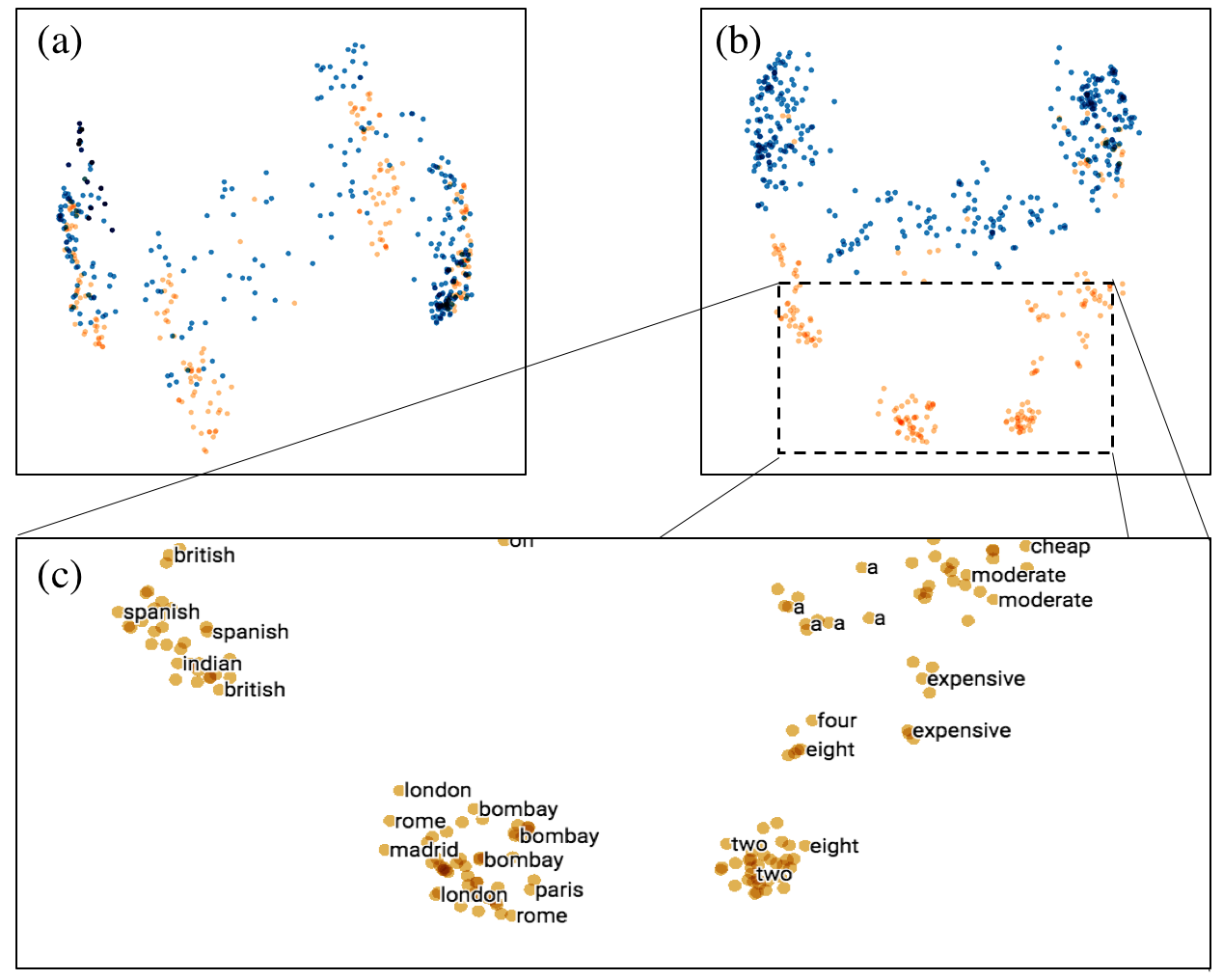}
\caption{Principal component analysis of Mem2Seq query vectors in hop (a) 1 and (b) 6. (c) a closer look at clustered tokens.}
\label{FIG:hopsQuery-Mem2Seq}
\end{figure}

\paragraph{Query Vectors} In Figure~\ref{FIG:hopsQuery-Mem2Seq}, the principal component analysis (PCA) of Mem2Seq query vectors are shown for different hops. 
Each dot is a query vector during each decoding time step, and it has a corresponding generated word. 
The blue dots are the words generated from $P_{vocab}$, which trigger the sentinel, and orange ones are from $P_{ptr}$. 
One can find that in (a), hop 1, there is no clear separation of the dots of the two different colors but they tend to group together. The separation becomes clearer in (b), hop 6, as dots of each color clusters into several groups, such as location, cuisine, and number. Our model tends to retrieve more information in the first hop, and points into the memories in the last hop.

\begin{figure}[H]
\centering
\includegraphics[width=\linewidth]{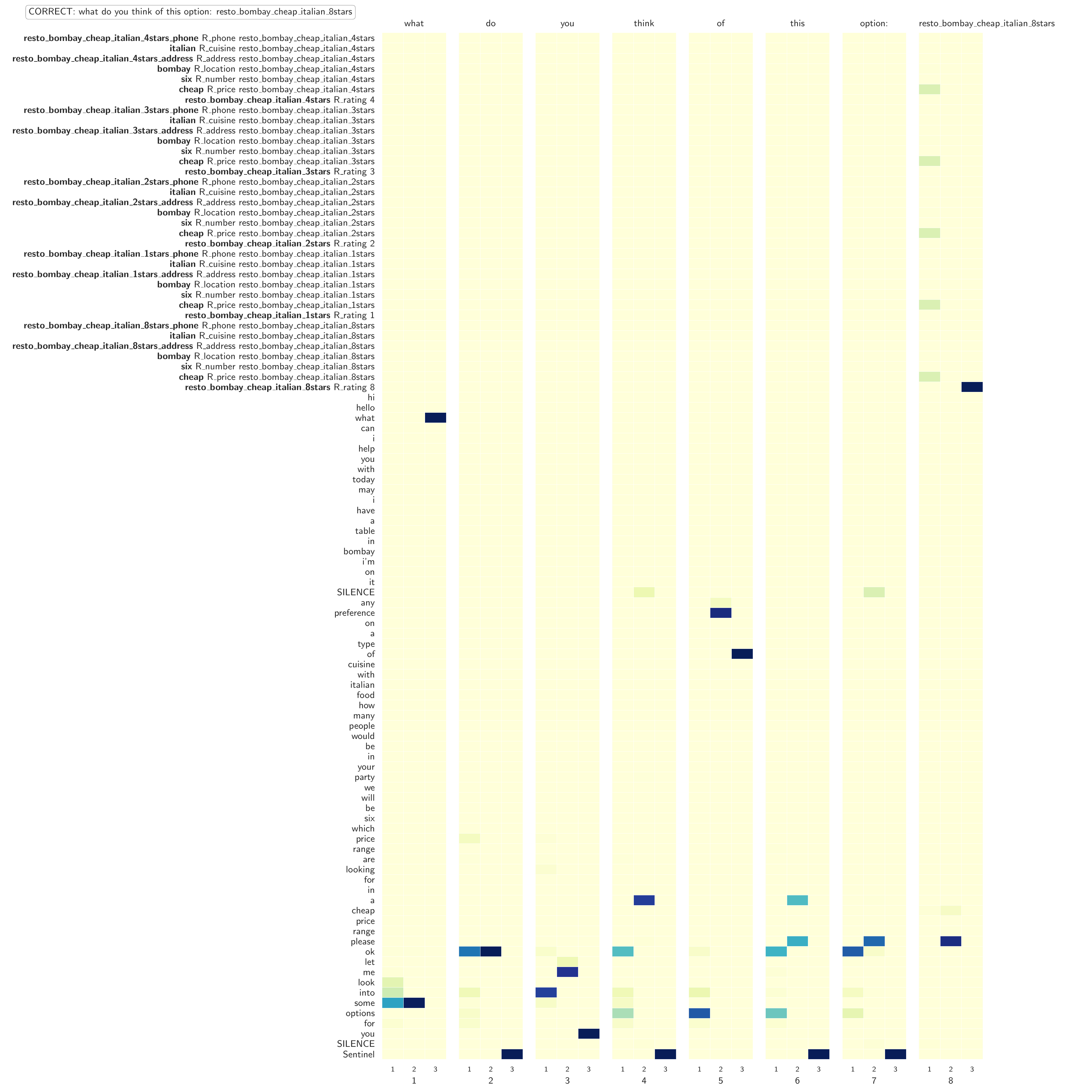}
\caption{Mem2Seq multi-hop memory attention visualization. Each decoding step on the x-axis has three hops, from loose attention to sharp attention.}
\label{3hop-mem2seq}
\end{figure}

\paragraph{Multiple Hops} In Figure~\ref{3hop-mem2seq}, Mem2Seq shows how multiple hops improve the model performance on several datasets. 
Task 3 in the bAbI dialogue dataset serves as an example, in which the systems need to recommend restaurants to users based on restaurant ranking from highest to lowest. Users can reject the recommendation and the system has to reason over the next highest restaurant. We find two common patterns between hops among different samples: 1) the first hop is usually used to score all the relevant memories and retrieve information; and 2) the last hop tends to focus on a specific token and makes mistakes when the attention is not sharp. 

\subsection{Short Summary}

We present an end-to-end trainable memory-to-sequence (Mem2Seq) model for task-oriented dialogue systems. Mem2Seq combines the multi-hop attention mechanism in end-to-end memory networks with the idea of pointer networks to incorporate external information.
It is a simple generative model that is able to incorporate KB information with promising generalization ability. 
We discover that the entity F1 score may be a more comprehensive evaluation metric than per-response accuracy or BLEU score, as humans can normally choose the right entities but have very diversified responses. 
Lastly, we empirically show our model's ability to produce relevant answers using both the external KB information and the predefined vocabulary, and visualize how the multi-hop attention mechanism helps in learning correlations between memories. Mem2Seq is fast, general, and able to achieve state-of-the-art results on three different datasets.

\newpage

\newpage

\section{Global-to-Local Memory Pointer Networks}
In the previous section, we discussed the first generative model that combines memory-augmented neural networks with copy mechanism to memorize long dialogue context and external knowledge. We empirically show the promising results of generated system responses in terms of BLEU score and entity F1 score on three different dialogue datasets. However, we found Mem2Seq tends to make the following errors while generating responses:
1) Wrong entity copying, as shown in Table~\ref{TB:EXAMPL_DIALOG_5_MEM2SEQ}. Although Mem2Seq can achieve the highest entity F1 score compared to existing baselines, it is only 33.4\%. 
2) The responses generated sometimes are not fluent, or even with several grammar mistakes, as shown in Table~\ref{TB:EXAMPL_DIALOG_4_MEM2SEQ}.  
Therefore, how to further improve the copy mechanism to obtain correct entity values becomes the next challenge, and how to maintain the fluency while balancing between generating from the vocabulary space and copying from the external knowledge.

In the section, we introduce the global-to-local memory pointer (GLMP) networks~\cite{wu2019global}, which is an extension of Mem2Seq. 
GLMP sketches system responses with unfilled slots, strengthens the copy mechanism using double pointers, and sharing memory representation in external knowledge for encoder and decoder. This model is composed of a global memory encoder, a local memory decoder, and a shared external knowledge. 
Unlike existing approaches with copy ability~\citep{gulcehre2016pointing,Gu2016IncorporatingCM,eric-manning:2017:EACLshort,mem2seq}, in which the only information passed to the decoder is the encoder hidden states, GLMP shares the external knowledge and leverages the encoder and the external knowledge to learn a global memory pointer and global contextual representation.
The global memory pointer modifies the external knowledge by softly filtering words that are not necessary for copying. 
Afterward, instead of generating system responses directly, the local memory decoder first uses a sketch RNN to obtain sketch responses without slot values but with sketch tags, which can be considered as learning latent dialogue management to generate dialogue action template. 
A similar intuition for generation sketching can be found in~\cite{mori2014flowgraph2text}, \cite{coarse2fine} and \cite{cai2018skeleton}.
Then the decoder generates local memory pointers to copy words from external knowledge and instantiate sketch tags. 

We empirically show that GLMP can achieve superior performance using the combination of global and local memory pointers. 
In simulated OOV tasks on the bAbI dialogue dataset~\cite{bordes2016learning}, GLMP achieves 92.0\% per-response accuracy and surpasses existing end-to-end approaches by 7.5\% on OOV full dialogue. 
On a human-human dialogue dataset~\citep{ericKVR2017}, GLMP is able to surpass the previous state-of-the-art, including Mem2Seq, on both automatic and human evaluation.

\subsection{Model Description}
\begin{figure}[t]
  \centering
  \includegraphics[width=0.75\linewidth]{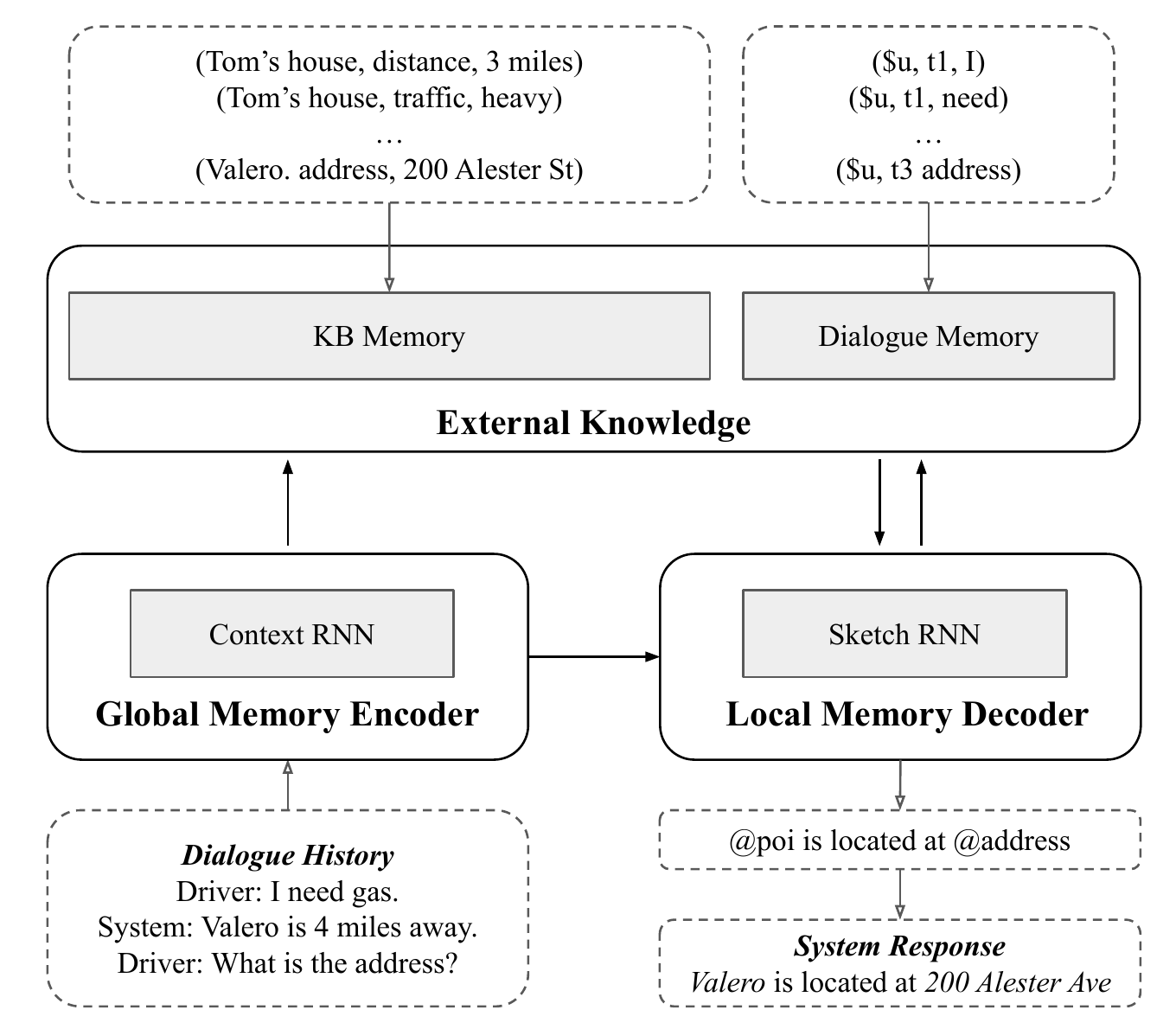}
\caption{The block diagram of global-to-local memory pointer networks. There are three components: global memory encoder, shared external knowledge, and local memory decoder.}
\label{FIG:glmp_blockdiagram}
\end{figure}

The GLMP model~\footnote{The code is available at \url{https://github.com/jasonwu0731/GLMP}} is composed of three parts: a global memory encoder, external knowledge, and local memory decoder, as shown in Figure~\ref{FIG:glmp_blockdiagram}. The dialogue history and the KB information are the input, and the system response is the expected output.
First, the global memory encoder uses a context RNN to encode dialogue history and writes its hidden states into the external knowledge. 
Then the last hidden state is used to read the external knowledge and generate the global memory pointer at the same time. 
On the other hand, during the decoding stage, the local memory decoder first generates sketch responses by a sketch RNN. 
Then the global memory pointer and the sketch RNN hidden state are passed to the external knowledge as a filter and a query. 
The local memory pointer returned from the external knowledge can copy text from the external knowledge to replace the sketch tags and obtain the final system response.

\subsubsection{External Knowledge}
\begin{figure}[t]
  \centering
  \includegraphics[width=0.5\linewidth]{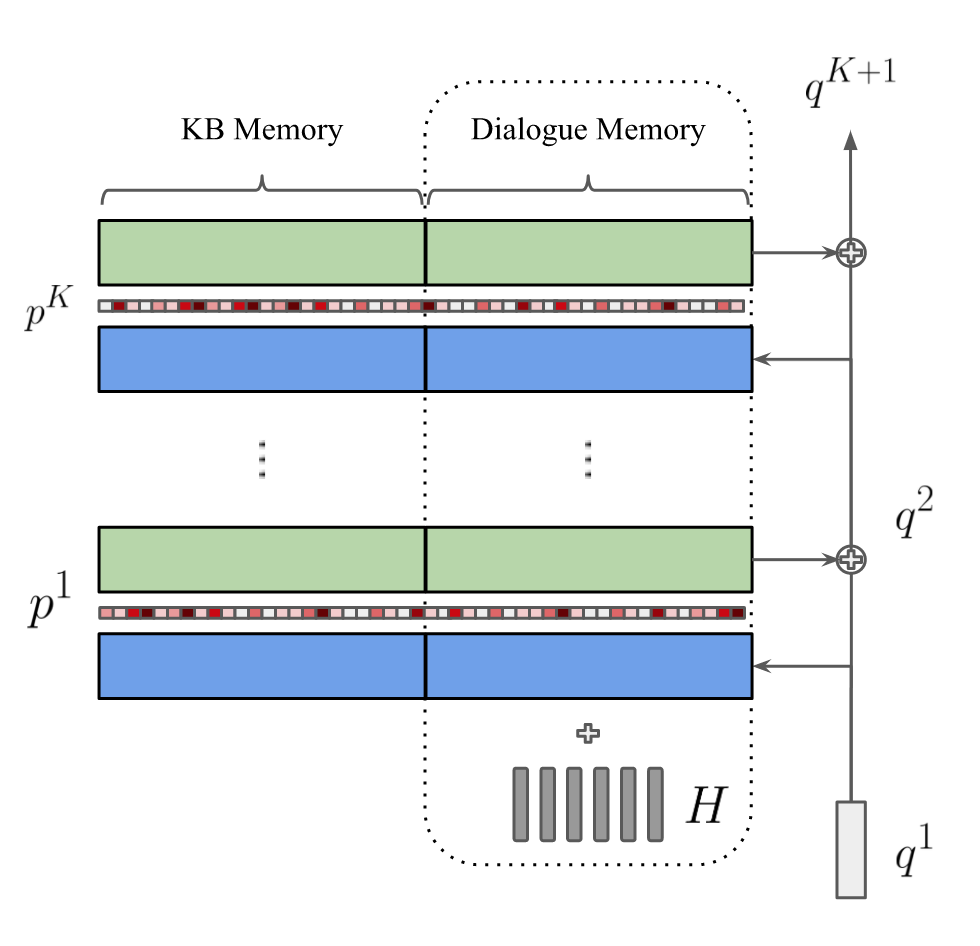}
\caption{GLMP external knowledge architecture, KB memory and dialogue memory. }
\label{FIG:glmp-externalknowledge}
\end{figure}

The external knowledge contains the global contextual representation that is shared with the encoder and the decoder. 
To incorporate external knowledge into a learning framework, end-to-end memory networks are used to store word-level information for both structural KB (KB memory) and temporal-dependent dialogue history (dialogue memory), as shown in Figure~\ref{FIG:glmp-externalknowledge}. 
In addition, the MN is well-known for its multiple hop reasoning ability ~\citep{sukhbaatar2015end}, which is appealing to strengthen the copy mechanism. 

\paragraph{Global Contextual Representation} In the KB memory module, each element is represented in the triplet format as a \textit{(Subject, Relation, Object)} structure, which is a common format used to represent KB nodes~\citep{millerEtAl2016,ericKVR2017}. 
Meanwhile, the dialogue context is stored in the dialogue memory module, where the speaker and temporal encoding are included, as in ~\cite{bordes2016learning}, in a triplet format. 
For the two memory modules, a bag-of-words representation is used as the memory embeddings. 
As in the design in~\cite{mem2seq}, during the inference time, we copy the object word once a memory position is pointed to. For example, \textit{3 miles} will be copied if the triplet \textit{(Toms house, distance, 3 miles)} is selected. 
We denote the $Object(.)$ function as getting the object word from a triplet.

\subsubsection{Global Memory Encoder}
\begin{figure}[t]
  \centering
  \includegraphics[width=0.7\linewidth]{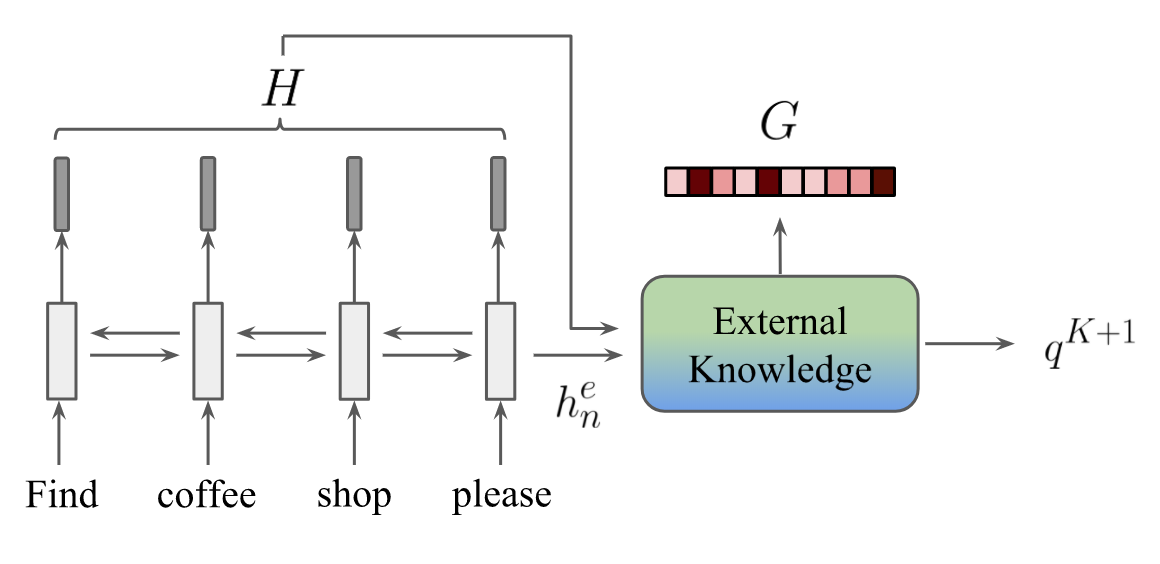}
\caption{GLMP global memory encoder architecture. }
\label{FIG:glmp-encoder}
\end{figure}

In Figure~\ref{FIG:glmp-encoder}, a context RNN is used to model the sequential dependency and encode the context. 
Then the hidden states are written into the external knowledge, as shown in Figure~\ref{FIG:glmp-externalknowledge}. 
Afterward, the last encoder hidden state serves as the query to read the external knowledge and get two outputs, the global memory pointer and the memory readout. Intuitively, since it is hard for MN architectures to model the dependencies between memories~\citep{dqmem8461426}, which is a serious drawback, especially in conversational related tasks, writing the hidden states to the external knowledge can provide sequential and contextualized information. 
With meaningful representation, our pointers can correctly copy out words from the external knowledge, and the common OOV challenge can be mitigated. 
In addition, using the encoded dialogue context as a query can encourage our external knowledge to read out memory information related to the hidden dialogue states or user intention. 
Moreover, the global memory pointer that learns a global memory distribution is passed to the decoder along with the encoded dialogue history and KB information.

\paragraph{Context RNN} A bi-directional GRU is used to encode dialogue history into the hidden states, and the last hidden state is used to query the external knowledge as the encoded dialogue history. 
In addition, the hidden states are written into the dialogue memory module in the external knowledge by summing up the original memory representation with the corresponding hidden states via:
\begin{equation}
    \begin{array}{c}
        c^k_i = c^k_i + h^{enc}_{m_i}  \quad \textrm{if} \quad m_i \in X \text{ and } \forall k \in [1,K+1]. \\
    \end{array}
\end{equation}

\paragraph{Global memory pointer} $G = (g_1,\dots,g_M)$ is a vector containing real values between 0 and 1, where $M$ is the total number of memories. 
Unlike the conventional attention mechanism in which all the weights sum to one, each element in $G$ is an independent probability.
We first query the external knowledge encoded dialogue history, and instead of applying the Softmax function as in (\ref{Eq:mnattention}), we perform an inner product followed by the Sigmoid function. 
The memory distribution we obtained is the global memory pointer $G$, which is passed to the decoder. 
To further strengthen the global pointing ability, we add an auxiliary loss to train the global memory pointer as a multi-label classification task. 
We show in an ablation study that adding this additional supervision does improve the performance. 
Lastly, the memory readout $q^{K+1}$ is used as the encoded KB information.

In the auxiliary task, we define the label $G^{label} = (g^l_1,\dots,g^l_{M})$ by checking whether the object words in the memory exists in the expected system response $Y$. 
Then the global memory pointer is trained using the binary cross-entropy loss $Loss_{g}$ between $G$ and $G^{label}$:
\begin{gather}
        g_i = \text{Sigmoid}((q^K)^T c^K_i), \quad 
        {g^l_i} = 
        \begin{cases} 
        1 &\mbox{if } Object(m_i) \in Y \\ 
        0 &\mbox{otherwise} 
        \end{cases}, \\
        Loss_g = -\sum_{i=1}^{M} [g^l_i \times \log{g_i}+(1-g^l_i) \times \log{(1-g_i)}]. 
\end{gather}

\begin{figure}[t]
  \centering
  \includegraphics[width=\linewidth]{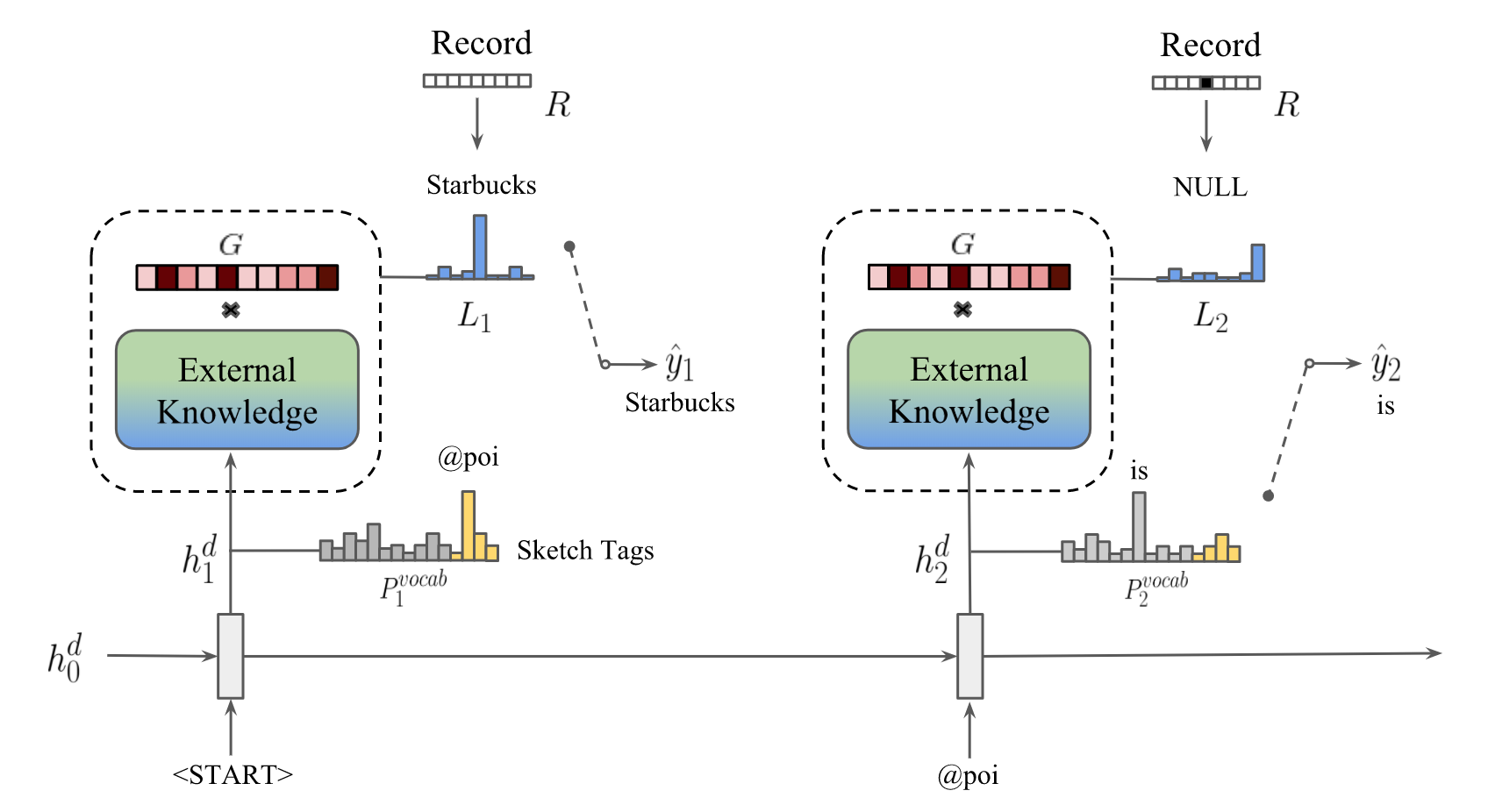}
\caption{GLMP local memory decoder architecture. }
\label{FIG:glmp-decoder}
\end{figure}

\subsubsection{Local Memory Decoder}
Given the encoded dialogue history, the encoded KB information, and the global memory pointer, our local memory decoder first initializes its sketch RNN using the concatenation of the encoded dialogue history and KB information, and generates a sketch response that excludes slot values but includes the sketch tags. 
For example, sketch RNN will generate ``\textit{@poi is @distance away}'' instead of ``\textit{Starbucks is 1 mile away.}'' 
At each decoding time step, the hidden state of the sketch RNN is used for two purposes: 1) predicting the next token in the vocabulary, which is the same as standard sequence-to-sequence (Seq2Seq) learning; and 2) serving as the vector to query the external knowledge. 
If a sketch tag is generated, the global memory pointer is passed to the external knowledge, and the expected output word will be picked up from the local memory pointer. 
Otherwise, the output word is the word generated by the sketch RNN. 
For example in Figure~\ref{FIG:glmp-decoder}, a @poi tag is generated at the first time step. Therefore, the word \textit{Starbucks} is picked up from the local memory pointer as the system output word.

\paragraph{Sketch RNN} We use a GRU to generate a sketch response $Y^s = (y^s_1,\dots,y^s_{|Y|})$ without real slot values. The sketch RNN learns to generate a dynamic dialogue action template based on the encoded dialogue and KB information. At each decoding time step $t$, the sketch RNN hidden state $h^{dec}_t$ and its output distribution $P^{vocab}_t$ are defined as
\begin{equation}
h^{dec}_t = \text{GRU}(C^1(\hat{y}^s_{t-1}), h^{dec}_{t-1}), \quad
P^{vocab}_t = \text{Softmax}(W h^{dec}_t).
\end{equation}
We use the standard cross-entropy loss to train the sketch RNN, and define $Loss_v$ as
\begin{equation}
Loss_v = \sum_{t=1}^{|Y|} - \log(P^{vocab}_t(y^s_t)).
\end{equation}
We transform the slot values in $Y$ into sketch tags based on the provided entity table. The sketch tags $ST$ are all the possible slot types that start with a special token. For example, \textit{@address} stands for all the addresses and \textit{@distance} stands for all the distance information. 

\paragraph{Local memory pointer} $L = (L_1,\dots,L_{|Y|})$ contains a sequence of pointers. 
At each time step $t$, the global memory pointer $G$ first modifies the global contextual representation using its attention weights,  
\begin{equation}
    \begin{array}{c}
        c^k_i = c^k_i \times g_i, \quad \forall i \in [1,M] \text{ and } \forall k \in [1,K+1],
    \end{array}
\end{equation}
and then the sketch RNN hidden state $h^{dec}_t$ queries the external knowledge. 
The memory attention in the last hop is the corresponding local memory pointer $L_t$, which is represented as the memory distribution at time step $t$. 
To train the local memory pointer, supervision on top of the last hop memory attention in the external knowledge is added. 
We first define the position label of local memory pointer $L^{label}$ at the decoding time step $t$ as
\begin{equation}
    L^{label}_t = 
    \begin{cases} 
        max(z) &\mbox{if } \exists z \ \text{s.t.} \ y_t = Object(m_z), \\ 
        M+1  &\mbox{otherwise.} 
    \end{cases} 
\end{equation}
The position $M$+1 is a null token in the memory that allows us to calculate the loss function even if $y_t$ does not exist in the external knowledge. 
Then, the loss between $L$ and $L^{label}$ is defined as 
\begin{equation}
Loss_l = \sum_{t=1}^{|Y|} - \log(L_t(L^{label}_t)).
\end{equation}
Furthermore, a record $R \in \mathbb{R}^{n+l}$ is utilized to prevent copying the same entities multiple times. All the elements in $R$ are initialized as 1 in the beginning. During the decoding stage, if a memory position has been pointed to, its corresponding position in $R$ will be masked out. During the inference time, $\hat{y}_t$ is defined as
\begin{equation}
    \hat{y}_t = 
    \begin{cases} 
        \argmax(P^{vocab}_t) &\mbox{if } \argmax(P^{vocab}_t) \not\in ST, \\ 
        Object(m_{\argmax(L_t \odot R)})  &\mbox{otherwise,} 
    \end{cases} 
\end{equation}
where $\odot$ is the element-wise multiplication. 
Lastly, all the parameters are jointly trained by minimizing the weighted-sum of three losses ($\alpha, \beta, \gamma$ are hyper-parameters):
\begin{equation}
Loss = \alpha Loss_g + \beta Loss_v + \gamma Loss_l.
\end{equation}

\subsection{Experimental Setup}
\subsubsection{Training}
The model is trained end-to-end using the Adam optimizer~\cite{kingma2014adam}, and learning rate annealing starts from $1e^{-3}$ to $1e^{-4}$. The number of hops $K$ is set to 1, 3, or 6 to compare the performance difference. The weights $\alpha$, $\beta$, and $\gamma$ summing the three losses, are set to 1. All the embeddings are initialized randomly, and a simple greedy strategy is used without beam-search during the decoding stage. The hyper-parameters such as hidden size and dropout rate are tuned with grid-search over the development set (per-response accuracy for bAbI dialogue and BLEU score for the In-Car Assistant).  In addition, to increase model generalization and simulate an OOV setting, we randomly mask a small number of input source tokens into an unknown token. 

\subsection{Results and Discussion}

\subsubsection{bAbI Dialogue}

\begin{table}[h]
\begin{center}
\caption{GLMP per-response accuracy and completion rate on bAbI dialogues.}
\label{TB:BABI-GLMP}
\resizebox{\linewidth}{!}{
\begin{tabular}{r|ccccc|ccc}
\hline
\multicolumn{1}{c|}{\textbf{Task}} & MN & GMN & S2S+Attn & Ptr-Unk & Mem2Seq & GLMP H1 & \multicolumn{1}{l}{GLMP H3} & \multicolumn{1}{l}{GLMP H6} \\ \hline
T1 & 99.9 (99.6) & 100 (100) & 100 (100) & 100 (100) & 100 (100) & 100 (100) & 100 (100) & 100 (100) \\
T2 & 100 (100) & 100 (100) & 100 (100) & 100 (100) & 100 (100) & 100 (100) & 100 (100) & 100 (100) \\
T3 & 74.9 (2.0) & 74.9 (0) & 74.8 (0) & 85.1 (19.0) & 94.7 (62.1) & \textbf{96.3 (75.6)} & 96.0 (69.4) & 96.0 (68.7) \\
T4 & 59.5 (3.0) & 57.2 (0) & 57.2 (0) & 100 (100) & 100 (100) & 100 (100) & 100 (100) & 100 (100) \\
T5 & 96.1 (49.4) & 96.3 (52.5) & 98.4 (87.3) & 99.4 (91.5) & 97.9 (69.6) & 99.2 (88.5) & 99.0 (86.5) & 99.2 (89.7) \\ \hline
T1 oov & 72.3 (0) & 82.4 (0) & 81.7 (0) & 92.5 (54.7) & 94.0 (62.2) & \textbf{100 (100)} & \textbf{100 (100)} & 99.3 (95.9) \\
T2 oov & 78.9 (0) & 78.9 (0) & 78.9 (0) & 83.2 (0) & 86.5 (12.4) & \textbf{100 (100)} & \textbf{100 (100)} & 99.4 (94.6) \\
T3 oov & 74.4 (0) & 75.3 (0) & 75.3 (0) & 82.9 (13.4) & 90.3 (38.7) & 95.5 (65.7) & \textbf{96.7 (72.9)} & 95.9 (67.7) \\
T4 oov & 57.6 (0) & 57.0 (0) & 57.0 (0) & 100 (100) & 100 (100) & 100 (100) & 100 (100) & 100 (100) \\
T5 oov & 65.5 (0) & 66.7 (0) & 65.7 (0) & 73.6 (0) & 84.5 (2.3) & \textbf{92.0 (21.7)} & 91.0 (17.7) & 91.8 (21.4) \\ \hline
\end{tabular}
}
\end{center}
\end{table}

In Table~\ref{TB:BABI-GLMP}, we follow~\citet{bordes2016learning} to compare the performance of our model and baselines based on per-response accuracy and task-completion rate. 
Note that utterance retrieval methods, such as QRN, MN, and GMN, cannot correctly recommend options (T3) and provide additional information (T4), and a poor generalization ability is observed in the OOV setting, which shows around 30\% performance difference on Task 5. 
Although previous generation-based approaches (Ptr-Unk, Mem2Seq) mitigate the gap by incorporating a copy mechanism, the simplest cases such as generating and modifying API calls (T1, T2) still face a 6--17\% OOV performance drop. 
On the other hand, GLMP achieves the highest, 92.0\%, task-completion rate on the full dialogue task and surpasses baselines by a big margin, especially in the OOV setting. 
No per-response accuracy loss cab be seen for T1, T2, and T4 using only the single hop, and it only decreases 7--9\% on the full dialogue task. 

\subsubsection{In-Car Assistant}

\begin{table}[h]
\begin{center}
\caption{GLMP performance on In-Car Assistant dataset using automatic evaluation (BLEU and entity F1) and human evaluation (appropriate and humanlike).}
\label{TB:SMD-GLMP}
\resizebox{\linewidth}{!}{
\begin{tabular}{rccccccccc}
\multicolumn{10}{c}{Automatic Evaluation} \\ \hline
\multicolumn{1}{l|}{} & Rule-Based* & \multicolumn{1}{c|}{KVR*} & S2S & S2S + Attn & Ptr-Unk & \multicolumn{1}{c|}{Mem2Seq} & GLMP K1 & GLMP K3 & GLMP K6 \\ \hline
\multicolumn{1}{r|}{BLEU} & 6.6 & \multicolumn{1}{c|}{13.2} & 8.4 & 9.3 & 8.3 & \multicolumn{1}{c|}{12.6} & 13.83 & \textbf{14.79} & 12.37 \\ \hline
\multicolumn{1}{r|}{Entity F1} & 43.8 & \multicolumn{1}{c|}{48.0} & 10.3 & 19.9 & 22.7 & \multicolumn{1}{c|}{33.4} & 57.25 & \textbf{59.97} & 53.54 \\ \hline
\multicolumn{1}{r|}{Schedule F1} & 61.3 & \multicolumn{1}{c|}{62.9} & 9.7 & 23.4 & 26.9 & \multicolumn{1}{c|}{49.3} & 68.74 & \textbf{69.56} & 69.38 \\
\multicolumn{1}{r|}{Weather F1} & 39.5 & \multicolumn{1}{c|}{47.0} & 14.1 & 25.6 & 26.7 & \multicolumn{1}{c|}{32.8} & 60.87 & \textbf{62.58} & 55.89 \\
\multicolumn{1}{r|}{Navigation F1} & 40.4 & \multicolumn{1}{c|}{41.3} & 7.0 & 10.8 & 14.9 & \multicolumn{1}{c|}{20.0} & 48.62 & \textbf{52.98} & 43.08 \\ \hline
\multicolumn{1}{l}{} & \multicolumn{1}{l}{} & \multicolumn{1}{l}{} &  & \multicolumn{1}{l}{} & \multicolumn{1}{l}{} & \multicolumn{1}{l}{} & \multicolumn{1}{l}{} & \multicolumn{1}{l}{} & \multicolumn{1}{l}{} \\
\multicolumn{10}{c}{Human Evaluation} \\ \hline
\multicolumn{1}{l|}{} & \multicolumn{3}{c|}{Mem2Seq} & \multicolumn{3}{c|}{GLMP} & \multicolumn{3}{c}{Human} \\ \hline
\multicolumn{1}{r|}{Appropriate} & \multicolumn{3}{c|}{3.89} & \multicolumn{3}{c|}{4.15} & \multicolumn{3}{c}{4.6} \\
\multicolumn{1}{r|}{Humanlike} & \multicolumn{3}{c|}{3.80} & \multicolumn{3}{c|}{4.02} & \multicolumn{3}{c}{4.54} \\ \hline
\end{tabular}
}
\end{center}
\end{table}

For a human-human dialogue scenario, we follow previous dialogue works~\cite{ericKVR2017,zhao2017generative,mem2seq} to evaluate our system on two automatic evaluation metrics, BLEU and entity F1 score. 
As shown in Table~\ref{TB:SMD-GLMP}, GLMP achieves the highest BLEU and entity F1 scores of 14.79 and 59.97\% respectively, which represent a slight improvement in BLEU but a huge gain in entity F1. 
In fact, for unsupervised evaluation metrics in task-oriented dialogues, we argue that the entity F1 might be a more comprehensive evaluation metric than per-response accuracy or BLEU, as it is shown in~\cite{ericKVR2017} that humans are able to choose the right entities but have very diversified responses. 
Note that the results of rule-based and KVR are not directly comparable because they simplify the task by mapping the expression of entities to a canonical form using named entity recognition and linking. 
For example, they compared in ``\textit{@poi is @poi\_distance away},'' instead of ``\textit{Starbucks is 1\_mile away.''}

Moreover, human evaluation of the generated responses is reported. 
We compare GLMP with the state-of-the-art model Mem2Seq and the original dataset responses as well. 
We randomly select 200 different dialogue scenarios from the test set to evaluate three different responses. 
Amazon Mechanical Turk is used to evaluate system appropriateness and human-likeness on a scale from 1 to 5. 
With the results shown in Table~\ref{TB:SMD-GLMP}, we can see that GLMP outperforms Mem2Seq on both measures, which is consistent with the previous observation. 
We also see that human performance on this assessment sets the upper bound on scores, as expected. 


\subsubsection{Ablation Study}

\begin{table}[h]
\centering
\caption{Ablation study using single hop model. Numbers in parentheses indicate how seriously the performance drops.}
\label{TB:ABLATION-GLMP}
\resizebox{0.85\linewidth}{!}{
\begin{tabular}{r|ccccc|c}
\hline
\multicolumn{1}{l|}{} & \multicolumn{5}{c|}{\begin{tabular}[c]{@{}c@{}}bAbI Dialogue OOV\\ Per-response Accuracy\end{tabular}} & \begin{tabular}[c]{@{}c@{}}In-Car Assistant\\ Entity F1\end{tabular} \\ \cline{2-7} 
 & T1 & T2 & T3 & T4 & T5 & All \\ \hline
GLMP & 100 (-) & 100 (-) & 95.5 (-) & 100 (-) & 92.0 (-) & 57.25 (-) \\
GLMP w/o H & 90.4 (-9.6) & 85.6 (-14.4) & 95.4 (-0.1) & 100 (-0) & 86.2 (-5.3) & 47.96 (-9.29) \\
GLMP w/o G & 100 (-0) & 91.7 (-8.3) & 95.5 (-0) & 100 (-0) & 92.4 (+0.4) & 45.78 (-11.47) \\ \hline
\end{tabular}
}
\end{table}

The contributions of the global memory pointer $G$ and the memory writing of dialogue history $H$ are shown in Table~\ref{TB:ABLATION-GLMP}. 
We compare the results using GLMP with one hop in the bAbI OOV setting and In-Car Assistant. 
GLMP without $H$ means that the context RNN in the global memory encoder does not write the hidden states into external knowledge. 
As one can observe, GLMP without $H$ has 5.3\% more loss on the full dialogue task. 
On the other hand, GLMP without $G$ means that we do not use the global memory pointer to modify the external knowledge, and an 11.47\% entity F1 drop can be observed on the In-Car Assistant dataset. 
Note that a 0.4\% increase can be observed on task 5, which suggests that the use of the global memory pointer may impose too strong prior entity probability. 

\begin{figure}[t]
\centering
\includegraphics[width=\linewidth]{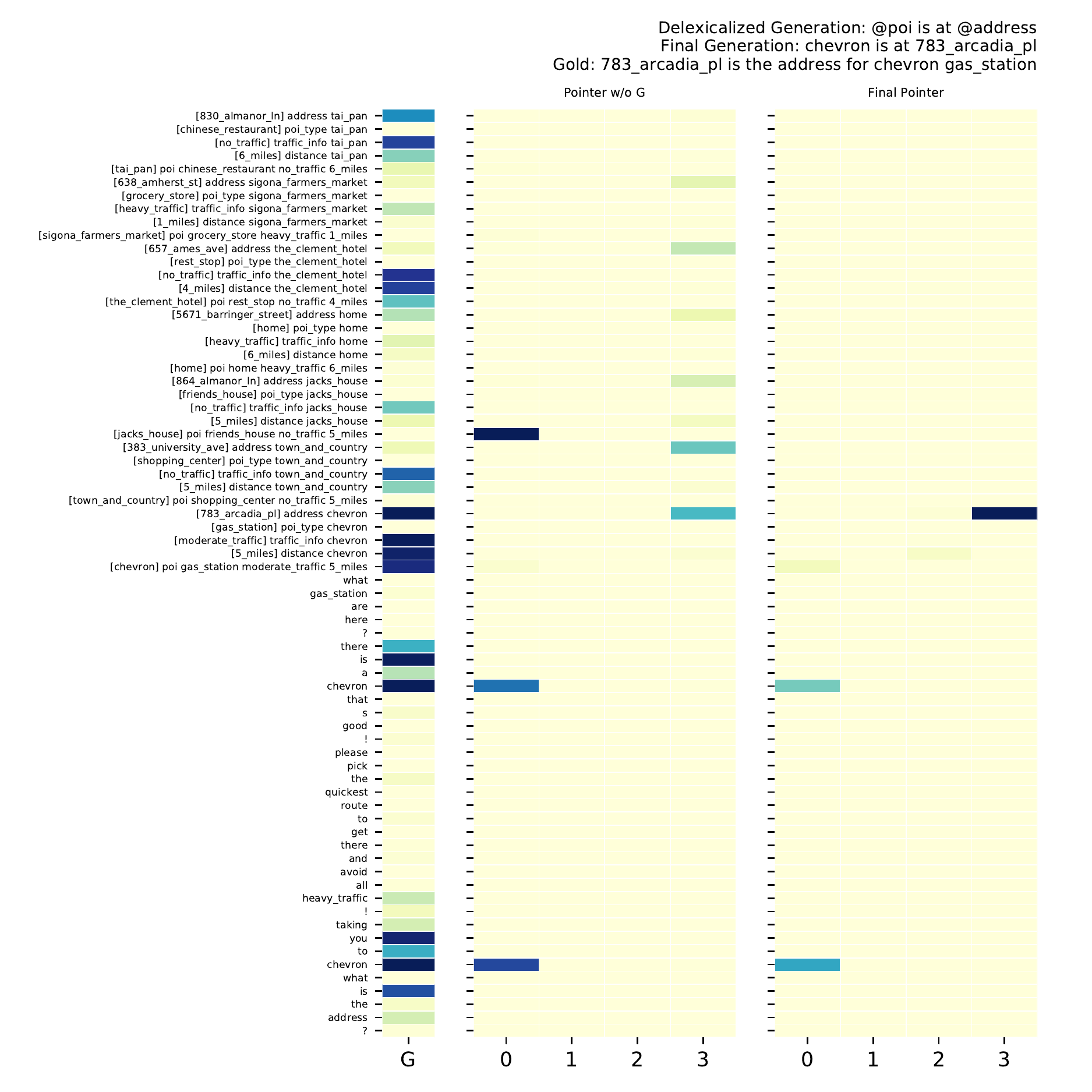} 
\caption{Memory attention visualization in the In-Car Assistant dataset. The left column is the memory attention of global memory pointer, the right column is the local memory pointer over four decoding steps. The middle column is the local memory pointer without weighted by global memory pointer.}
\label{FIG:VIZ-GLMP}
\end{figure}

\subsubsection{Visualization and Qualitative Evaluation}
Analyzing the attention weights has been frequently used to interpret deep learning models. In Figure~\ref{FIG:VIZ-GLMP}, we show the attention vector in the last hop for each generation time step. 
The Y-axis is the external knowledge that we can copy, including the KB information and dialogue history. 
Based on the question ``\textit{What is the address?}'' asked by the driver in the last turn, the gold answer and our generated response are on the top, and the global memory pointer $G$ is shown in the left column. 
One can observe that in the right column, the final memory pointer has successfully copied the entity \textit{chevron} in step 0 and its address \textit{783 Arcadia Pl} in step 3 to fill in the sketch utterance. 
Memory attention without global weighting is reported in the middle column, and one can find that even if the attention weights focus on several points of interest and addresses in step 0 and step 3, the global memory pointer can mitigate the issue as expected. 

\subsubsection{Error Analysis}
We also check the main mistakes made by our model and analyze the errors.
For the bAbI dialogues, the mistakes are mainly from task 3, which is recommending restaurants based on their rating from high to low. 
We find that sometimes the system will keep sending restaurants with a higher score even if the user has rejected them in the previous turns. 
Meanwhile, the In-Car Assistant is more challenging for response generation. 
First, we find that the model makes mistakes when the KB has several options corresponding to the user intention. 
For example, once the user has more than one doctor's appointment in the table, the model can barely recognize them. 
In addition, since we do not include the domain specific and user intention supervision, wrong delexicalized responses may be generated, which results in an incorrect entity copy. 
Lastly, we find that the copied entities may not match the generated sketch tags. 
For example, an address tag may result in a totally unrelated entity copy such as ``4 miles''.

\subsection{Short Summary}
We present an end-to-end trainable model called a global-to-local memory pointer network (GLMP) for task-oriented dialogues to effectively incorporate long dialogue context and external knowledge. It is an extension of the Mem2Seq generation model, to address the wrong entity copying and response fluency problems. The global memory encoder and the local memory decoder are designed to incorporate shared external knowledge into the learning framework. We empirically show that the global and the local memory pointer are able to effectively produce system responses even in the out-of-vocabulary scenario, and visualize how the global memory pointer helps as well. As a result, our model achieves state-of-the-art results on both simulated and human-human dialogue datasets.

%% file: chapter6.tex
\chapter{Conclusion}

In this thesis, we focus on learning task-oriented dialogue systems with deep learning models. We pointed out challenges in existing approaches to modeling long dialogue context and external knowledge in conversation and optimizing dialogue systems end-to-end. To effectively address these challenges, we incorporated and strengthened the neural copy mechanism with memory-augmented neural networks, and leveraged this strategy to achieve state-of-the-art performance in multi-domain dialogue state tracking, retrieval-based dialogue systems, and generation-based dialogue systems. In this chapter, we conclude the thesis and discuss possible future work.

In Chapter 3, we showed an end-to-end generative model with a copy mechanism can be used in dialogue state tracking, and sharing in multiple domains can further improve the performance. Our proposed dialogue state generator (TRADE) achieved state-of-the-art results in multi-domain dialogue state tracking. We also demonstrated how to track unseen domains by transferring knowledge from learned domains, or to quickly adapt to a new domain without forgetting the learned domains.

In Chapter 4, we leveraged recurrent entity networks and proposed dynamic query memory networks for end-to-end retrieval-based dialogue learning. We obtained state-of-the-art per-response/per-dialogue accuracy via modeling sequential dependencies and external knowledge using memory-augmented neural networks. In addition, we demonstrated how a simple copy mechanism, recorded delexicalized copying, can be used to reduce learning complexity and improve model generalization ability.

In Chapter 5, we introduced two neural models with a copy mechanism that achieved state-of-the-art performance on end-to-end response generation tasks. We presented the first machine learning model (Mem2Seq) that combines a multi-hop attention mechanism with the idea of pointer networks. Then we presented its extension model (GLMP), which strengthens the copying accuracy with double pointers and response sketching. We showed that both models outperform existing generation approaches, not only in terms of automatic evaluation metrics, such as BLEU and F1 score, but also on human evaluation metrics, such as appropriateness and human likeness.

Finally, conventional task-oriented dialogue systems~\cite{williams2007partially}, which are still widely used in commercial systems, require significant amounts human effort in system design and data collection. 
End-to-end dialogue systems, although not perfect yet, require much less human involvement, especially in the dataset construction, as raw conversational text and KB information can be used directly without the need of heavy pre-processing, e.g., named-entity recognition and dependency parsing. 

In future works, several methods could be applied (e.g. Reinforcement Learning~\cite{ranzato2015sequence}, Beam Search~\cite{wiseman2016sequence}) to improve both responses relevance and entity F1 score. However, we preferred to keep our model as simple as possible in order to show that it works well even without advanced training methods. Also it is possible to further improve the copying accuracy using different kinds of memory-augmented neural networks or using hierarchical structures to store external knowledge. In multi-domain dialogue state tracking, we expect to further improve zero-shot or few-shot learning by collecting a larger dataset with more domains. We believe end-to-end systems can be further improved by taking steps toward multi-task training as well, e.g., joint training dialogue state tracking in response generation tasks. It will be interesting to see how the concept of conventional pipeline dialogue approaches can help the design of end-to-end dialogue learning in task-oriented dialogue systems.

%% file: publication.tex
\null\skip0.2in
\begin{center}
{\bf \Large \underline{List of Publications}}
\end{center}
\vspace{12mm}

(* denotes equal contribution)

\begin{enumerate}
    \item Chien-Sheng Wu, Andrea Madotto, Ehsan Hosseini-Asl, Caiming Xiong, Richard Socher, and Pascale Fung. ``Transferable Multi-Domain Dialogue State Generators for Task-Oriented Dialogue Systems.''  In \textit{Proceedings of the 57th Annual Meeting of the Association for Computational Linguistics (ACL)}
    

    \item Chien-Sheng Wu, Richard Socher, and Caiming Xiong. ``Global-to-local Memory Pointer Networks for Task-Oriented Dialogue'' In \textit{Proceedings of the 7th International Conference on Learning Representations (ICLR)}, 2019.
    
    \item Chien-Sheng Wu*, Andrea Madotto*, and Pascale Fung. ``Mem2Seq: Effectively Incorporating Knowledge Bases into End-to-End Task-Oriented Dialog Systems.'' In \textit{Proceedings of the 56th Annual Meeting of the Association for Computational Linguistics (ACL)} (Volume 1: Long Papers). Vol. 1. 2018.
    
    \item Chien-Sheng Wu, Andrea Madotto, Genta Winata, and Pascale Fung. ``End-to-End Dynamic Query Memory Network for Entity-Value Independent Task-Oriented Dialog.'' In \textit{2018 IEEE International Conference on Acoustics, Speech and Signal Processing (ICASSP)}, pp. 6154-6158. IEEE, 2018.
    
    \item Chien-Sheng Wu*, Andrea Madotto*, Genta Winata, and Pascale Fung. ``End-to-end recurrent entity network for entity-value independent goal-oriented dialog learning.'' In \textit{Dialog System Technology Challenges Workshop, DSTC6}. 2017.
    
\end{enumerate}

%% file: reference.tex
\bibliographystyle{IEEEtran}
\bibliography{reference} 